\documentclass{article}

\PassOptionsToPackage{numbers, compress}{natbib}
    \usepackage[final,main]{neurips_2026}

\usepackage[utf8]{inputenc} %
\usepackage[T1]{fontenc}    %
\usepackage[pagebackref]{hyperref}       %
\renewcommand*{\backrefalt}[4]{%
  \ifcase#1 %
  \or
    \,Cited on p.~#2.%
  \else
    \,Cited on pp.~#2.%
  \fi
}
\usepackage{url}            %
\usepackage{booktabs}       %
\usepackage{longtable}      %
\usepackage{multirow}       %
\usepackage{amsfonts}       %
\usepackage{nicefrac}       %
\usepackage{microtype}      %
\usepackage[table]{xcolor}  %

\usepackage{enumitem}
\usepackage{mleftright}

\title{Variational Monte Carlo for Quantum Excited States via Nested Low-Rank Approximation}

\author{%
    \textbf{Minchan Jeong}\textsuperscript{1}\thanks{The two authors contributed equally to this work.}$\hphantom{^{*}}$\thanks{Work done while a Ph.D. student at KAIST.}$\hphantom{^{\dagger}}$,
    \textbf{J.~Jon~Ryu}\textsuperscript{2}${^{*}}$\thanks{Work done while a postdoctoral researcher at MIT.}$\hphantom{^{*}}$,
    \textbf{Se-Young Yun}\textsuperscript{3},
    \textbf{Gregory W. Wornell}\textsuperscript{4}\\
    \textsuperscript{1}Center for AI and Natural Sciences, KIAS, Seoul 02455, South Korea\\
    \textsuperscript{2}Department of CSE, Penn State University, University Park, PA 16802, USA \\
    \textsuperscript{3}Kim Jaechul Graduate School of AI, KAIST, Daejeon 34141, South Korea\\
    \textsuperscript{4}Department of EECS, MIT, Cambridge, MA 02139, USA \\
    \texttt{\href{mailto:mcjeong@kias.re.kr}{mcjeong@kias.re.kr}, 
    \href{mailto:jongha@psu.edu}{jongha@psu.edu},
    \href{mailto:yunseyoung@kaist.ac.kr}{yunseyoung@kaist.ac.kr}, \href{mailto:gww@mit.edu}{gww@mit.edu}}
}

\usepackage[most]{tcolorbox}

\usepackage{xspace}
\usepackage{bbm}
\input{mathlig}
\usepackage{mathtools}
\usepackage{relsize}
\usepackage{mathrsfs}
\usepackage{dsfont}
\DeclarePairedDelimiterX{\inp}[2]{\langle}{\rangle}{#1, #2}
\makeatletter
\newcommand*\bigcdot{\mathpalette\bigcdot@{.5}}
\newcommand*\bigcdot@[2]{\mathbin{\vcenter{\hbox{\scalebox{#2}{$\m@th#1\bullet$}}}}}
\makeatother

\newcommand{\muspace}{\mspace{1mu}}

\DeclareRobustCommand{\scond}{\mathchoice{\muspace\vert\muspace}{\vert}{\vert}{\vert}}
\mathlig{|}{\scond}

\DeclareRobustCommand{\discint}{\mathchoice{\mspace{-1.5mu}:\mspace{-1.5mu}}{\mspace{-1.5mu}:\mspace{-1.5mu}}{:}{:}}
\mathlig{::}{\discint}

\def\tr{\mathop{\rm tr}\nolimits}%
\def\Cov{\mathop{\rm Cov}\nolimits}%
\def\diag{\mathop{\rm diag}\nolimits}%
\def\Re{\mathop{\rm Re}\nolimits}%
\def\sgn{\mathop{\rm sgn}\nolimits}%

\newcommand{\Cc}{\mathcal{C}}

\newcommand{\Ec}{\mathcal{E}}

\newcommand{\Hc}{\mathcal{H}}

\newcommand{\Jc}{\mathcal{J}}

\newcommand{\Lc}{\mathcal{L}}

\newcommand{\Vc}{\mathcal{V}}

\renewcommand{\Pr}{\mathbb{P}}

\newcommand{\Xv}{{\bf X}}

\newcommand{\xv}{{\bf x}}

\def\a{\alpha}

\def\th{\theta}

\DeclareMathOperator\E{\mathsf{E}}

\def\textiid{i.i.d.\@\xspace}
\newcommand\iid{\ifmmode\text{ i.i.d. } \else \textiid \fi}

\newcommand{\Complex}{\mathbb{C}}

\newcommand{\Real}{\mathbb{R}}

\newcommand{\ones}{\mathds{1}}
\newcommand{\half}{\frac{1}{2}}%

\def\clap#1{\hbox to 0pt{\hss#1\hss}}
\def\mathclap{\mathpalette\mathclapinternal}
\def\mathclapinternal#1#2{%
  \clap{$\mathsurround=0pt#1{#2}$}}

\let\oldstackrel\stackrel
\renewcommand{\stackrel}[2]{\oldstackrel{\mathclap{#1}}{#2}}

\DeclarePairedDelimiterX{\infdivx}[2]{(}{)}{%
  #1\;\delimsize\|\;#2%
}

\catcode`~=11 %
\newcommand{\urltilde}{\kern -.06em\lower -.06em\hbox{~}\kern .02em}
\catcode`~=13 %

\DeclarePairedDelimiterX{\norm}[1]{\lVert}{\rVert}{#1}
\DeclarePairedDelimiterX{\abs}[1]{\lvert}{\rvert}{#1}

\usepackage{xparse}

\newcommand*\diff{\mathop{}\!\mathrm{d}}

\let\oldpartial\partial
\renewcommand*{\partial}{\mathop{}\!\oldpartial}
\newcommand{\defeq}{\mathrel{\mathop{:}}=}

\newcommand{\Iop}{\mathpzc{I}}
\newcommand{\ip}[2]{\langle #1 ~|~ #2 \rangle}
\newcommand{\Top}{\mathpzc{T}} %
\newcommand{\Oop}{\mathpzc{O}} %

\newcommand{\Imat}{\mathsf{I}}

\renewcommand{\E}{\mathbb{E}}
\newcommand{\hmltn}{\mathpzc{H}}

\renewcommand{\Hc}{\mathscr{H}}

\newcommand{\rv}{\mathbf{r}}
\newcommand{\Rv}{\mathbf{R}}
\newcommand{\numel}{{n_{\mathsf{el}}}}
\newcommand{\numnu}{{n_{\mathsf{nu}}}}

\usepackage{yhmath}

\newcommand{\lora}{\mathsf{LoRA}}

\newcommand{\algo}{{NestedLoRA-VMC}\xspace}

\newcommand{\kidx}{k}

\newcommand{\totalhmltn}{\mathfrak{H}}
\newcommand{\Sm}{\mathsf{S}}
\newcommand{\Hm}{\mathsf{H}}
\newcommand{\idop}{\Iop}
\newcommand{\locmat}{\mathsf{E}_{\mathsf{loc}}}

\newcommand{\matmask}{\mathsf{M}}
\newcommand{\eigvalparam}{\lambda}

\newcommand{\stopgrad}{\mathsf{sg}}

\usepackage{makecell}
\usepackage{tablefootnote}
\usepackage{pdflscape}

\usepackage{eufrak}
\DeclareMathAlphabet{\mathpzc}{OT1}{pzc}{m}{it}

\usepackage{amsmath}
\usepackage{amssymb}
\usepackage{mathtools}
\usepackage{amsthm}

\usepackage{thmtools}
\usepackage{thm-restate}
\declaretheorem[name=Theorem,numberwithin=section]{theorem}
\declaretheorem[name=Proposition,numberwithin=section]{proposition}
\declaretheorem[name=Lemma,numberwithin=section]{lemma}
\declaretheorem[name=Corollary,numberwithin=section]{corollary}

\usepackage{scrwfile}
\TOCclone[\appendixname]{toc}{atoc}
\newcommand\StartAppendixEntries{}
\AfterTOCHead[toc]{%
  \renewcommand\StartAppendixEntries{\value{tocdepth}=-10000\relax}%
}
\AfterTOCHead[atoc]{%
  \value{tocdepth}=-10000\relax%
  \renewcommand\StartAppendixEntries{\value{tocdepth}=2\relax}%
}
\usepackage[normalem]{ulem}

\newcommand{\Rm}{\mathsf{R}}

\newcommand{\seq}{\mathsf{seq}}

\usepackage{tikz}
\usetikzlibrary{positioning, arrows.meta, calc, fit, backgrounds, shapes.geometric}

\usepackage{titlesec}
\titleformat{\paragraph}[runin]
  {\normalfont\bfseries}{}{0pt}{}[]
\titlespacing*{\paragraph}{0pt}{\medskipamount}{0.5em}

\begin{document}

\maketitle

\begin{abstract}
The electronic Schr\"odinger equation encodes a molecule's electronic structure and quantum properties, but solving it scales exponentially with the number of electrons.
Solving this equation with neural-network variational Monte Carlo (VMC) has reached chemical accuracy on the ground state of small molecules; extending this success to excited states is constrained by two existing paradigms.
Penalty-based methods need $K$ Laplacians per step but require tuning the overlap-penalty weights against the energy gaps they aim to compute.
Natural excited states VMC (NES-VMC) is penalty-free, but it combines the $K$ ansatzes into one determinant, whose local energy requires $K^2$ Laplacians per step.
We resolve this trade-off via low-rank approximation (LoRA) reparameterization of the rank-$K$ variational objective.
The reparameterized objective admits a VMC-compatible stochastic gradient estimator via MCMC sampling from the per-state Born densities, and a sequential nesting scheme recovers the lowest $K$ eigenstates in eigenvalue order.
The resulting algorithm, NestedLoRA-VMC, inherits a global optimality guarantee from the LoRA principle.
On the first-row atoms Li through Ne at $K=10$, its accuracy is comparable to that of NES-VMC and of a state-of-the-art penalty-based method.
\end{abstract}

\section{Introduction}
Ab initio quantum chemistry aims to compute the electronic states of molecules by solving the Schr\"odinger equation $\hmltn\psi=E\psi$.
Accurate computation of these states remains challenging due to the curse of dimensionality: the wavefunction lives on the $3\numel$-dimensional configuration space of $\numel$ electrons, and the computational cost of traditional high-accuracy methods grows steeply with $\numel$.
Recently, neural-network variational Monte Carlo (VMC) has emerged as a promising alternative, achieving chemical accuracy for the ground states of small molecules with architectures such as FermiNet~\citep{Pfau--Spencer--Matthews--Foulkes2020}, PauliNet~\citep{Hermann--Schatzle--Noe2020}, and Psiformer~\citep{glehn2023a}.
However, extending this success to excited states remains challenging.
While minimizing the Rayleigh quotient naturally characterizes the ground state, computing multiple excited states requires an additional mechanism to recover mutually orthogonal eigenfunctions and identify their energy ordering.
Moreover, such a mechanism must be compatible with VMC, where optimization relies on Monte Carlo samples drawn from the Born densities of the neural-network ansatzes.

Existing excited-state VMC methods can be broadly categorized into two approaches.
On the one hand, \emph{penalty-based methods}~\citep{Entwistle--Schatzle--Erdman--Hermann--Noe2023,Pathak--Busemeyer--Rodrigues--Wagner2021,Wheeler--Kleiner--Wagner2024,Szabo--Schatzle--Entwistle--Noe2024} minimize the sum of the Rayleigh quotients of multiple ansatzes, augmented with penalties on their pairwise overlaps to encourage orthogonality.
While conceptually straightforward, these methods require careful tuning of the penalty weights.
In particular, \citet{Wheeler--Kleiner--Wagner2024} show that recovering the desired eigenstates at the global minimum requires sufficiently large penalty weights relative to the energy gaps, which are themselves unknown quantities to be estimated.
On the other hand, \emph{natural excited states VMC} (NES-VMC)~\citep{Pfau--Axelrod--Sutterud--von-Glehn--Spencer2024} takes an elegant, penalty-free approach by transforming the excited-state problem into a ground-state problem of an extended quantum system.
By constructing a determinant from the individual ansatzes, NES-VMC implicitly enforces orthogonality without additional regularization.
The individual eigenstates are then recovered by diagonalizing the Hamiltonian projected onto the learned subspace.
However, this construction increases computational cost: evaluating its local energy requires $K^2$ Laplacian evaluations per optimization step for $K$ states, compared to $K$ for penalty-based methods.
Thus, the existing approaches present a tradeoff between penalty tuning and computational efficiency (Table~\ref{tab:positioning}).

We pursue a different variational perspective based on the low-rank approximation (LoRA) principle.
Specifically, we build on NestedLoRA~\citep{Ryu--Xu--Erol--Bu--Zheng--Wornell2024}, a general framework for learning ordered eigenfunctions of an operator using neural networks, without explicitly imposing orthogonality constraints or introducing penalty weights.
This framework has been successfully applied to low-dimensional partial differential equations and representation learning~\citep{Ryu--Xu--Erol--Bu--Zheng--Wornell2024}, as well as stochastic dynamical systems~\citep{Jeong--Ryu--Yun--Wornell2025}.
Despite these appealing properties, the LoRA objective is not directly compatible with VMC.
The main obstacle is that it depends on the norms of the learned functions, which involve high-dimensional integrals that cannot be directly estimated from samples of their normalized Born densities.

\paragraph{Contributions.}
We overcome this obstacle through a simple reparameterization, representing each function in the LoRA objective as a normalized neural-network ansatz multiplied by a trainable scale.
This makes the resulting objective compatible with the Born sampling used in VMC.
Building on this formulation, we propose \algo{}, which learns eigenstates directly in energy order through the \emph{sequential nesting} technique~\citep{Gao--Li--Lu2021,Ryu--Xu--Erol--Bu--Zheng--Wornell2024}.
The proposed framework is penalty-free, requires no post-training diagonalization, and retains the global optimality characterization of the underlying LoRA principle.
Moreover, \algo{} requires only $K$ Laplacian evaluations per optimization step, matching the cost of penalty-based methods rather than the $K^2$ evaluations of NES-VMC.
We evaluate \algo{} on first-row atoms from Li to Ne with $K=10$ states and demonstrate accuracy comparable to NES-VMC and the state-of-the-art penalty-based method of \citet{Szabo--Schatzle--Entwistle--Noe2024}, without requiring gap-dependent penalty weights.

\paragraph{Organization.}
Section~\ref{sec:prelim} introduces the problem and existing excited-state VMC approaches.
Section~\ref{sec:proposed_method} develops the proposed VMC-compatible LoRA formulation and its theoretical properties.
Section~\ref{sec:experiments} presents experimental results, and Section~\ref{sec:conclusion} concludes with limitations and future directions.
Proofs and additional technical details are deferred to the appendix.

\section{Preliminaries}
\label{sec:prelim}
Our goal is to recover the $K$ lowest eigenpairs $\{(\varphi_k, E_k)\}_{k=1}^K$ of the electronic Hamiltonian $\hmltn$ of a molecule on the space $\Hc$ of its wavefunctions, both defined below.
We order them as $E_1 \le E_2 \le \cdots \le E_K$.
Consider a molecule with $\numel$ electrons at positions $\rv_1, \ldots, \rv_\numel \in \mathbb{R}^3$ and $\numnu$ fixed nuclei of charges $Z_I$ at positions $\Rv_I$, $I \in [\numnu]$.
In the Born--Oppenheimer approximation and with $\xv = (\rv_1, \ldots, \rv_\numel)$, the non-relativistic electronic Hamiltonian is
\[
\hmltn \defeq \sum_{i=1}^\numel \biggl\{ -\half \nabla_{\rv_i}^2 - \sum_{I=1}^\numnu \frac{Z_I}{\|\rv_i - \Rv_I\|} \biggr\} + \sum_{1 \le i < j \le \numel} \frac{1}{\|\rv_i - \rv_j\|},
\]
and the time-independent Schr\"odinger equation reads $\hmltn \psi = E \psi$.
Each electron also has a spin, up or down.
We fix the numbers $n_\uparrow$ and $n_\downarrow$ of spin-up and spin-down electrons and assign the spins to the electrons, so that a wavefunction is a function of $\xv$ alone that must be antisymmetric under exchange of two electrons with the same spin (Appendix~\ref{app:sec:spin_content}).
The neural-network architecture enforces this antisymmetry (Appendix~\ref{app:sec:architectures}).
The wavefunctions therefore live in the space $\Hc$ of complex-valued functions in $L^2(\mathbb{R}^{3\numel})$ with this antisymmetry.
As a closed subspace of $L^2(\mathbb{R}^{3\numel})$, $\Hc$ is a Hilbert space with the standard inner product $\ip{f}{g} \defeq \int f^*(\xv)g(\xv)\,d\xv$.
Since $\hmltn$ commutes with permutations of same-spin electrons, it maps $\Hc$ into itself.
It is moreover self-adjoint on $\Hc$, a property that requires its adjoint to have the same domain and that the Kato--Rellich theorem establishes despite the Coulomb singularities (Appendix~\ref{app:sec:hamiltonian_structure}).
We assume that $E_1,\ldots,E_K$ lie strictly below the essential spectrum of $\hmltn$, which begins at the first ionization limit (Appendix~\ref{app:sec:spectral_hypothesis}).
Since $\hmltn$ has real coefficients in position space, its eigenstates can be chosen real-valued, and we restrict every ansatz to a real-valued function.\footnote{This relies on the absence of external magnetic fields and of relativistic effects such as spin-orbit coupling, both outside the present scope.}

\subsection{Variational Monte Carlo for ground state computation}
VMC minimizes the Rayleigh quotient of a wavefunction ansatz $\psi_\th$ with parameters $\th$,
\begin{align*}
\min_{\th} \Lc(\th),\quad \Lc(\th) \defeq \frac{\ip{\psi_\th}{\hmltn\psi_\th}}{\ip{\psi_\th}{\psi_\th}}=\E_{p_\th(\xv)}\bigl[\Ec_\th(\xv)\bigr],
\end{align*}
where $\Ec_\th(\xv)\defeq \frac{\hmltn\psi_\th(\xv)}{\psi_\th(\xv)}$ is called the \emph{local energy}, and $p_\th(\xv)\defeq \frac{|\psi_\th(\xv)|^2}{\ip{\psi_\th}{\psi_\th}}$ is the induced probability density from the wavefunction ansatz, called the \emph{Born density}.
Then the gradient of the Rayleigh quotient is
\begin{align}
\label{eq:vmc_grad}
\nabla_\th \Lc(\th)
&= 2 \E_{p_\th(\xv)}\Bigl[\Re\Bigl((\Ec_\th(\xv)-\E_{p_\th(\xv)}[\Ec_\th(\xv)]) \nabla_\th \log \psi_\th^*(\xv) \Bigr)\Bigr].
\end{align}
See Lemma~\ref{lem:grad_basic} for its proof.
The gradient~\eqref{eq:vmc_grad} can be estimated unbiasedly from \emph{walkers}, samples drawn from the Born density $p_\th$ by a Markov chain Monte Carlo (MCMC) algorithm~\citep{Foulkes--Mitas--Needs--Rajagopal2001}.
Since integrals over $\mathbb{R}^{3\numel}$ are otherwise intractable, VMC needs an objective, or at least its gradient, written as expectations under the Born densities of the ansatzes; we call such an objective or gradient \emph{VMC-compatible}.

\subsection{Prior works on excited states computation}
\label{sec:prelim_excited_prior_works}
Two VMC-compatible approaches exist: penalty-based methods tune penalty weights against the energy gaps, and NES-VMC needs $K^2$ Laplacians per step.

\textbf{(1) Penalty-Based Methods.}
Penalty-based methods
\citep{Entwistle--Schatzle--Erdman--Hermann--Noe2023,Pathak--Busemeyer--Rodrigues--Wagner2021,Wheeler--Kleiner--Wagner2024,Szabo--Schatzle--Entwistle--Noe2024}
keep the per-state Rayleigh quotient and add an explicit pairwise overlap penalty to enforce orthogonality:
\begin{align*}
\Lc_{\mathsf{penalty}}(\psi_{1:K})
&= \sum_{\kidx=1}^K 
\E_{p_\kidx}[\Ec_\kidx] 
+ \sum_{k<\ell} \a_{k\ell}\,\Pi(\tilde{\Sm}_{k\ell}),
\quad \tilde{\Sm}_{k\ell}\defeq \frac{\ip{\psi_{k}}{\psi_{\ell}}}{\|\psi_{k}\| \|\psi_{\ell}\|},
\end{align*}
where $\Ec_\kidx(\xv)\defeq \frac{\hmltn\psi_\kidx(\xv)}{\psi_\kidx(\xv)}$ denotes the local energy of the $\kidx$-th ansatz and $p_\kidx(\xv)\propto|\psi_{\kidx}(\xv)|^2$ the corresponding Born density, $\a_{k\ell}>0$ are penalty weights, and $\Pi$ is a penalty kernel, such as $\Pi(s)=|s|^2$, which admits an unbiased estimator~\citep{Pathak--Busemeyer--Rodrigues--Wagner2021,Wheeler--Kleiner--Wagner2024}.
\citet{Wheeler--Kleiner--Wagner2024} established an ensemble variational principle showing that an energy-weighted version of this objective recovers the ordered lowest $K$ eigenfunctions of $\hmltn$ once the overlap penalty exceeds a spectral-gap-dependent threshold (Appendix~\ref{app:sec:penalty_explicit}).
This threshold depends on the energy gaps, so the tuning depends on the very spectrum the optimization aims to compute.
Based on this criterion, \citet{Szabo--Schatzle--Entwistle--Noe2024} set each $\a_{k\ell}$ at every step to a global scale $\tilde\alpha$ times the larger of two exponential moving averages (EMAs) over training steps: that of the energy gap between states $k$ and $\ell$ and that of the local-energy standard deviation of state $k$ (Appendix~\ref{app:sec:penalty_explicit}).
Whether the global minimum gives the lowest $K$ eigenstates thus depends on how accurate these estimates are.

\textbf{(2) NES-VMC.}
\citet{Pfau--Axelrod--Sutterud--von-Glehn--Spencer2024} propose an extended-system construction for the lowest $K$ states.
NES-VMC evaluates the $K$ ansatzes $\psi_1,\ldots,\psi_K$ on $K$ electron configurations $\Xv = (\xv^1,\ldots,\xv^K)$ and forms the matrix ansatz $\Psi(\Xv)\in\Complex^{K\times K}$ with $\Psi_{ik}(\Xv)=\psi_k(\xv^i)$.
The total ansatz is the determinant $\Phi(\Xv)\defeq\det\Psi(\Xv)$.
NES-VMC minimizes the energy of this extended system, in which $\hmltn$ acts on each of the $K$ configurations $\xv^i$ and the results are summed.
This energy has two forms (Theorems~\ref{thm:nes_rq_basic} and~\ref{thm:nes_rq_vmc} in Appendix~\ref{app:sec:nes_statements})
\begin{align}
\label{eq:nes_two_forms}
\Lc_{\mathsf{NES\text{-}VMC}}(\psi_{1:K})
= \tr(\Sm^{-1}\Hm)
= \E_{\Xv\sim|\Phi|^2}\bigl[\tr(\locmat(\Xv))\bigr],
\end{align}
where $\Sm\defeq[\ip{\psi_k}{\psi_\ell}]_{k,\ell=1}^K$ and $\Hm\defeq[\ip{\psi_k}{\hmltn\psi_\ell}]_{k,\ell=1}^K$ are the overlap and energy matrices, and $\locmat(\Xv)\defeq\Psi(\Xv)^{-1}(\hmltn\Psi)(\Xv)$ is the \emph{total local energy matrix} with $(\hmltn\Psi)(\Xv) \defeq [\,(\hmltn\psi_k)(\xv^i)\,]_{i,k=1}^K$.
The matrix $\Sm^{-1}\Hm$ represents $\hmltn$ projected onto $\mathrm{span}\{\psi_{1:K}\}$, so the middle form is the sum of the eigenvalues of this projected operator, minimized with value $E_1+\cdots+E_K$ by any $K$ ansatzes whose span is that of $\varphi_1,\ldots,\varphi_K$.
The right form of Eq.~\eqref{eq:nes_two_forms} is the ground-state VMC form on the $K$-fold configuration space under the Born density $|\Phi|^2/\ip{\Phi}{\Phi}$ of the total ansatz, written $\Xv\sim|\Phi|^2$.
The objective and its gradient are thus expectations under this density, so the objective is VMC-compatible.
Since the minimum does not single out eigenfunctions within this span, NES-VMC orders and orthogonalizes the states in a post-processing step.
This step diagonalizes the sample mean of $\locmat$, an estimate of $\Sm^{-1}\Hm$ (Proposition~\ref{prop:nes_locmat_mean}), and reads the individual energies from the eigenvalues of this mean.
With this procedure, NES-VMC recovers accurate vertical excitation energies, including double excitations, on molecules up to the size of benzene.

NES-VMC incurs three structural costs, all from the total ansatz $\Phi=\det\Psi$ through which it estimates~\eqref{eq:nes_two_forms}.
First, $\locmat$ contains a cross term $(\hmltn\psi_k)(\xv^i)$ for every pair $(i,k)$, and these $K^2$ Laplacians per step dominate the wall time (Table~\ref{tab:positioning}).
Second, the Born density of $\Phi$ couples all $K$ states in a single Markov chain, whereas penalty-based methods and \algo sample each state in its own chain, in parallel.
Lastly, the gradient flows through the matrix inverse $\Psi(\Xv)^{-1}$, which can become ill-conditioned when two ansatzes are nearly linearly dependent.
We elaborate each point in Appendix~\ref{app:sec:nes_costs} and survey the numerical-stability measures of \citet{Pfau--Axelrod--Sutterud--von-Glehn--Spencer2024} in Appendix~\ref{app:sec:nan_handling}.
Appendix~\ref{app:related-work} presents more related work on excited states and on learning eigenfunctions with neural networks.

\section{Proposed method: \algo}
\label{sec:proposed_method}

\begin{table}[t]
\centering
\footnotesize
\setlength{\tabcolsep}{3pt}
\caption{Comparison of \algo with the two existing excited-state VMC approaches. Theoretical scaling excludes spin-penalty overhead (Appendix~\ref{app:sec:nes_costs}). $^\dagger$Measured by \citet[Figure~1]{Szabo--Schatzle--Entwistle--Noe2024} for $K\in\{1,\ldots,30\}$ on Ne.}\label{tab:positioning}
\begin{tabular*}{\textwidth}{@{\extracolsep{\fill}}lccc@{}}
\toprule
 & \textbf{NES-VMC}~\citep{Pfau--Axelrod--Sutterud--von-Glehn--Spencer2024}
 & \textbf{Penalty-based}~\citep{Entwistle--Schatzle--Erdman--Hermann--Noe2023,Szabo--Schatzle--Entwistle--Noe2024}
 & \textbf{\algo} \\
\midrule
\textbf{Objective}
 & Energy of the extended system
 & Energies + overlap penalty
 & Rank-$K$ approximation of $-\hmltn$ \\[2pt]
\textbf{Overlap-penalty weights}
 & -
 & $\a_{k\ell} > |E_k - E_\ell|$, per pair
 & - \\[2pt]
\textbf{Sampling}
 & One chain over $K\numel$ electrons
 & $K$ chains over $\numel$ electrons
 & $K$ chains over $\numel$ electrons \\[2pt]
\textbf{Laplacians per step}
 & $K^2$
 & $K$
 & $K$ \\[2pt]
\textbf{Theoretical scaling}
 & $O(K^2 \numel^4 + K^3)$
 & $O(K \numel^4 + K^2 \numel^3)$
 & $O(K \numel^4 + K^2 \numel^3)$ \\[2pt]
\textbf{Wall-time scaling on Ne}
 & $O(K^{2.5})^\dagger$
 & $O(K^{1.1})^\dagger$
 & $\approx$ penalty-based (Figure~\ref{fig:scaling}) \\[2pt]
\textbf{Output}
 & Subspace, diagonalized post hoc
 & Ordered states
 & Ordered states \\
\bottomrule
\end{tabular*}
\end{table}

This section adapts the NestedLoRA framework, developed for general operator decomposition, to excited-state VMC.
The resulting algorithm, \algo, is penalty-free and needs $K$ Laplacians per step.
It inherits a global optimality guarantee from the LoRA principle and learns the ordered eigenfunctions without explicit diagonalization.
Sections~\ref{sec:lora_background} and~\ref{sec:nestedlora_vmc} build the objective of \algo and its guarantee, and Sections~\ref{sec:implementation} and~\ref{sec:scaling} estimate it by Monte Carlo and compare its per-step cost with the two existing approaches.

\subsection{Operator spectral decomposition via low-rank approximation}
\label{sec:lora_background}
For a self-adjoint operator $\Top$ that is bounded above, suppose that we wish to recover its top-$K$ eigenvalues $\mu_1\ge \mu_2\ge \ldots \ge \mu_K$ and corresponding orthonormal eigenfunctions $|\varphi_1\rangle,\ldots,|\varphi_K\rangle$.
In our case, we will set $\Top=-\hmltn$ such that $\mu_k=-E_k$.
Recently, \citet{Ryu--Xu--Erol--Bu--Zheng--Wornell2024} proposed the low-rank approximation (LoRA) objective
\begin{align}
\Lc_{\lora}(f_{1:K})&\defeq
-2\sum_{\kidx=1}^K {\ip{f_\kidx}{\Top f_\kidx}}
+\sum_{\kidx=1}^K \sum_{\ell=1}^K 
{\ip{f_\kidx}{f_{\ell}}\ip{f_{\ell}}{f_{\kidx}}},
\label{eq:lora_obj}
\end{align}
as a training objective for parametric spectral decomposition.
For a Hilbert--Schmidt operator $\Top$, this objective is equivalent to the rank-$K$ approximation error in Hilbert--Schmidt norm $\|\Top-\sum_{k=1}^K |f_k\rangle\langle f_k|\|_{\mathsf{HS}}^2$.
If this $\Top$ is moreover positive semidefinite and $\mu_K>\mu_{K+1}$, a global minimizer $f_{1:K}^\star$ of $\Lc_{\lora}(f_{1:K})$ recovers the top-$K$ eigenspace of $\Top$ as $\sum_{k=1}^K |f_k^\star\rangle\langle f_k^\star|=\sum_{k=1}^K\mu_k|\varphi_k\rangle\langle\varphi_k|$; for a self-adjoint $\Top$ with an orthonormal basis of eigenfunctions, not necessarily compact or positive semidefinite, the minimizer recovers the top-$\min(K,r)$ eigenspace, where $r$ is the number of positive eigenvalues~\citep[Theorems~3.1 and~C.5]{Ryu--Xu--Erol--Bu--Zheng--Wornell2024}.

While neither result covers our quantum chemistry case $\Top=-\hmltn$, which is unbounded and has a continuous spectrum of scattering states, we show that the LoRA principle is still applicable when the lowest $K$ energies are eigenvalues below its essential spectrum (Appendix~\ref{app:sec:spectral_hypothesis}).
That is, a global minimizer recovers the top-$K$ eigenspace as above whenever $\mu_K > \max(\mu_{K+1},\sup\sigma_{\mathrm{ess}}(\Top),0)$, with $\sigma_{\mathrm{ess}}(\Top)$ the essential spectrum; the proof applies the Eckart--Young theorem to $\Top$ projected onto the span of $f_{1:K}$, the only part of $\Top$ on which the LoRA objective depends (Appendix~\ref{sec:lora_detailed_discussion}).

\subsection{NestedLoRA-VMC}
\label{sec:nestedlora_vmc}
The LoRA objective requires a reparameterization to be \textit{VMC-compatible}.
\citet{Ryu--Xu--Erol--Bu--Zheng--Wornell2024} further propose an optimization technique called \emph{nesting} to learn ordered eigenfunctions without explicit diagonalization, and call the framework \emph{NestedLoRA}.

\paragraph{Reparameterization.} The key idea for VMC-compatibility is to reparameterize each function $f_{k}$ in \eqref{eq:lora_obj} as $\sqrt{\eigvalparam_k}\,\psi_k / \|\psi_k\|$, with the ansatzes $\psi_{1:K}$ parameterized as neural networks with parameters $\th$ and $\eigvalparam_{1:K} \in \Real_{\ge 0}^K$ as additional trainable scale parameters.
Under the sequential nesting introduced below, each $\eigvalparam_k$ plays the role of the $k$-th eigenvalue estimate.
This yields the objective
\begin{align*}
\Jc_{\lora}(\psi_{1:K},\eigvalparam_{1:K})
\defeq -2\sum_{k=1}^K \eigvalparam_k\Rm_{kk}^{\Top}
+ \sum_{k,\ell=1}^K \eigvalparam_k\eigvalparam_\ell \Rm_{k\ell}^{\Iop}\Rm_{\ell k}^{\Iop},
\end{align*}
where we define the generic functional $\Rm_{k\ell}^{\Oop} \defeq \frac{\ip{\psi_k}{\Oop \psi_\ell}}{\ip{\psi_k}{\psi_k}} = \E_{p_k}[\frac{\Oop  \psi_\ell}{\psi_k}]$ for a symmetric operator $\Oop\in\{\Top,\idop\}$.
The objective is VMC-compatible: drawing electronic configurations $\Xv$ from the product distribution $|\psi_1(\xv^1)|^2\cdots |\psi_K(\xv^K)|^2$ yields unbiased estimates of the objective and its gradient $\nabla_\theta \Jc_{\lora}$ (Theorem~\ref{thm:lora_grad}). 

\paragraph{Sequential nesting for learning ordered excited states.}
\label{sec:nesting}
As in \citet{Ryu--Zhou--Wornell2025}, we order the learned eigenfunctions by sequential nesting.
Sequential nesting updates each network $k$ by the gradient with respect to $\psi_k$ of the order-$k$ objective $\Jc_{\lora}(\psi_{1:k},\eigvalparam_{1:k})$, the LoRA objective of the first $k$ states, and updates all $K$ networks at once.
This update can be read as $K$ nested problems solved together: if states $1,\ldots,k-1$ are the first $k-1$ eigenpairs, the order-$k$ objective is minimized by $\psi_k\propto\varphi_k$ and $\lambda_k=\mu_k$, so the fixed point of the update is the ordered eigenpairs (Appendix~\ref{app:sec:nesting_fixed_point}).

We define the sequential nesting gradient as
\begin{align}
\label{eq:nested_gradient}
\nabla_\th^\seq
&\defeq \frac14\sum_{k=1}^K \ip{\nabla_\th \psi_k}{\partial_{\psi_k}\Jc_{\lora}(\psi_{1:k},\eigvalparam_{1:k})} 
= \sum_{k=1}^K \eigvalparam_k\Re\Bigl\{ - \tilde{\Delta}_{kk}^{\Top} + \sum_{\ell=1}^K \matmask_{k\ell}^\seq\eigvalparam_\ell\Rm_{\ell k}^{\idop}\tilde{\Delta}_{k\ell}^{\idop} \Bigr\},
\end{align}
where $\partial_{\psi_k}$ is the functional derivative with respect to $\psi_k$, $\matmask_{k\ell}^\seq = \ones\{k>\ell\}$ is the strictly lower-triangular mask and $\tilde{\Delta}_{k\ell}^{\Oop} \defeq \Cov_{p_k}(\Oop\psi_\ell / \psi_k, \nabla_\th \log \psi_k^*)$ is the covariance, under $p_k$, of the operator action $\Oop\psi_\ell/\psi_k$ against the per-state score $\nabla_\th \log \psi_k^*$ (using score in the same log-derivative sense as in eq.~\eqref{eq:vmc_grad}), for $\Oop \in \{\Top, \idop\}$.
Appendices~\ref{app:sec:unified_gradient_derivation} and~\ref{app:sec:proof_sequential_nesting} derive Eq.~\eqref{eq:nested_gradient}.
For real-valued ansatzes the gradient is VMC-compatible, a sum over states of expectations under the Born densities $p_k$, estimated by averages over samples $\xv\sim p_k$, for each $k\in[K]$:
\begin{align}
\nabla_\th^\seq &= \sum_{k=1}^K \E_{p_k(\xv)}\!\Bigl[(g_{k}(\xv) - \E_{p_k}[g_{k}])\,\nabla_\th \log |\psi_k|\Bigr],
\nonumber
\\
\label{eq:def_g}
g_{k}(\xv) &\defeq \lambda_k\Bigg(-\,\frac{\Top \psi_k(\xv)}{\psi_k(\xv)} + \sum_{\ell=1}^K \matmask_{k\ell}^\seq\,\lambda_\ell\,\Rm_{\ell k}^{\idop}\,\frac{\psi_\ell(\xv)}{\psi_k(\xv)}\Bigg).
\end{align}

In our implementation, we drop the prefactor $\lambda_k$ in $g_k(\xv)$ to normalize the gradient norm, which can be equivalently understood as using the state-dependent learning rate $\eta/\lambda_k$ for a base learning rate $\eta$; this rescales the gradient of each state by a positive factor and keeps the fixed point.
A similar state-dependent learning rate was recently studied for $k$-SVD by gradient descent with the LoRA principle~\citep{Jedra--Shah2025}.

\subsection{Implementation details}
\label{sec:implementation}

In Sections~\ref{sec:lora_background} and~\ref{sec:nestedlora_vmc}, we derive a parametric gradient for finding the ordered eigenpairs as a fixed point out of iterative gradient-based updates. 
In practical optimization, there exist several issues that hinder effective training with a direct implementation of the gradient above: heavy-tails of the local energy and overlap statistics, the step-size sensitivity of the gradient-based eigenvalue updates, the conditioning of each network's objective, and unstable order of the states at the beginning of training.
The first is a statistical issue, while the rest are related to optimization.
In this section, we provide a solution to each problem without altering the fixed-point property of the desired solution.

\subsubsection{Asymmetric clipping for local energy and overlap statistics}
\label{sec:clipping}
To implement the gradient in Eq.~\eqref{eq:def_g}, we need the local energy statistics $\Top\psi_k/\psi_k$ and local overlap statistics $\psi_\ell/\psi_k$. 
Since both statistics may blow up around the nodes of $\psi_k$ (i.e., where $\psi_k\approx 0$), it is a common practice in the literature~\citep{Pfau--Spencer--Matthews--Foulkes2020,glehn2023a,Pfau--Axelrod--Sutterud--von-Glehn--Spencer2024,Szabo--Schatzle--Entwistle--Noe2024} to apply a clipping on these values.
However, we empirically find that an asymmetric clipping strategy, clipping local energies while using raw local overlap statistics, leads to much more stable training; we attribute the effectiveness of the asymmetric clipping rule to differences in how the statistics behave near convergence, as explained below.

First, note that the local energy $\Top\psi_k/\psi_k$ is constant at the fixed point, so the clipping is inactive there; Appendix~\ref{app:sec:energy_statistics} gives the clipping rule.
In contrast, the local overlap $\psi_\ell/\psi_k$ stays heavy-tailed at the fixed point; theoretically, we can show that, under the Born density $p_k$, the probability that its magnitude exceeds $t$ decays as $t^{-3}$ under regularity conditions; see Appendix~\ref{app:sec:overlap_heavy_tail}.
This suggests that, for example, the clipping rule of \citet{Szabo--Schatzle--Entwistle--Noe2024}, which clips outside the median $\pm 10$ median absolute deviations, may still unexpectedly perturb some walkers even around convergence, since the heavy tail places some walkers outside the clip window; see Figure~\ref{fig:overlap_clip_bias}c in Appendix~\ref{app:sec:overlap_measurements}.
The perturbation can bias the batch mean of the ratio, whose expectation $\Rm^{\idop}_{k\ell}$ vanishes at the fixed point.

We demonstrate empirically on Li that clipping the ratio substantially biases its batch mean, both early in training and near convergence.
Early in training, when the overlap remains nonzero and the overlap term helps separate the networks, clipping shifts the batch mean away from a reference estimate obtained using long Metropolis chains targeting a mixture of $p_k$ and $p_\ell$, by an amount many times larger than its variation across batches; see Figure~\ref{fig:overlap_clip_bias}.
At the final training step, near a fixed point at which the overlap should vanish, the clip still changes the walkers of many ratios and can leave their batch means spuriously nonzero; see Appendix~\ref{app:sec:overlap_measurements}.

\subsubsection{Analytical eigenvalue updates}
\label{sec:analytic_eigvals_subsec}
A gradient update of the eigenvalue parameters $\lambda_{1:K}$ together with the networks needs a step size for $\lambda$, which may be hard to tune together with training the neural ansatzes.
To simplify the updates, we instead propose to track the eigenvalues by an analytical update at each step.
Since the objective $\Jc_{\lora}(\psi_{1:k},\eigvalparam_{1:k})$ is strictly convex quadratic in $\eigvalparam_k$, we solve its stationary equation $\partial_{\eigvalparam_k}\Jc_{\lora}=0$ to obtain an analytical update.
Mirroring the sequential update of $\psi_k$, we hold the lower-state parameters $\{\eigvalparam_j\colon j < k\}$ fixed and solve $\partial_{\eigvalparam_k}\Jc_{\lora} = 0$ sequentially for $k = 1, \ldots, K$, yielding
\begin{align}
\label{eq:analytic_eigvals}
\boldsymbol{\eigvalparam}^\star
&= \bigl(\Imat + \matmask^{\seq}\odot(\Rm^{\idop}\odot(\Rm^{\idop})^\intercal)\bigr)^{-1}\diag(\Rm^{\Top}),
\end{align}
which can be computed in one cheap $O(K^2)$ triangular sweep per step; Appendix~\ref{app:sec:analytic_eigvals} gives the step-by-step derivation. 
Because the system is lower-triangular, a heavy-tailed estimate at state $m$ enters $\eigvalparam^\star_j$ only for $j \ge m$.
In practice, we plug EMAs of the estimates of $\Rm^{\idop}$ and $\diag(\Rm^{\Top})$ into Eq.~\eqref{eq:analytic_eigvals} to smooth the trajectory across steps.
With the exact $\Rm^{\idop}$ and $\diag(\Rm^{\Top})$ in place of these estimates, the solve returns $\lambda_k=\mu_k$ at the ordered eigenpairs, so the fixed point is unchanged.

\subsubsection{Per-network operator shifts for curvature shaping}
\label{sec:shifted_operator}

Another source of instability in the sequential nesting gradient of Eq.~\eqref{eq:def_g} is the imbalance between its energy and overlap terms.
While the energy term distinguishes eigenstates through spectral gaps, the overlap term is weighted by $\lambda_\ell\approx -E_\ell$, whose magnitude can be much larger than the spectral gaps.
This creates steep directions toward lower eigenstates and relatively flat directions toward higher ones, leading to unstable convergence.

To mitigate this imbalance, we introduce a network-specific operator shift.
Since network $k$ is updated only through the $k$-th term of Eq.~\eqref{eq:nested_gradient}, it can use $\Top_k\defeq E_{\mathrm{ref},k}\idop-\hmltn$, with $E_{\mathrm{ref},k}$ fixed during gradient computation.
The shift adds a constant $-E_{\mathrm{ref},k}$ to the local energy, which vanishes upon centering, while changing each lower-state overlap coefficient from $\lambda_\ell\approx-E_\ell$ to $\lambda_\ell+E_{\mathrm{ref},k}\approx E_{\mathrm{ref},k}-E_\ell$ near orthogonality.
Thus, choosing $E_{\mathrm{ref},k}$ near $E_k$ reduces the relative weight of the overlap term without changing the centered energy contribution.

To see its effect on curvature, assume $\psi_\ell\propto\varphi_\ell$ for $\ell<k$ and expand $q_k\defeq\frac{\psi_k}{\|\psi_k\|}=\sum_n a_{kn}\varphi_n+\chi_k$, where $\varphi_n$ are the discrete eigenfunctions of $\hmltn$ with energies $E_n$, and $\chi_k$ belongs to its essential spectral subspace.
The effective objective underlying the $k$-th gradient, with the prefactor $\lambda_k$ removed, is approximately $E_k + \sum_{\ell<k}(E_{\mathrm{ref},k}-E_k)\,a_{k\ell}^2 + \sum_{n>k}(E_n-E_k)\,a_{kn}^2 + \ip{\chi_k}{(\hmltn-E_k)\chi_k}$.
Thus, the lower-state curvature decreases from $-E_k$ to $E_{\mathrm{ref},k}-E_k$, while the higher-state curvature remains unchanged.

In our implementation, we choose $E_{\mathrm{ref},k}=-\lambda_k+c\,\sigma_k$, where $-\lambda_k$ estimates $E_k$ via Eq.~\eqref{eq:analytic_eigvals}, $\sigma_k$ is the walker standard deviation of the clipped local energy, and $c=5$.
The margin $c\,\sigma_k$ helps ensure $E_{\mathrm{ref},k}>E_k$, so that the target state retains a positive eigenvalue under $\Top_k$ and remains recoverable by LoRA.
Near convergence, the lower-state curvature becomes approximately $c\,\sigma_k$ rather than $|E_k|$.
Whenever $E_{\mathrm{ref},k}>E_k$, $\varphi_k$ remains a minimizer, up to degeneracies.
Appendix~\ref{app:sec:shifted_operator} analyzes the curvature imbalance for carbon, and Section~\ref{sec:exp_shifted_operator} demonstrates the effect of the shift.

\subsubsection{Online state ordering for faster convergence}
\label{sec:state_ordering}
We find that online state reordering lets the networks learn the lowest $K$ states without waiting for their energies to line up with their indices during training.
We assign, at every step, the order-$k$ objective to the network whose EMA energy ranks $k$-th within its spin sector (Section~\ref{sec:exp_setup} and Appendix~\ref{app:sec:ewm_ordering}).
The ranking reads only the EMA energies, so it is the same under every shift $E_{\mathrm{ref},k}$ of Section~\ref{sec:shifted_operator}.
The assignment changes only which network carries which index, so the fixed point is unchanged.

\subsection{Choice of parameterization and per-iteration efficiency}
\label{sec:scaling}

\algo needs $K$ Laplacians per step, against $K^2$ for NES-VMC.
Like the penalty methods, \algo applies $\hmltn$ to each state only at that state's own walkers, $(\hmltn\psi_k)(\xv^k)$, while the local-energy matrix of NES-VMC needs $(\hmltn\psi_k)(\xv^i)$ at every pair $(i,k)$.
A Laplacian is the most expensive evaluation in a step: it sums $3\numel$ second derivatives, so with the forward Laplacian~\citep{Li--Ye--Jiang--Wen--Wang--Li--Li--He--Chen--Ren--Wang2024} it costs about $3\numel$ forward evaluations, $O(\numel^4)$ against $O(\numel^3)$ for one forward evaluation (Appendix~\ref{app:sec:folx}).

FermiNet~\citep{Pfau--Spencer--Matthews--Foulkes2020} and the Psiformer~\citep{glehn2023a} compute a feature of each electron with a \emph{backbone} of permutation-equivariant layers, map these features to an \emph{orbital matrix}, and sum the determinants of such matrices.
Within one forward evaluation, the determinant of the orbital matrix sets the order $\numel^3$, and on first-row atoms most of the cost is in the dense layers of the backbone, which act on a $256$-dimensional feature of each electron.
These costs, times the number of forward evaluations per step that Appendix~\ref{app:sec:nes_costs} counts by component, give the theoretical scaling in Table~\ref{tab:positioning}.
The per-step wall time measured on Ne follows it for NES-VMC, the penalty class~\citep[Figure~1]{Szabo--Schatzle--Entwistle--Noe2024} and \algo (Appendix~\ref{app:sec:scaling}).

NES-VMC uses one \emph{joint} backbone with $K$ heads, each of which maps the shared features of the electrons to the orbital matrices of one state.
Whether this costs less than $K$ disjoint networks depends on where $\hmltn$ is applied.
In NES-VMC every state needs its Laplacian at every walker set, and a joint backbone computes the backbone's part once per walker set for all $K$ heads.
In \algo and the penalty methods each state needs its Laplacian only at its own walkers, so the Laplacian costs the same under either parameterization, and a joint backbone saves only on the overlaps, which carry no factor $3\numel$.
We use $K$ disjoint networks, following \citet{Szabo--Schatzle--Entwistle--Noe2024}.

\section{Experiments}
\label{sec:experiments}

We evaluate \algo on the first-row atoms, Li through Ne, with the Psiformer architecture~\citep{glehn2023a}, for the lowest $K{=}10$ eigenstates of $\hmltn$.
Its excited-state energies are comparable in accuracy to those of NES-VMC~\citep{Pfau--Axelrod--Sutterud--von-Glehn--Spencer2024} and the penalty-based method of \citet{Szabo--Schatzle--Entwistle--Noe2024}, against term centroids from the National Institute of Standards and Technology (NIST) (Section~\ref{sec:exp_results}).
On carbon, the per-network operator-shift of Section~\ref{sec:shifted_operator} brings the five networks of the degenerate $2p^2\,{}^1\!D$ level close together in energy, whereas with $\Top=-\hmltn$ they stay spread over the level (Section~\ref{sec:exp_shifted_operator}).\footnote{Source code for \algo is available at \url{https://github.com/MinchanJeong/NestedLoRA-VMC}.}

\begin{figure}[t]
  \centering
  \includegraphics[width=\linewidth]{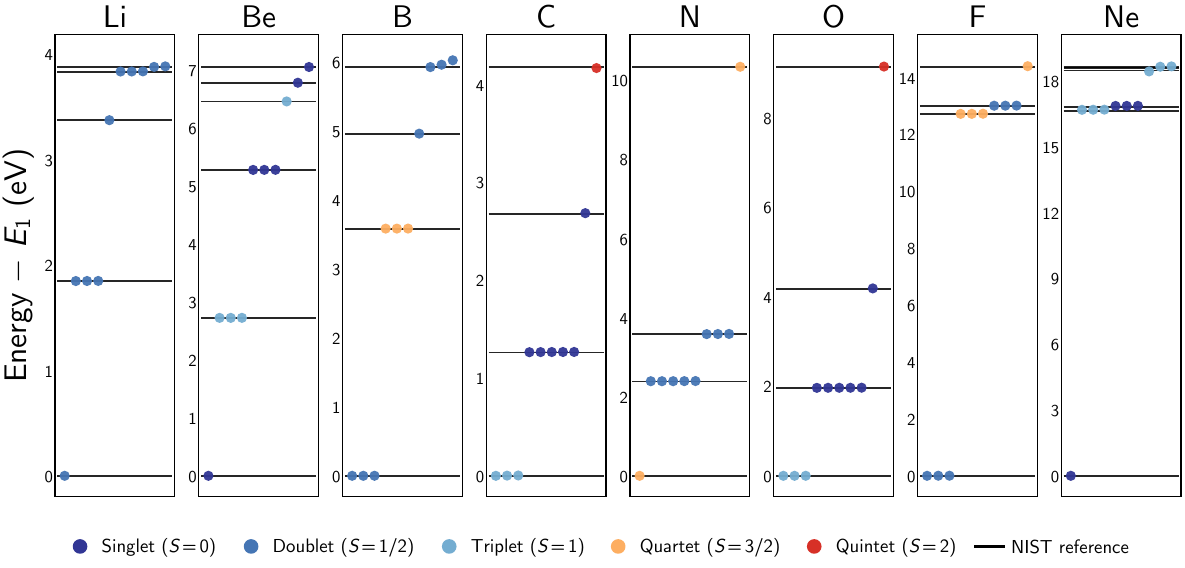}
  \caption{First-row $K{=}10$ energy-level diagram (Li--Ne). Per-atom
  recovered spectra (\algo) on top of NIST term
  centroids~\citep{Sansonetti--Martin2005} (horizontal segments).
  Markers are colored by total spin $S$ (per-network measured
  $\langle \hat S^2 \rangle$). Within each subplot, markers are placed
  left-to-right in order of energy, so states close in energy separate
  horizontally. For an atom with several seeds in
  Table~\ref{tab:atoms_mae}, each marker is the mean over seeds of the
  energy and $\langle \hat S^2 \rangle$ of the state at that rank.
  Table~\ref{tab:per_state_master} gives the term and
  configuration of each state.}
  \label{fig:atom-spectra}
\end{figure}

\subsection{Setup}
\label{sec:exp_setup}

\paragraph{Ansatz and MCMC sampler.}
Every network is a Psiformer~\citep{glehn2023a} and samples its own Born density with one Metropolis chain; Appendix~\ref{app:sec:samplers} gives the sampler settings.

\paragraph{Pretraining.}
For every atom we pretrain to a complete-active-space self-consistent-field (CASSCF) target with \texttt{aug-cc-pVTZ} as the underlying single-particle basis.
The active space $(N_e, N_o)$, $N_e$ electrons in $N_o$ orbitals, is chosen per atom to cover the $K{=}10$ states of its sectors (Table~\ref{tab:atom_setup}).
The per-state CASSCF targets and the pretraining schedule are in Appendix~\ref{app:sec:pretrain_details}.

\paragraph{Spin sectors and spin penalty.}
To obtain pure spin states more accurately, we run \algo on $\Top=-(\hmltn+\hat S^2)$ so that each state has one total spin $S$.
The numbers $n_\uparrow$ and $n_\downarrow$ of Section~\ref{sec:prelim} fix $m_S=(n_\uparrow-n_\downarrow)/2$, so the \emph{spin sector} of a given $m_S$ holds every state with $S\ge|m_S|$ (cf.~\citet{Pfau--Spencer--Matthews--Foulkes2020,glehn2023a}).
Added to $\hmltn$, the spin penalty $\hat S^2$ raises each state by $S(S{+}1)$ and keeps the eigenfunctions, so the lowest states of a sector take its smallest spin $S=|m_S|$ (Appendix~\ref{app:sec:lora_remarks}).
In training, each atom has one sector $m_S=S$ for each spin multiplicity $2S{+}1$ among its lowest $K{=}10$ states; the sectors, with fixed numbers of networks that sum to $K{=}10$, are trained in one run, and their states are merged by energy.
\citet{Szabo--Schatzle--Entwistle--Noe2024} use this penalty for every atom and fix $m_S$ for each higher spin multiplicity, which gives the same sectors (Appendix~\ref{app:sec:atom_setup}).

\paragraph{Reference data.}
A term $^{2S+1}\!L$ has total spin $S$ and total orbital angular momentum $L$, with $L=0,1,2$ written S, P, D; Table~\ref{tab:atom_setup} gives the terms of each atom.
We compare energies with term centroids from the NIST atomic-spectroscopy compilation of \citet{Sansonetti--Martin2005}.
The centroid of a term is the mean energy of its fine-structure levels $J=|L-S|,\ldots,L+S$, each weighted by its degeneracy $2J{+}1$; spin--orbit coupling splits a term into these levels, and $\hmltn$ gives them one energy.
NIST gives each level as its energy above the ground level, so every comparison uses the excitation energy $\Delta E$ of each state above the ground-term centroid.

\paragraph{Training.}
Training runs are $100{,}000$ steps at an electron batch of $4{,}096$ with the Kronecker-factored approximate curvature (KFAC) optimizer~\citep{pmlr-v37-martens15}, a natural-gradient approximation applied per state (Appendix~\ref{app:sec:ngd_kfac}).

\subsection{Excited-state energies on first-row atoms}
\label{sec:exp_results}

The per-atom mean absolute error (MAE) of \algo in Table~\ref{tab:atoms_mae} is below chemical accuracy on every atom from Li to F and comparable to that of NES-VMC~\citep{Pfau--Axelrod--Sutterud--von-Glehn--Spencer2024} and of the penalty-based method of \citet{Szabo--Schatzle--Entwistle--Noe2024}.
NES-VMC reaches its MAE with $200{,}000$ training steps per atom~\citep[Table~S1]{Pfau--Axelrod--Sutterud--von-Glehn--Spencer2024}, twice the $100{,}000$ of \algo and of \citet[SI Table~S6]{Szabo--Schatzle--Entwistle--Noe2024}.

\begin{table}[ht]
\centering
\footnotesize
\setlength{\tabcolsep}{4pt}
\caption{Per-atom mean absolute error (mHa) of $K{=}10$ excited-state energies against $(2J{+}1)$-weighted NIST term centroids~\citep{Sansonetti--Martin2005}. An entry $m_{\pm s}$ is the mean $m$ and standard deviation $s$ over random seeds; the others are from one run.}
\label{tab:atoms_mae}
\begin{tabular}{lcccccccc}
\toprule
Method & Li & Be & B & C & N & O & F & Ne \\
\midrule
NES-VMC~\citep{Pfau--Axelrod--Sutterud--von-Glehn--Spencer2024} & 0.02 & 0.04 & 0.54 & 0.09 & 0.16 & 0.26 & 1.90 & 1.95 \\
\citet{Szabo--Schatzle--Entwistle--Noe2024} (Penalty-based)     & 1.06 & 0.20 & 0.63 & 0.59 & 2.95 & 1.43 & 1.40 & 2.86 \\
\midrule
\rowcolor{blue!15}
\textbf{\algo} & {$0.06_{\pm0.01}$} & {$0.03_{\pm0.01}$} & {$0.64_{\pm0.23}$} & {$0.18_{\pm0.01}$} & {0.28} & {0.27} & {0.65} & {$1.55_{\pm0.56}$} \\
\bottomrule
\end{tabular}
\end{table}

Figure~\ref{fig:atom-spectra} shows the recovered $K{=}10$ spectra on each atom.
All three methods are scored against the same NIST term centroids, with the NES-VMC and penalty errors computed from the per-state values published in their papers.
Per-state energies of \algo are the mean of the per-walker local energy over the last $3{,}000$ training steps, evaluated with exact interparticle distances (Appendix~\ref{app:sec:per_state_mae}).
The compute time available set the number of training seeds per atom: three for Li, Be and B, two for C and Ne, and one for N, O and F.
Appendix~\ref{app:sec:per_state_mae} gives the scoring, the per-state errors of all eight atoms, and in Table~\ref{tab:mae_intervals} the $95\%$ interval of each run's MAE.

\begin{figure}[t]
  \centering
  \includegraphics[width=\linewidth]{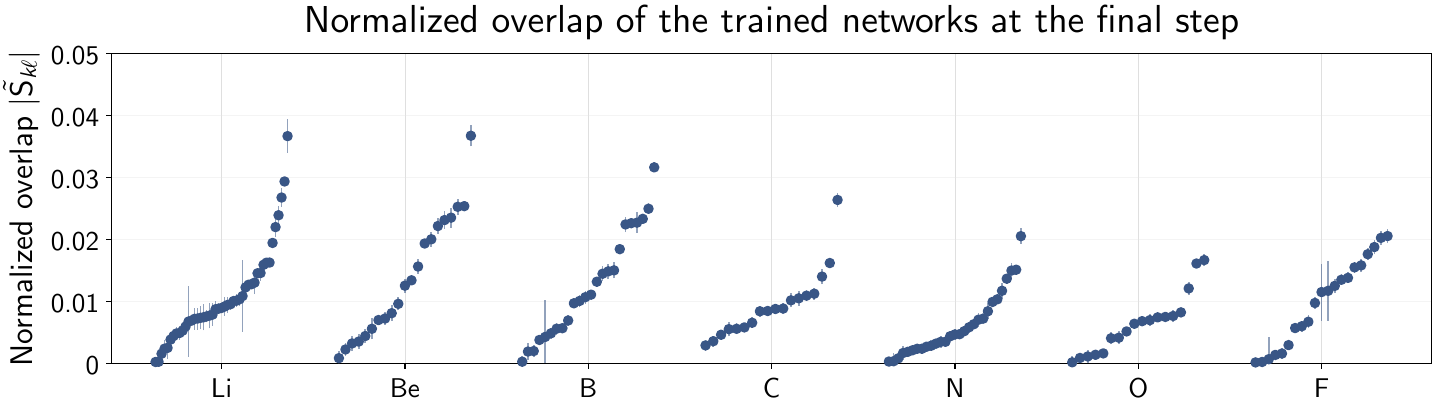}
  \caption{Normalized overlap $|\tilde{\Sm}_{k\ell}|$ of every pair of networks within a spin sector at the final training step of the seed-$1$ run of each atom from Li to F, with the pairs of an atom ordered by $|\tilde{\Sm}_{k\ell}|$. Each value comes from two long Metropolis chains (Appendix~\ref{app:sec:overlap_measurements}), and Table~\ref{tab:overlap_x2_classes} gives their convergence. Error bars are bootstrap standard errors (Appendix~\ref{app:sec:overlap_measurements}).}
  \label{fig:overlap_x2}
\end{figure}

\subsection{Overlap of the trained networks}
\label{sec:exp_overlap}

Exact eigenstates are mutually orthogonal, so we measure the normalized overlap of every pair of networks within a spin sector at the final training step as a surrogate for how well the networks learn the eigenstates, with the long chains of Appendix~\ref{app:sec:overlap_measurements} (Figure~\ref{fig:overlap_x2}).
The trained networks are close to orthogonal on every atom from Li to F.

\begin{figure}[t]
  \centering
  \includegraphics[width=\linewidth]{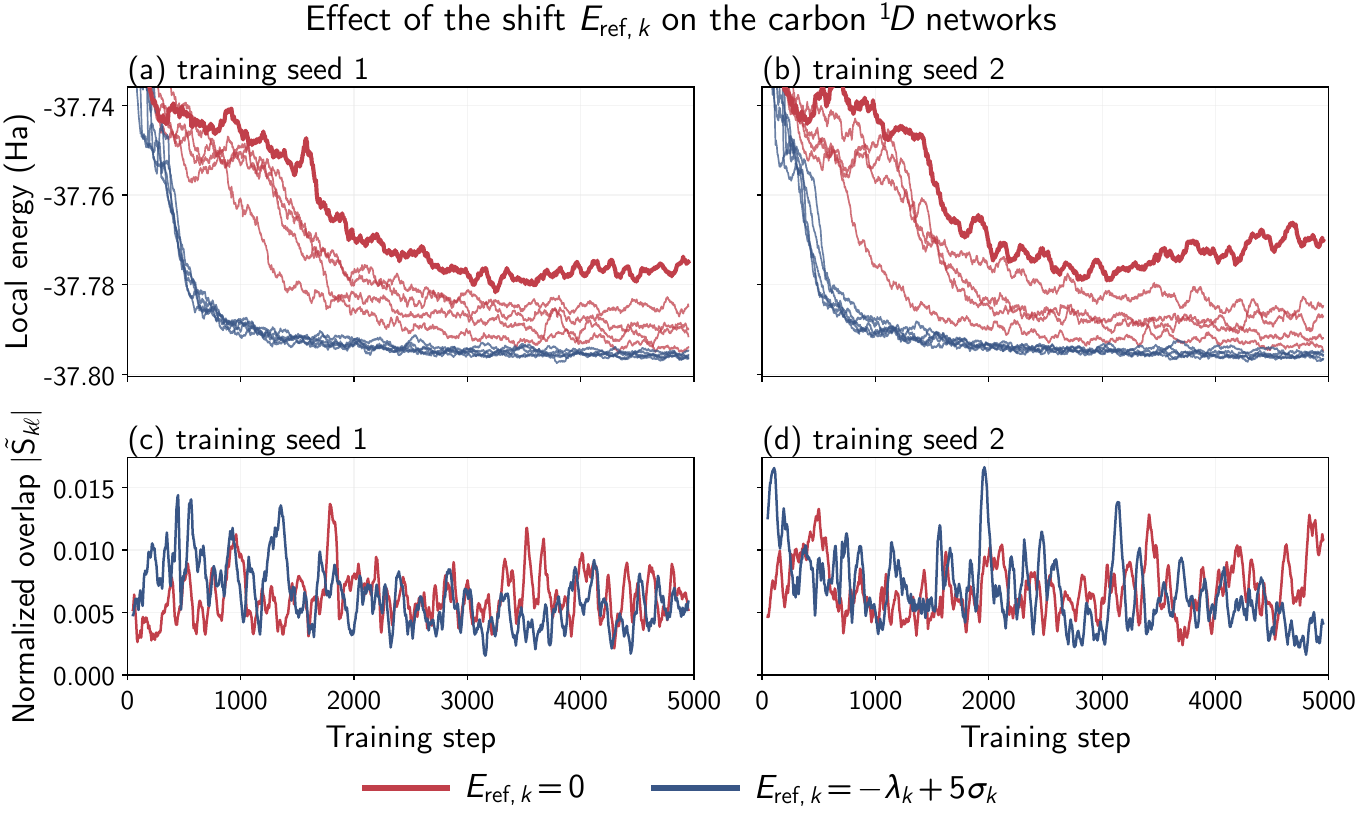}
  \caption{Carbon runs of Section~\ref{sec:exp_shifted_operator}, with $E_{\mathrm{ref},k}=0$ or with the shift $E_{\mathrm{ref},k}=-\lambda_k+5\sigma_k$, for training seed $1$ (a, c) and seed $2$ (b, d). (a, b) Batch mean of the clipped local energy of the five networks of the $^1\!D$ term; the thick line is the network furthest above the others with $E_{\mathrm{ref},k}=0$. (c, d) Normalized overlap from the unclipped batch mean, averaged over the $10$ pairs of these networks. Every curve is a mean over a sliding window of $100$ steps, taken before the absolute value in (c, d).}
  \label{fig:carbon_shift}
\end{figure}

\subsection{On the effectiveness of the per-network operator-shift on carbon}
\label{sec:exp_shifted_operator}

On carbon we test whether the shift of Section~\ref{sec:shifted_operator} improves the energy convergence of the five networks of the $2p^2$ $^1\!D$ term, whose exact energies are equal.
We train $5{,}000$ steps from the pretrained start of the seed-$1$ carbon run of Table~\ref{tab:atoms_mae}, with the setup of Section~\ref{sec:exp_setup} and training seeds $1$ and $2$.
For each seed one run uses $E_{\mathrm{ref},k}=0$, that is $\Top=-\hmltn$, and one the shift $E_{\mathrm{ref},k}=-\lambda_k+5\sigma_k$ (Figure~\ref{fig:carbon_shift}).
With $E_{\mathrm{ref},k}=0$ the five networks stay spread over the term.
With the shift their mean energies over the last $1{,}000$ steps span less than a tenth as much.

\section{Conclusion}
\label{sec:conclusion}

We present \algo, a new approach to excited-state VMC based on nested low-rank approximation.
Unlike penalty-based methods and NES-VMC, \algo exploits the structure of the spectral approximation problem to recover eigenstates in energy order while retaining per-network VMC optimization.
This provides a principled alternative to existing approaches, with computational savings in Laplacian evaluations that become increasingly significant as the number of target states grows.

An important next step is to extend \algo to molecular systems and larger $K$.
Its VMC-compatible formulation also opens opportunities to incorporate advanced optimization techniques, such as Wasserstein gradient flows~\citep{Neklyudov--Nys--Thiede--Carrasquilla--Liu--Welling--Makhzani2023} and improved discretizations of imaginary-time evolution~\citep{Su--Liu2024}.

\paragraph{Limitations.}
Empirical validation is limited to first-row atoms (Li--Ne) at $K{=}10$.
For molecular systems, reference excitation energies at fixed nuclear geometries are best estimates whose uncertainty grows with system size, motivating evaluation criteria beyond agreement with reference values.
Our multi-spin calculations also require preselecting spin sectors and allocating networks among them; automating this selection remains an open problem.
Finally, \algo retains the $O(\numel^4)$ Laplacian cost of standard neural-network VMC, which remains a major bottleneck for larger molecules.
Reducing this cost requires further mathematical and computational advances.

\clearpage
\begin{ack}
JJR and GWW were supported, in part, by the MIT-IBM Watson AI Lab under Agreement No. W1771646.
MJ and SYY were partly supported by the Institute of Information \& Communications Technology Planning \& Evaluation (IITP) grant funded by the Korean government (MSIT) (No.~RS-2019-II190075, Artificial Intelligence Graduate School Program (KAIST); No.~RS-2024-00457882, National AI Research Lab Project).
MJ was supported in part by Korea Institute for Advanced Study (KIAS) grant funded by the Korea government, and by the Center for Advanced Computation at KIAS.
\end{ack}

\section*{Competing interests}
All authors declare no competing interests.

{\small
\bibliographystyle{plainnat}
\bibliography{ref}
}

\newpage
\appendix
\addtocontents{toc}{\protect\StartAppendixEntries}
\listofatoc

\clearpage

\raggedbottom

\section{Notation summary}
\label{app:sec:notation}

Table~\ref{tab:notation} collects the symbols used throughout the body and appendix in one place.
Each symbol is defined where it first appears.

{\small
\setlength{\LTpre}{0pt}
\begin{longtable}{@{}lp{0.6\textwidth}@{}}
\caption{Notation summary.}
\label{tab:notation}\\
\toprule
\textbf{Symbol} & \textbf{Meaning} \\
\midrule
\endfirsthead
\toprule
\textbf{Symbol} & \textbf{Meaning} \\
\midrule
\endhead
\bottomrule
\endlastfoot
\multicolumn{2}{@{}l}{\textit{Particle and walker variables}} \\
$\numel,\,\numnu$ & Number of electrons / nuclei in the molecule \\
$K$ & Rank of the LoRA approximation; number of states, the ground state included \\
$\rv_i,\,\Rv_I$ & 3D positions of the $i$-th electron, $i \in [\numel]$, and of the $I$-th nucleus, $I \in [\numnu]$ \\
$\xv = (\rv_1,\ldots,\rv_{\numel})$ & Single-walker $\numel$-electron configuration \\
$\Xv = (\xv^1,\ldots,\xv^K)$ & $K$ electron configurations of NES-VMC, of which $\xv^i$ is the $i$-th \\
\addlinespace
\multicolumn{2}{@{}l}{\textit{Operators and spectrum}} \\
$\Hc$ & Hilbert space of wavefunctions antisymmetric under the exchange of two same-spin electrons \\
$\hmltn$ & Molecular electronic Hamiltonian \\
$\hat S^2$ & Total spin-squared operator \\
$\Top \defeq -\hmltn$ or $-(\hmltn + \hat S^2)$ & Negated Hamiltonian without and with spin penalty, target of HS approximation \\
$\idop$ \,/\, $\Imat$ & Identity operator (script) / identity matrix (sans-serif) \\
$\sigma_{\mathrm{ess}}(\Top)$ & Essential spectrum of $\Top$ \\
\addlinespace
\multicolumn{2}{@{}l}{\textit{Wavefunction ansatzes and architecture parameters}} \\
$\psi_k$ & Per-state wavefunction ansatz, $k \in [K]$ \\
$f_k$ & Unnormalized rank-1 LoRA component \\
$\varphi_k$ & Eigenfunction of $\Top$ \\
$\Psi(\Xv) \in \Complex^{K\times K}$ & Matrix $[\psi_k(\xv^i)]_{i,k=1}^K$ \\
$\Phi(\Xv) \defeq \det\Psi(\Xv)$ & NES-VMC total (determinantal) ansatz \\
$\theta$ & Neural-network parameters of the ansatz $\psi_\theta$ \\
\addlinespace
\multicolumn{2}{@{}l}{\textit{Local energies and Gram-type matrices}} \\
$\Ec_\th(\xv),\,\Ec_k(\xv)$ & Local energy $(\hmltn\psi_\th)(\xv)/\psi_\th(\xv)$ of the ansatz $\psi_\th$, and its per-state form $(\hmltn\psi_k)(\xv)/\psi_k(\xv)$ \\
$\locmat(\Xv) \defeq \Psi(\Xv)^{-1}(\hmltn\Psi)(\Xv)$ & NES-VMC local energy matrix ($K\!\times\!K$) \\
$\Sm_{k\ell} \defeq \ip{\psi_k}{\psi_\ell}$ & Overlap matrix ($K\!\times\!K$) \\
$\Hm_{k\ell} \defeq \ip{\psi_k}{\hmltn\psi_\ell}$ & Energy matrix ($K\!\times\!K$, NES-VMC) \\
\addlinespace
\multicolumn{2}{@{}l}{\textit{Born densities, expectations, and statistics}} \\
$p_k(\xv) \defeq |\psi_k(\xv)|^2 / \ip{\psi_k}{\psi_k}$ & Per-state Born density \\
$\E_{p_k}[\,\cdot\,]$ & Expectation under the per-state Born density $p_k$ \\
$\E_{\Xv\sim|\Phi|^2}[\,\cdot\,]$ & Expectation under the joint NES-VMC sampling distribution \\
$\Cov_{p_k}(\,\cdot\,,\,\cdot\,)$ & Covariance under $p_k$ \\
$\sigma_k$ & Walker standard deviation of the local energy of network $k$ in a step \\
$E_{\mathrm{ref},k},\,c$ & Reference energy $-\lambda_k + c\,\sigma_k$ of network $k$, with margin $c{=}5$ (Section~\ref{sec:shifted_operator}) \\
\addlinespace
\multicolumn{2}{@{}l}{\textit{LoRA objective and nesting}} \\
$\Lc_{\lora}(f_{1:K})$ & HS rank-$K$ LoRA objective in $f$ variables \\
$\Jc_{\lora}(\psi_{1:K},\eigvalparam_{1:K})$ & Normalized LoRA objective in $(\psi,\eigvalparam)$ variables \\
$\eigvalparam_k$ & Per-state scale parameter ($=\mu_k$ at the nesting fixed point) \\
$\Rm_{k\ell}^{\Oop} \defeq \ip{\psi_k}{\Oop\psi_\ell}/\ip{\psi_k}{\psi_k}$ & Unified ratio functional, $\Oop \in \{\Top, \idop\}$ \\
$\tilde{\Delta}_{k\ell}^{\Oop}$ & Score-coupled covariance (Appendix~\ref{app:sec:proof_sequential_nesting}) \\
$\matmask_{k\ell}^{\seq} \defeq \ones\{k > \ell\}$ & Sequential nesting mask (strictly lower triangular) \\
$\odot$ & Hadamard (entry-wise) product \\
\end{longtable}
}

\clearpage
\section{Related work}
\label{app:related-work}

\algo is penalty-free like NES-VMC and needs $K$ Laplacians per step like the penalty-based methods.
It takes its objective from the neural-network eigensolver NestedLoRA and learns the states in energy order through sequential nesting (Section~\ref{sec:nesting}).

\paragraph{Ab initio methods for excited states.}
The accurate methods of ab initio quantum chemistry grow steeply in cost with the number $\numel$ of electrons.
Full configuration interaction expands $\psi$ in Slater determinants and gives the ordered eigenstates exactly, but needs exponentially many determinants~\citep{Pfau--Spencer--Matthews--Foulkes2020}.
Coupled cluster with single, double and perturbative triple excitations, CCSD(T), costs $O(\numel^7)$ for the ground state~\citep{glehn2023a}, and equation-of-motion coupled cluster costs $O(\numel^6)$ or more for excited states~\citep{Stanton--Bartlett1993,Krylov2008}.
Time-dependent density functional theory is far cheaper, but its usual approximations fail for long-range charge-transfer excitations and miss double excitations~\citep{Dreuw--Weisman--Head-Gordon2003,Maitra--Zhang--Cave--Burke2004}.

\paragraph{Neural-network VMC for the ground state.}
Neural-network VMC represents the wavefunction by a neural network and minimizes its energy with stochastic optimizers~\citep{Kingma--Ba2014,pmlr-v37-martens15}, using samples of the Born density as in Section~\ref{sec:prelim}; \citet{Hermann--Spencer--Choo--Mezzacapo--Foulkes--Pfau--Carleo--Noe2023} review the field.
On the ground states of small molecules, FermiNet~\citep{Pfau--Spencer--Matthews--Foulkes2020}, PauliNet~\citep{Hermann--Schatzle--Noe2020} and Psiformer~\citep{glehn2023a} reach chemical accuracy, an error below $1$\,kcal/mol or about $1.6$ millihartree (mHa).
With the forward Laplacian~\citep{Li--Ye--Jiang--Wen--Wang--Li--Li--He--Chen--Ren--Wang2024}, each optimization step costs $O(\numel^4)$ for a fixed network size, against $O(\numel^7)$ for CCSD(T).
\algo keeps this machinery for every state: each $\psi_k$ is its own network and is sampled from its own Born density.

\paragraph{Excited-state VMC.}
Penalty-based methods~\citep{Pathak--Busemeyer--Rodrigues--Wagner2021,Entwistle--Schatzle--Erdman--Hermann--Noe2023,Wheeler--Kleiner--Wagner2024,Szabo--Schatzle--Entwistle--Noe2024} train one ansatz per state on the sum of the energies plus pairwise overlap penalties, and the penalty weights set whether the minimum is the ordered eigenstates.
\citet{Gao--Grutschus--Noe--Guennemann2026} train one network for many excited states across molecular geometries on a penalty objective, and estimate the overlaps from samples pooled over all states.

NES-VMC~\citep{Pfau--Axelrod--Sutterud--von-Glehn--Spencer2024} minimizes the energy of an extended system whose ansatz is the determinant of the $K$ ansatzes evaluated on $K$ configurations, and obtains the individual states by a diagonalization after training.
The inverse of this $K\times K$ matrix enters its local energy, which therefore develops heavy tails when two ansatzes are nearly linearly dependent.
\citet{Hendry--Sinibaldi--Carleo2026} write this objective as the energy of a $K$-dimensional subspace and derive a stochastic reconfiguration on the manifold of such subspaces, and \citet{Kahn--Gravina--Vicentini2026} derive the same formalism and apply it to time evolution.
Section~\ref{sec:prelim_excited_prior_works} describes the penalty-based methods and NES-VMC, and Appendix~\ref{app:prior-methods} states their objectives in full.

\paragraph{Neural-network eigensolvers.}
Outside VMC, several methods learn the leading eigenfunctions of an operator with neural networks and estimate inner products under a fixed data or proposal distribution.
SpIN~\citep{Pfau--Petersen--Agarwal--Barrett--Stachenfeld2018} maximizes a trace objective whitened by the covariance of the learned functions and orders them with a masked gradient.
NeuralEF~\citep{Deng--Shi--Zhu2022} solves a sequence of Rayleigh-quotient problems in which each function is penalized against the earlier ones.
NestedLoRA~\citep{Ryu--Xu--Erol--Bu--Zheng--Wornell2024} minimizes the error of a rank-$K$ approximation of the operator, a polynomial in the inner products of the learned functions, and orders the eigenfunctions by nesting, which descends from the triangularized gradient of \citet{Gao--Li--Lu2021}.

\algo brings the NestedLoRA objective to VMC: the reparameterization of Section~\ref{sec:nestedlora_vmc} writes it as expectations under the Born densities of the ansatzes.
NestedLoRA has been applied to low-dimensional partial differential equations and representation learning~\citep{Ryu--Xu--Erol--Bu--Zheng--Wornell2024} and to the Koopman operators of stochastic dynamical systems~\citep{Jeong--Ryu--Yun--Wornell2025}; \citet{Ryu--Zhou--Wornell2025} apply nesting to the objective of the orbital minimization method, and \citet{Jedra--Shah2025} study $k$-SVD by gradient descent with the same low-rank-approximation principle.

\section{Technical details for prior methods}
\label{app:prior-methods}

This section states the objectives of the prior excited-state methods that the main text compares with \algo, in the notation of this paper.

\subsection{Natural-excited-states}
\label{app:sec:nes_statements}

We state the two forms of the natural-excited-states Rayleigh quotient in Eq.~\eqref{eq:nes_two_forms} and the expectation of the total local energy matrix.
Let the matrix ansatz $\Psi(\Xv)\in\Complex^{K\times K}$ and the total ansatz $\Phi(\Xv)=\det\Psi(\Xv)$ be defined as in Section~\ref{sec:prelim_excited_prior_works}, and let the extended Hamiltonian be $\totalhmltn = \sum_{i=1}^K \totalhmltn_i$ with $\totalhmltn_i = \idop^{\otimes(i-1)}\otimes\hmltn\otimes\idop^{\otimes(K-i)}$, which acts as $\hmltn$ on the $i$-th configuration $\xv^i$ and as the identity on the other $K{-}1$.

\begin{theorem}
\label{thm:nes_rq_basic}
Let $\Sm \defeq [\ip{\psi_k}{\psi_\ell}]_{k,\ell=1}^K$ be the overlap matrix and $\Hm \defeq [\ip{\psi_k}{\hmltn\psi_\ell}]_{k,\ell=1}^K$ the energy matrix.
Then

\begin{align*}
\frac{\ip{\Phi}{\totalhmltn\Phi}}{\ip{\Phi}{\Phi}}
&= \tr(\Sm^{-1}\Hm).
\end{align*}

\end{theorem}

\begin{theorem}[\citep{Pfau--Axelrod--Sutterud--von-Glehn--Spencer2024}]
\label{thm:nes_rq_vmc}

\begin{align*}
\frac{\ip{\Phi}{\totalhmltn\Phi}}{\ip{\Phi}{\Phi}}
=\E_{\Xv\sim |\Phi|^2}\biggl[\frac{\totalhmltn\Phi(\Xv)}{\Phi(\Xv)}\biggr]
&= \E_{\Xv\sim |\Phi|^2}[\tr(\locmat(\Xv))],
\end{align*}
where the \emph{total local energy matrix} is $\locmat(\Xv)\defeq \Psi(\Xv)^{-1}(\hmltn\Psi)(\Xv) \in \Complex^{K\times K}$, with $(\hmltn\Psi)(\Xv) \defeq \bigl((\hmltn\otimes\cdots\otimes \hmltn)\Psi\bigr)(\Xv) = [\,(\hmltn\psi_k)(\xv^i)\,]_{i,k=1}^K$.
\end{theorem}

\begin{proposition}
\label{prop:nes_locmat_mean}
If $\psi_{1:K}$ are linearly independent, then $\E_{\Xv\sim|\Phi|^2}[\locmat(\Xv)]=\Sm^{-1}\Hm$.
Its trace gives $\tr(\Sm^{-1}\Hm)=\E_{\Xv\sim|\Phi|^2}[\tr(\locmat(\Xv))]$, the equality of the right-hand sides of Theorems~\ref{thm:nes_rq_basic} and~\ref{thm:nes_rq_vmc}.
\end{proposition}
\begin{proof}
By Cramer's rule, $\Phi(\Xv)\,[\locmat(\Xv)]_{jk}$ is the determinant of $\Psi(\Xv)$ with its $j$-th column replaced by $[(\hmltn\psi_k)(\xv^i)]_{i=1}^K$.
The Andr\'eief identity $\int\bigl(\det[f_b(\xv^a)]_{a,b}\bigr)^*\det[g_b(\xv^a)]_{a,b}\,d\Xv=K!\,\det[\ip{f_a}{g_b}]_{a,b}$ then gives $\int\Phi^*(\Xv)\,\Phi(\Xv)\,[\locmat(\Xv)]_{jk}\,d\Xv=K!\det\Sm^{(j,k)}$, where $\Sm^{(j,k)}$ is $\Sm$ with its $j$-th column replaced by the $k$-th column of $\Hm$, and $\int|\Phi(\Xv)|^2\,d\Xv=K!\det\Sm$, which is nonzero by linear independence.
The ratio $\det\Sm^{(j,k)}/\det\Sm$ is $(\Sm^{-1}\Hm)_{jk}$, again by Cramer's rule.
\end{proof}

\subsection{Penalty-based methods}
\label{app:sec:penalty_explicit}

\paragraph{PauliNet penalty (Entwistle et al., 2023).}
\citet{Entwistle--Schatzle--Erdman--Hermann--Noe2023} extended PauliNet to excited states with the divergent overlap penalty
\begin{align*}
\Lc_{\mathsf{penalty}}(\psi_{1:K})
&= \sum_{\kidx=1}^K \E_{p_{\kidx}}[\Ec_{\kidx}]
+ \a \sum_{i<j} \frac{1}{1-|\tilde{\Sm}_{ij}|},
\quad
\tilde{\Sm}_{ij}\defeq \frac{\ip{\psi_{i}}{\psi_{j}}}{\|\psi_{i}\| \|\psi_{j}\|}.
\end{align*}
The symmetrized overlap is estimated consistently by the product of expectations
\begin{align*}
\tilde{\Sm}_{ij}
&=\sgn\biggl(\E_{p_{i}}\biggl[\frac{\psi_{j}}{\psi_{i}}\biggr]\biggr) \times \sqrt{\E_{p_{i}}\biggl[\frac{\psi_{j}}{\psi_{i}}\biggr] \E_{p_{j}}\biggl[\frac{\psi_{i}}{\psi_{j}}\biggr]}.
\end{align*}

\paragraph{Pathak--Wheeler form.}
\citet{Pathak--Busemeyer--Rodrigues--Wagner2021,Wheeler--Kleiner--Wagner2024} use a squared overlap that admits a direct unbiased estimator,
\begin{align*}
\Lc_{\mathsf{penalty}}(\psi_{1:K})
&= \sum_{\kidx=1}^K \E_{p_{\kidx}}[\Ec_{\kidx}]
+ \sum_{i<j} \a_{ij}\,|\tilde{\Sm}_{ij}|^2.
\end{align*}
\citet{Wheeler--Kleiner--Wagner2024} established an ensemble variational principle for this family.
Its objective carries strictly decreasing energy weights $w_1 > \cdots > w_K > 0$ on $\sum_i w_i \E_{p_i}[\Ec_i]$ and a single global penalty $\a\sum_{i<j}|\tilde{\Sm}_{ij}|^2$.
The ordered bottom-$K$ eigenfunctions of $\hmltn$ are then the global minimum once $\a > \a_c = \max_{i<j}\bigl[(E_j-E_i)\,w_iw_j/(w_i-w_j)\bigr]$, with $E_i$ the eigenvalues the optimum identifies.
The strictly decreasing weights break the rotational degeneracy of the bottom-$K$ subspace: at equal weights $\a_c \to \infty$ and the minimum is the subspace, not its ordered basis.
Because $\a_c$ scales with the spectral gaps $E_j - E_i$, the penalty schedule is coupled to the spectrum the optimization computes.
The per-pair $\a_{ij}$ form displayed above is the practical variant; \citet{Szabo--Schatzle--Entwistle--Noe2024} scale $\a_{ij}$ to running gap estimates (below).

\citet{Szabo--Schatzle--Entwistle--Noe2024} adopted the Pathak--Wheeler form with three modifications.
First, the orthogonality constraint is applied with respect to lower-lying states only: gradients of the overlap term flow only into $\psi_j$ for $j>i$, while $\psi_i$ is detached from the computational graph (their Section~2.1.2).
Second, an explicit per-state spin penalty $\beta\sum_i\langle \hat{S}^2\rangle_i$ (their Eq.~7) selects the target spin sector at training time; Appendix~\ref{app:sec:spin_penalty_impl} gives the corresponding operator shift in our framework, and Appendix~\ref{app:sec:atom_setup} the spin sectors that their setting implies.
Third, $\a_{ij}$ is set at every step to (their Eq.~17)
\begin{align*}
\a_{ij} = \tilde\a \cdot \max\bigl\{ |\mathrm{ema}(\hat{E}_j - \hat{E}_i)|,\ \mathrm{ema}(\mathrm{std}_{p_i}(\Ec_i)),\ 0.001 \bigr\}\,\ones\{i<j\},
\end{align*}
where $\hat E_i$ is the batch mean of $\Ec_i$, $\mathrm{std}_{p_i}(\Ec_i)$ is its batch standard deviation, and $\mathrm{ema}(\,\cdot\,)$ is an EMA over training steps.
This reduces the user-tunable hyperparameter to a single global scale $\tilde\a$ (default $\tilde\a=4$, with $\tilde\a=2$ for Li, Be and $\tilde\a=1$ for benzene per their SI Table~S6).

\section{Discussion on the mathematical formulation}
\label{sec:lora_detailed_discussion}

The LoRA objective of \algo is well defined for the molecular Hamiltonian, its global minimizers span the lowest $K$ eigenstates, and the fixed point of sequential nesting gives these eigenstates in energy order.
The first claim holds for any $K$ trial functions in $\Hc$, because the objective of the unbounded $\Top$ is the Hilbert--Schmidt objective of $\Top$ projected onto their span, an operator of finite rank (Appendix~\ref{app:sec:lora_well_defined}).
The other two claims hold when the lowest $K$ energies of $\hmltn$ lie strictly below its essential spectrum, which we assume in Appendix~\ref{app:sec:spectral_hypothesis}, and are shown in Appendices~\ref{app:lora-consistency} and~\ref{app:sec:nesting_fixed_point}.

\subsection{Mathematical preliminaries}

The three claims are statements about $\hmltn$ on the spatial space $\Hc$, and this subsection gives the mathematical facts about $\Hc$ and the spectrum of $\hmltn$ that their proofs use.

\subsubsection{Reduction to the spatial space $\Hc$}
\label{app:sec:spin_content}

The eigenstates of the electrons with $n_\uparrow$ spins up correspond one-to-one to the eigenfunctions of $\hmltn$ in $\Hc$, with the same energies, and their levels are those of total spin $S\ge|m_S|$, where $m_S=(n_\uparrow-n_\downarrow)/2$.

The space $\Hc$ is the subspace of $L^2(\mathbb{R}^{3\numel})$ of functions of the electron positions $\xv=(\rv_1,\ldots,\rv_\numel)$ that are antisymmetric under the exchange of two electrons of the same spin, where electrons $1,\ldots,n_\uparrow$ have spin up and the other $n_\downarrow=\numel-n_\uparrow$ have spin down.
The antisymmetrizer over the permutations within each spin is an orthogonal projection with range $\Hc$, so $\Hc$ is closed and is a Hilbert space with the inner product of $L^2(\mathbb{R}^{3\numel})$.
Our network architecture restricts the functions it can represent to $\Hc$ by construction (Appendix~\ref{app:sec:architectures}).

An electronic state (\emph{state} in this section) is a function of the positions and spins of the electrons, an element of $L^2(\mathbb{R}^{3\numel})\otimes(\mathbb{C}^2)^{\otimes\numel}$ that changes sign when two electrons exchange both position and spin.
The states with $n_\uparrow$ spins up correspond one-to-one to the functions in $\Hc$, and the correspondence is a linear isomorphism: it maps $\psi\in\Hc$ to the antisymmetrization $\Psi$, over all electrons, of $\psi$ times the spin function with electrons $1,\ldots,n_\uparrow$ up.
Since $\hmltn$ acts on the positions alone and is symmetric in the electrons, it acts on states as $\hmltn\otimes\idop$, which commutes with this antisymmetrization, so $(\hmltn\otimes\idop)\Psi=E\Psi$ exactly when $\hmltn\psi=E\psi$.

The total spin operators $\hat S^2$, with eigenvalues $S(S{+}1)$, and $\hat S_z$ act on the spins of the states, and the states with $n_\uparrow$ spins up form the eigenspace of $\hat S_z$ with eigenvalue $m_S$.
Both commute with $\hmltn\otimes\idop$, so each of its eigenspaces is spanned by multiplets of total spin $S$, each consisting of $2S+1$ states with $\hat S_z$ eigenvalues $-S,\ldots,S$~\citep[Sections~5.2--5.3]{Pauncz1995}.
A multiplet contains one state with $n_\uparrow$ spins up exactly when $S\ge|m_S|$, so the states with $n_\uparrow$ spins up form the orthogonal sum of their subspaces of total spin $S=|m_S|,|m_S|+1,\ldots,\numel/2$, each left invariant by $\hmltn\otimes\idop$.
Through the isomorphism, $\Hc=\bigoplus_{S\ge|m_S|}\Hc^{(S)}$, where $\Hc^{(S)}$ corresponds to total spin $S$ and is left invariant by $\hmltn$, and the eigenvalues of $\hmltn$ in $\Hc$ are the energies of total spin $S\ge|m_S|$, with one eigenfunction per multiplet.

\paragraph{Example: lithium.}
The antisymmetry that defines $\Hc$ excludes functions with energy below the physical ground state.
Consider lithium with $m_S$ fixed at $1/2$, that is, $n_\uparrow=2$ and $n_\downarrow=1$.
The spatial function $1s(\rv_1)\,1s(\rv_2)\,1s(\rv_3)$ of three hydrogenic orbitals is symmetric in the three electrons and, for a suitable orbital exponent, has an energy below the physical ground state; it lies outside $\Hc$, because it is symmetric under the exchange of the two spin-up electrons.

\subsubsection{Self-adjoint operators and their spectrum}
\label{app:sec:operator_spectrum}

The min--max values of a self-adjoint operator $\Top$ bounded above list its eigenvalues above the essential spectrum from the top, and the paragraphs below define these terms for a general $\Top$.

\paragraph{Self-adjoint operators.}
A linear operator $\Top$ on a dense domain $D(\Top)$ of a Hilbert space is \emph{bounded} if $\|\Top f\|\le C\|f\|$ for a constant $C$ and all $f\in D(\Top)$.
It is \emph{symmetric} if $\ip{f}{\Top g}=\ip{\Top f}{g}$ for all $f,g\in D(\Top)$, and \emph{self-adjoint} if, in addition, its adjoint has the same domain, $D(\Top^\dagger)=D(\Top)$.

\paragraph{Spectrum.}
For a self-adjoint $\Top$, the spectrum $\sigma(\Top)$ is the set of $z\in\Real$ for which $\Top-z$ has no bounded inverse on the Hilbert space.
The discrete spectrum $\sigma_{\mathrm{disc}}(\Top)$ is the set of isolated eigenvalues of finite multiplicity, and the essential spectrum is $\sigma_{\mathrm{ess}}(\Top)\defeq\sigma(\Top)\setminus\sigma_{\mathrm{disc}}(\Top)$.

\paragraph{Spectral theorem.}
By the spectral theorem in projection-valued-measure form~\citep[Theorem~VIII.6]{ReedSimon1}, a self-adjoint $\Top$ satisfies $\Top=\int z\,dP_\Top(z)$, where the spectral projection $P_\Top(B)$ of $\Top$ onto a Borel set $B\subset\Real$ is an orthogonal projection.
It vanishes when $B$ is disjoint from $\sigma(\Top)$, and $P_\Top(\{z\})$ projects onto the eigenspace of an eigenvalue $z$~\citep[pp.~236 and~264, Theorem~VIII.5(e)]{ReedSimon1}.
For $\psi\in D(\Top)$ this gives $\ip{\psi}{\Top\psi}=\int z\,d\nu_\psi(z)$ with the spectral measure $\nu_\psi(B)\defeq\|P_\Top(B)\psi\|^2$.

\paragraph{Min--max values.}
For a self-adjoint $\Top$ bounded above and $j\ge1$, the min--max values
\begin{align}\label{eq:minmax_values}
\mu_j \defeq \sup_{\Vc\subset D(\Top),\,\dim\Vc=j}\ \min_{\psi\in\Vc,\,\|\psi\|=1}\ip{\psi}{\Top\psi}
\end{align}
satisfy $\mu_1\ge\mu_2\ge\cdots$.
By the min--max principle for $-\Top$, which is bounded below~\citep[Theorem~XIII.1]{ReedSimon4}, the $\mu_j$ greater than $\sup\sigma_{\mathrm{ess}}(\Top)$ are the eigenvalues of $\Top$ greater than this value, in decreasing order and counted with multiplicity, and every other $\mu_j$ equals $\sup\sigma_{\mathrm{ess}}(\Top)$.

\subsubsection{The molecular Hamiltonian on $\Hc$}
\label{app:sec:hamiltonian_structure}

The molecular Hamiltonian $\hmltn$ is self-adjoint on $\Hc$, so it has a real spectrum.

The non-relativistic electronic Coulomb Hamiltonian (cf.~Section~\ref{sec:prelim})
\begin{align*}
\hmltn = -\frac{1}{2}\sum_{i=1}^{\numel} \nabla_{\rv_i}^2
- \sum_{i=1}^{\numel}\sum_{I=1}^{\numnu} \frac{Z_I}{\|\rv_i - \Rv_I\|}
+ \sum_{1 \le i < j \le \numel} \frac{1}{\|\rv_i - \rv_j\|}
\end{align*}
is unbounded but self-adjoint on $L^2(\mathbb{R}^{3\numel})$ with the Sobolev space $H^2(\mathbb{R}^{3\numel})$ as its dense domain, by the Kato--Rellich theorem~\citep{Kato1951}, since the Coulomb potential satisfies the Kato smallness condition with respect to the kinetic operator.

The restriction of $\hmltn$ to $D(\hmltn) \defeq H^2(\mathbb{R}^{3\numel})\cap\Hc$ is self-adjoint on $\Hc$, because $\hmltn$ commutes with the exchange of two same-spin electrons and the antisymmetrizer maps $H^2(\mathbb{R}^{3\numel})$ into itself, which also makes $D(\hmltn)$ dense in $\Hc$.
We work with $\Top = -\hmltn$ on $D(\Top) = D(\hmltn)$ throughout, and every function on which $\Top$ acts is taken in $D(\Top)$ without further mention.

\subsubsection{Spectrum and bound states}
\label{app:sec:spectral_hypothesis}

On $\Hc$, the spectrum of $\hmltn$ consists of isolated eigenvalues of finite multiplicity below the ionization threshold $\Sigma$ and the essential spectrum $\sigma_\mathrm{ess}(\hmltn)=[\Sigma,\infty)$ above it.
On each $\Hc^{(S)}$, the Hunziker--van Winter--Zhislin (HVZ) theorem~\citep{Hunziker1966, Zhislin1960, ReedSimon4} in its form for a fixed total spin~\citep{Zhislin--Mandel1969} gives the essential spectrum $[\Sigma_S,\infty)$, where $\Sigma_S$ is the lowest energy of the $(\numel{-}1)$-electron system with total spin $S-\frac12$ or $S+\frac12$.
Since $\Hc$ is the finite orthogonal sum of the $\Hc^{(S)}$, each left invariant by $\hmltn$ (Appendix~\ref{app:sec:spin_content}), the essential spectrum on $\Hc$ is the union of these intervals, $[\Sigma,\infty)$ with $\Sigma=\min_S\Sigma_S$, and below $\Sigma$ the discrete spectrum is the union of those on the $\Hc^{(S)}$.
The threshold $\Sigma$ is the lower of the ground-state energies of the two $(\numel{-}1)$-electron systems obtained by removing one spin-up or one spin-down electron to infinity.

The eigenfunctions of $\hmltn$ in $\Hc$ are the bound states, and they span only part of $\Hc$.
The energies of the bound states below $\Sigma$ form the discrete spectrum, and these states are localized near the nuclei.
The essential spectrum $[\Sigma,\infty)$ contains the energies of scattering states, which have at least one electron at infinity and whose solutions of $\hmltn\psi=E\psi$ lie outside $\Hc$.
It can also contain eigenvalues: a bound state of total spin $S$ with energy between $\Sigma$ and $\Sigma_S$ is discrete on $\Hc^{(S)}$ and embedded in the essential spectrum on $\Hc$.
The orthogonal complement of the closed span of the bound states, the continuous spectral subspace, holds square-integrable states such as a wave packet of an electron moving away from the nuclei.
The orthogonal complement of the bound states below $\Sigma$, the essential spectral subspace, contains the continuous spectral subspace and the bound states with energies from $\Sigma$ upward.

We assume that $\Top = -\hmltn$ has at least $K$ discrete eigenvalues separated from the essential spectrum:
\begin{align}\label{eq:spectral-gap}
\mu_1 \ge \mu_2 \ge \cdots \ge \mu_K > \sup\, \sigma_{\mathrm{ess}}(\Top).
\end{align}
Since $\sup\sigma_{\mathrm{ess}}(\Top)=-\Sigma$, this is the existence of at least $K$ bound states below $\Sigma$, which holds for all systems in our experiments.
The number of bound states below $\Sigma$ depends on the system; neutral atoms and positive ions have infinitely many by Zhislin's theorem~\citep{Zhislin--Mandel1969}.

\subsection{The LoRA objective for an unbounded operator}
\label{app:sec:lora_well_defined}
\label{app:sec:galerkin_ey}

The LoRA objective of any trial functions $f_1,\ldots,f_K$ is the Hilbert--Schmidt objective of the projected operator $\Top_{\Vc} \defeq P_{\Vc}\, \Top\, P_{\Vc}$, an operator of finite rank, where $P_\Vc$ is the orthogonal projection onto $\Vc \defeq \mathrm{span}\{f_1, \ldots, f_K\}$; this holds also for the unbounded molecular $\Top$.

\paragraph{Hilbert--Schmidt operators.}
A bounded operator on $\Hc$ is \emph{Hilbert--Schmidt} (HS) if $\|\Top\|_{\mathsf{HS}}^2 \defeq \sum_n \|\Top e_n\|^2 < \infty$ for an orthonormal basis $(e_n)$ of $\Hc$, a sum that does not depend on the basis, and \emph{finite-rank} if its range is finite-dimensional; a finite-rank operator is HS.

The body's identity $\Lc_\lora(f_{1:K}) = \|\Top - \sum_k |f_k\rangle\langle f_k|\|_{\mathsf{HS}}^2 - \|\Top\|_{\mathsf{HS}}^2$ assumes $\Top$ to be HS, which fails for molecular $\Top$.
The objective itself is defined for any $f_1, \ldots, f_K$, since the matrix elements $\ip{f_k}{\Top f_\ell}$ are finite.
Like the Rayleigh quotient, which evaluates $\Top$ on its trial function alone, the objective depends on $\Top$ only through these matrix elements, that is, through $\Top$ on $\Vc$.
The projected operator $\Top_{\Vc}$ is finite-rank, hence HS, and has the same matrix elements, $\ip{f_k}{\Top f_\ell} = \ip{f_k}{\Top_{\Vc} f_\ell}$, so the identity holds exactly for $\Top_\Vc$:
\begin{align}\label{eq:galerkin-ey}
\Lc_{\lora}(f_{1:K})
&= \biggl\|\sum_{k=1}^K |f_k\rangle\langle f_k|
    - \Top_{\Vc}\biggr\|_{\mathsf{HS}}^2
    - \|\Top_{\Vc}\|_{\mathsf{HS}}^2.
\end{align}
Within a fixed $\Vc$, minimizing $\Lc_\lora$ is therefore the best positive-semidefinite (PSD) rank-$K$ approximation of $\Top_\Vc$, and the global minimization also moves $\Vc$.

\subsection{Global minimizers of $\Jc_\lora$}
\label{app:lora-consistency}

\begin{theorem}[Global minimizers of $\Jc_\lora$]
\label{thm:lora_vmc_consistency}
Let $\Top$ be self-adjoint and bounded above on $\Hc$, with the min--max values $\mu_j$ of Eq.~\eqref{eq:minmax_values}, and assume
\begin{align}\label{eq:consistency_hyp}
\mu_K > \max\bigl(\sup\sigma_{\mathrm{ess}}(\Top),\,0\bigr).
\end{align}
Then $\mu_1\ge\cdots\ge\mu_K$ are the top $K$ eigenvalues of $\Top$, counted with multiplicity, and
\begin{align*}
\min_{\psi_{1:K}\neq0,\ \eigvalparam_{1:K}\in\Real_{\ge0}^K} \Jc_{\lora}(\psi_{1:K},\eigvalparam_{1:K}) = -\sum_{k=1}^{K}\mu_k^2 .
\end{align*}
Every minimizer satisfies
\begin{align*}
\sum_{k=1}^K \eigvalparam_k \frac{|\psi_k\rangle\langle\psi_k|}{\ip{\psi_k}{\psi_k}} = \sum_{k=1}^K \mu_k |\varphi_k\rangle\langle\varphi_k|
\end{align*}
for some orthonormal $\varphi_{1:K}$ with $\Top\varphi_k=\mu_k\varphi_k$.
If moreover $\mu_K>\mu_{K+1}$, this operator is the same for every minimizer: it is the weighted spectral projector of $\Top$ onto its top-$K$ eigenspace.
\end{theorem}

The min--max formula defines $\mu_j$ for every self-adjoint $\Top$ bounded above, including the operator with the spin penalty of Appendix~\ref{app:sec:lora_remarks}, so Eq.~\eqref{eq:consistency_hyp} can be stated for any such $\Top$; the first claim is the min--max principle of Appendix~\ref{app:sec:operator_spectrum}, since $\mu_K>\sup\sigma_{\mathrm{ess}}(\Top)$.
For $\Top=-\hmltn$, $\sup\sigma_{\mathrm{ess}}(\Top)=-\Sigma$ with the ionization threshold $\Sigma\le0$ of Appendix~\ref{app:sec:spectral_hypothesis}, so Eq.~\eqref{eq:consistency_hyp} reduces to Eq.~\eqref{eq:spectral-gap}.

\subsubsection{Two lemmas}
\label{app:sec:lora_lemmas}

The proof uses two lemmas, with $\Top$ and $\mu_j$ as in Theorem~\ref{thm:lora_vmc_consistency} and Eq.~\eqref{eq:consistency_hyp} assumed.
The first is the lower bound, through the projected operator $\Top_\Vc$ of Appendix~\ref{app:sec:lora_well_defined}.
For a finite-rank self-adjoint $S$, write $S_+$ for its positive part, and $(x)_+\defeq\max(x,0)$.

\begin{lemma}[Lower bound]
\label{lem:galerkin_bound}
Assume Eq.~\eqref{eq:consistency_hyp}.
For $f_1,\ldots,f_K$, let
\begin{itemize}[leftmargin=*,itemsep=2pt,topsep=2pt]
\item $\Vc=\mathrm{span}\{f_{1:K}\}$ and $d=\dim\Vc$,
\item $A_\Vc=\sum_k|f_k\rangle\langle f_k|$, and
\item $\omega_1\ge\cdots\ge\omega_d$ the eigenvalues of $\Top_\Vc$ on $\Vc$.
\end{itemize}
Then
\begin{align*}
\Lc_\lora(f_{1:K}) \ge -\sum_{j=1}^d(\omega_j)_+^2 \ge -\sum_{k=1}^K\mu_k^2 .
\end{align*}
The first inequality is an equality only if $A_\Vc=(\Top_\Vc)_+$.
The second is an equality only if $d=K$ and $\omega_j=\mu_j$ for every $j$.
\end{lemma}
\begin{proof}
Consider
\begin{align*}
\Lc_\lora(f_{1:K})
&\stackrel{(a)}{=} \|\Top_\Vc - A_\Vc\|_{\mathsf{HS}}^2 - \|\Top_\Vc\|_{\mathsf{HS}}^2\\
&\stackrel{(b)}{\ge} \min_{A\succeq0,\,\mathrm{rank}(A)\le K}\|\Top_\Vc - A\|_{\mathsf{HS}}^2 - \|\Top_\Vc\|_{\mathsf{HS}}^2 \\
&\stackrel{(c)}{=} \|\Top_\Vc-(\Top_\Vc)_+\|_{\mathsf{HS}}^2-\|\Top_\Vc\|_{\mathsf{HS}}^2 \stackrel{(d)}{=} -\|(\Top_\Vc)_+\|_{\mathsf{HS}}^2 \stackrel{(e)}{=} -\sum_{j=1}^d(\omega_j)_+^2\\
&\stackrel{(f)}{\ge} -\sum_{k=1}^K\mu_k^2.
\end{align*}
The steps hold for the following reasons.
\begin{itemize}[leftmargin=*,itemsep=2pt,topsep=2pt]
\item (a) is Eq.~\eqref{eq:galerkin-ey}.
\item (b) holds since $A_\Vc$ is PSD of rank $\le K$.
\item (c) holds since $\Top_\Vc$ has rank $\le K$, so the rank constraint is inactive and the best PSD approximant is $(\Top_\Vc)_+$ (a PSD $A$ cannot match the negative eigenvalues of $\Top_\Vc$).
\item (d) and (e) are direct calculations using $\Top_\Vc-(\Top_\Vc)_+=(\Top_\Vc)_-\perp(\Top_\Vc)_+$.
\item (f) is due to the Courant--Fischer min--max principle, i.e., $\omega_j\le\mu_j$, and monotonicity under $(\cdot)_+^2$ since $\mu_j>0$.
\end{itemize}
The nearest PSD operator in Hilbert--Schmidt norm is unique, so (b) is tight only if $A_\Vc=(\Top_\Vc)_+$.
Since $(\omega_j)_+^2\le\mu_j^2$ for each $j\le d$ and $\mu_j>0$ for every $j\le K$, (f) is tight only if $d=K$ and $\omega_j=\mu_j$ for every $j$.
\end{proof}

Equality in Lemma~\ref{lem:galerkin_bound} fixes the eigenvalues of $\Top_\Vc$ on $\Vc$, and the second lemma is the induction step of the proof of the theorem: given eigenfunctions $\varphi_1,\ldots,\varphi_{j-1}$, it turns a vector of the subspace $W_j$ of Lemma~\ref{lem:complement_bound} that attains $\mu_j$ into an eigenfunction of $\Top$.

\begin{lemma}[Orthogonal complement]
\label{lem:complement_bound}
Let $1\le j\le K$, let $\varphi_1,\ldots,\varphi_{j-1}$ be orthonormal with $\Top\varphi_i=\mu_i\varphi_i$, and let $W_j$ be the orthogonal complement of their span.
Then $\ip{\psi}{\Top\psi}\le\mu_j$ for every unit $\psi\in W_j\cap D(\Top)$, and a unit $\varphi\in W_j\cap D(\Top)$ with $\ip{\varphi}{\Top\varphi}=\mu_j$ satisfies $\Top\varphi=\mu_j\varphi$.
\end{lemma}

\begin{proof}
Let $Q=P_\Top\bigl((\mu_j,\infty)\bigr)$; we show that $Q\psi=0$ for $\psi\in W_j$.
Eq.~\eqref{eq:consistency_hyp} and $j\le K$ give $\mu_j\ge\mu_K>\sup\sigma_{\mathrm{ess}}(\Top)$, so every point of the spectrum of $\Top$ in $(\mu_j,\infty)$ lies outside $\sigma_{\mathrm{ess}}(\Top)$ and is an isolated eigenvalue of finite multiplicity.
By the min--max principle of Appendix~\ref{app:sec:operator_spectrum}, an eigenvalue $z>\mu_j$ of multiplicity $m$ appears $m$ times in the list $\mu_1\ge\mu_2\ge\cdots$, at indices $i<j$.
The $m$ vectors $\varphi_i$ with these indices are orthonormal and satisfy $\Top\varphi_i=z\varphi_i$ by the hypothesis of the lemma, so they span the $m$-dimensional eigenspace of $z$.
Since $P_\Top$ vanishes on the gaps between these eigenvalues, $Q$ is the sum of their spectral projections, so its range is the sum of their eigenspaces, which lies in the span of $\varphi_1,\ldots,\varphi_{j-1}$ and is orthogonal to $W_j$.
For a unit $\psi\in W_j\cap D(\Top)$ with spectral measure $\nu_\psi$ under $\Top$, consider
\begin{align*}
\langle\psi|(\mu_j-\Top)\psi\rangle
\stackrel{(a)}{=}\int(\mu_j-z)\,d\nu_\psi(z)
\stackrel{(b)}{=}\int_{(-\infty,\mu_j]}(\mu_j-z)\,d\nu_\psi(z)
\stackrel{(c)}{\ge}0.
\end{align*}
The steps hold for the following reasons.
\begin{itemize}[leftmargin=*,itemsep=2pt,topsep=2pt]
\item (a) is the spectral theorem, with $\nu_\psi(\Real)=\|\psi\|^2=1$.
\item (b) holds since $\nu_\psi\bigl((\mu_j,\infty)\bigr)=\|Q\psi\|^2=0$, as $\psi\in W_j$ is orthogonal to the range of $Q$.
\item (c) holds since $\mu_j-z\ge0$ for $z\le\mu_j$.
\end{itemize}
Equality in (c) holds exactly when $\nu_\psi$ is concentrated at $\mu_j$, that is, when $\Top\psi=\mu_j\psi$.
\end{proof}

\subsubsection{Proof of Theorem~\ref{thm:lora_vmc_consistency}}

\begin{proof}[Proof of Theorem~\ref{thm:lora_vmc_consistency}]
Write $f_k=\sqrt{\eigvalparam_k}\,\psi_k/\|\psi_k\|$, so that $\Jc_\lora(\psi_{1:K},\eigvalparam_{1:K})=\Lc_\lora(f_{1:K})$ by the definition of $\Jc_\lora$ in Section~\ref{sec:nestedlora_vmc}.
The operator of the theorem is then $\sum_k\eigvalparam_k\tfrac{|\psi_k\rangle\langle\psi_k|}{\ip{\psi_k}{\psi_k}}=\sum_k|f_k\rangle\langle f_k|=A_\Vc$, with $A_\Vc$ as in Lemma~\ref{lem:galerkin_bound}.

\emph{Minimum.}
Lemma~\ref{lem:galerkin_bound} gives $\Lc_\lora(f_{1:K})\ge-\sum_k\mu_k^2$, and $f_k=\sqrt{\mu_k}\,\varphi_k$ for orthonormal eigenfunctions $\Top\varphi_k=\mu_k\varphi_k$, given by $\psi_k=\varphi_k$ and $\eigvalparam_k=\mu_k$, attains it, so a minimizer exists.

\emph{Form of every minimizer.}
Let $(\psi_{1:K},\eigvalparam_{1:K})$ be a minimizer, so that its $f_{1:K}$ attains the minimum and both inequalities of Lemma~\ref{lem:galerkin_bound} are tight: $\dim\Vc=K$, $\Top_\Vc$ has the eigenvalues $\mu_1,\ldots,\mu_K$ on $\Vc$, and $A_\Vc=(\Top_\Vc)_+$.
We show that $A_\Vc=\Top_\Vc$.
Let $\varphi_{1:K}$ be an orthonormal basis of $\Vc$ with $\Top_\Vc\varphi_j=\mu_j\varphi_j$; we show $\Top\varphi_j=\mu_j\varphi_j$ by induction on $j$.
If this holds for $\varphi_{1:j-1}$, then $\varphi_j$ is a unit vector in $W_j$ with $\ip{\varphi_j}{\Top\varphi_j}=\ip{\varphi_j}{\Top_\Vc\varphi_j}=\mu_j$, so $\Top\varphi_j=\mu_j\varphi_j$ by Lemma~\ref{lem:complement_bound}.
Since $\mu_k\ge\mu_K>0$, $\Top_\Vc=\sum_k\mu_k|\varphi_k\rangle\langle\varphi_k|$ is PSD, and $A_\Vc=(\Top_\Vc)_+=\Top_\Vc$.

\emph{Uniqueness.}
If $\mu_K>\mu_{K+1}$, at most $K$ orthonormal eigenfunctions have eigenvalues $\ge\mu_K$: for $K+1$ of them, $\ip{\psi}{\Top\psi}$ of a unit $\psi$ in their span is an average of their eigenvalues, hence $\ge\mu_K$, so the definition of $\mu_{K+1}$ gives $\mu_{K+1}\ge\mu_K$.
$\Vc$ is then the top-$K$ eigenspace, and $A_\Vc$ is the same for every minimizer.
If $\mu_K=\mu_{K+1}$, $\Vc$ contains the eigenfunctions with eigenvalues above $\mu_K$ and a subspace of $\ker(\Top-\mu_K)$ of the remaining dimension; the minimum and the eigenvalues of $A_\Vc$ do not depend on this choice, while $A_\Vc$ itself does.
\end{proof}

\subsubsection{Remarks}
\label{app:sec:lora_remarks}

\paragraph{Nonnegative scale parameters.}
The proof uses $\eigvalparam_k\ge0$ in step~(b) of Lemma~\ref{lem:galerkin_bound}, where $A_\Vc=\sum_k|f_k\rangle\langle f_k|$ is PSD.
Over $\eigvalparam_k<0$, $\Jc_\lora$ is unbounded below, since $\Top=-\hmltn$ is unbounded below.

\paragraph{Spin penalty.}
The spin penalty of Section~\ref{sec:exp_setup} replaces $\hmltn$ by $\hmltn+\hat S^2$; on $\Hc$ the total-spin operator $\hat S^2$ is bounded and commutes with $\hmltn$, so $\Top=-(\hmltn+\hat S^2)$ is self-adjoint and bounded above, its eigenfunctions are those of $\hmltn$ with each level raised by $S(S{+}1)$, and the theorem and Corollary~\ref{cor:nested_fixed_point} apply to it.

\paragraph{Relation to Theorem~C.5 of \citet{Ryu--Xu--Erol--Bu--Zheng--Wornell2024}.}
Theorem~\ref{thm:lora_vmc_consistency} holds for every self-adjoint $\Top$ bounded above under Eq.~\eqref{eq:consistency_hyp}, because it applies Eckart--Young to the finite-rank $\Top_\Vc$ (Lemma~\ref{lem:galerkin_bound}) and finds the eigenfunctions through the min--max values (Lemma~\ref{lem:complement_bound}).
Theorem~C.5 of \citet{Ryu--Xu--Erol--Bu--Zheng--Wornell2024} assumes an orthonormal basis of eigenfunctions of $\Top$.
The essential spectrum $(-\infty,-\Sigma]$ of $-\hmltn$ contains the continuous spectrum of the scattering states, whose solutions of the eigenvalue equation lie outside $\Hc$.
The eigenfunctions of $-\hmltn$ therefore span a proper subspace of $\Hc$ (Appendix~\ref{app:sec:spectral_hypothesis}).
A trial function $f_k$ has, in general, a component outside this subspace that an expansion of $f_k$ in eigenfunctions misses.
In exchange, Theorem~\ref{thm:lora_vmc_consistency} requires $\mu_K>0$, while Theorem~C.5 also covers an operator with only $r<K$ positive eigenvalues.

\paragraph{Scale parameters at a minimizer.}
With $f_k=\sqrt{\eigvalparam_k}\,\psi_k/\|\psi_k\|$ as in the proof, the objective depends on $f_{1:K}$ only through $A_\Vc=\sum_k|f_k\rangle\langle f_k|$.
For a minimizer with $A_\Vc=\sum_j\mu_j|\varphi_j\rangle\langle\varphi_j|$, the $f_k$ are exactly $f_k=\sum_{j=1}^K U_{jk}\sqrt{\mu_j}\,\varphi_j$ with $U$ a $K\times K$ unitary, because two families of $K$ vectors with the same rank-$K$ operator $\sum_k|f_k\rangle\langle f_k|$ differ by a unitary.
Then $\eigvalparam_k=\|f_k\|^2=\sum_j|U_{jk}|^2\mu_j$.
The joint minimum fixes the eigenvalues of $A_\Vc$ but not each $\eigvalparam_k$, which equals $\mu_k$ only if $U$ does not mix distinct eigenvalues.
Corollary~\ref{cor:nested_fixed_point} in Appendix~\ref{app:sec:nesting_fixed_point} shows that sequential nesting removes this freedom.

\subsection{Fixed point of sequential nesting}
\label{app:sec:nesting_fixed_point}

This section treats sequential nesting once more, through its objective.
The fixed point of sequential nesting is the ordered eigenpairs, up to a rotation within each degenerate level (Corollary~\ref{cor:nested_fixed_point}), while the joint minimum of Theorem~\ref{thm:lora_vmc_consistency} fixes the states only up to a rotation among all $K$ of them.
The sequential nesting gradient derived in Appendix~\ref{app:sec:proof_sequential_nesting} can be read as $K$ optimizations, one per network: network $k$ minimizes the order-$k$ objective $\Jc_\lora(\psi_{1:k},\eigvalparam_{1:k})$ over its own arguments, with the networks below it held fixed, and the corollary characterizes the fixed point of these $K$ optimizations.

If networks $1,\ldots,k-1$ hold the top $k-1$ eigenpairs, the order-$k$ objective minimized over $\eigvalparam_k$ depends on $\psi_k$ only through $q(\hat\psi_k)=\ip{\hat\psi_k}{\Top\hat\psi_k}-\sum_{\ell<k}\mu_\ell|\ip{\hat\psi_\ell}{\hat\psi_k}|^2$ with $\hat\psi_k=\psi_k/\|\psi_k\|$.
This $q$ is largest exactly at the eigenfunctions of eigenvalue $\mu_k$ orthogonal to the lower networks (Lemma~\ref{lem:complement_bound}).
Induction on $k$ then gives each network one eigenfunction and $\eigvalparam_k=\mu_k$.

\begin{corollary}[Fixed point of sequential nesting]
\label{cor:nested_fixed_point}
Assume Eq.~\eqref{eq:consistency_hyp}, and let $(\psi_{1:K},\eigvalparam_{1:K})$ be such that, for each $k=1,\ldots,K$, $(\psi_k,\eigvalparam_k)$ minimizes the order-$k$ objective $\Jc_\lora(\psi_{1:k},\eigvalparam_{1:k})$ over its own arguments with $(\psi_{1:k-1},\eigvalparam_{1:k-1})$ held fixed.
Then for every $k$, $\psi_k$ is an eigenfunction of $\Top$ with eigenvalue $\mu_k$, orthogonal to $\psi_{1:k-1}$, and $\eigvalparam_k=\mu_k$.
The statement also holds at $\mu_K=\mu_{K+1}$, and within a degenerate level $\psi_k$ may be any such eigenfunction.
\end{corollary}

\begin{proof}
By induction on $k$.
Suppose that $\hat\psi_\ell\defeq\psi_\ell/\|\psi_\ell\|$ for $\ell<k$ are orthonormal eigenfunctions of eigenvalue $\mu_\ell$ and that $\eigvalparam_\ell=\mu_\ell$, which is vacuous for $k=1$, and let $P$ be the orthogonal projection onto their span.
With $f_\ell=\sqrt{\eigvalparam_\ell}\,\hat\psi_\ell$, the order-$k$ objective is $-2\sum_{\ell\le k}\langle f_\ell|\Top f_\ell\rangle+\sum_{\ell,m\le k}|\langle f_\ell|f_m\rangle|^2$, and its terms with $f_k$ are $-2\langle f_k|\Top f_k\rangle$, the cross terms $2\sum_{\ell<k}|\langle f_\ell|f_k\rangle|^2$ and the diagonal term $\|f_k\|^4=\eigvalparam_k^2$.
Their sum is $-2\eigvalparam_k\,q(\hat\psi_k)+\eigvalparam_k^2$ with
\begin{align*}
q(\hat\psi)\defeq\langle\hat\psi|\Top\hat\psi\rangle-\sum_{\ell<k}\mu_\ell|\langle\hat\psi_\ell|\hat\psi\rangle|^2=\langle(1-P)\hat\psi|\Top(1-P)\hat\psi\rangle,
\end{align*}
where the last equality holds because $\Top$ maps the span of $\hat\psi_{1:k-1}$ into itself.
Over $\eigvalparam_k\ge0$ the minimum is $-q(\hat\psi_k)_+^2$, at $\eigvalparam_k=q(\hat\psi_k)_+$, so $\hat\psi_k$ maximizes $q$ over unit vectors.
For $q(\hat\psi_k)>0$ this $\eigvalparam_k$ is the analytic update of Eq.~\eqref{eq:analytic_eigvals} (Appendix~\ref{app:sec:analytic_eigvals}) with $\eigvalparam_\ell=\mu_\ell$ for $\ell<k$, so the corollary also covers $\eigvalparam_k$ set by that update.
By Lemma~\ref{lem:complement_bound}, $q(\hat\psi)\le\mu_k\|(1-P)\hat\psi\|^2\le\mu_k$, and since $\mu_k\ge\mu_K>0$ both bounds are attained exactly when $P\hat\psi=0$ and $\Top\hat\psi=\mu_k\hat\psi$; such a $\hat\psi$ exists because $\mu_{1:K}$ are counted with multiplicity.
Hence $\hat\psi_k$ is an eigenfunction of eigenvalue $\mu_k$ orthogonal to $\hat\psi_{1:k-1}$, and $\eigvalparam_k=\mu_k$.
\end{proof}

\section{Deferred proofs}
\label{app:sec:proofs}

This section derives the gradient expressions and the eigenvalue update that the main text states without proof.

\subsection{Gradient expressions}

\subsubsection{Proof of Eq.~\eqref{eq:vmc_grad} (VMC gradient)}
The gradient expression of the VMC Rayleigh quotient can be immediately derived from the following lemma.
\begin{lemma}\label{lem:grad_basic}
For a symmetric operator $\Oop$, 
\begin{align*}
\frac{\nabla_\th \ip{\psi_\th}{\Oop\psi_\th}}{\ip{\psi_\th}{\psi_\th}}
&= 2\Re\biggl(\E_{p_\th(\xv)}\biggl[
\frac{\Oop\psi_\th(\xv)}{\psi_\th(\xv)}
\nabla_\th \log\psi_\th^*(\xv)
\biggr]\biggr).
\end{align*}
\end{lemma}
\begin{proof}
The inner product between a vector-valued function and a scalar function denotes the coordinate-wise operation.
By the chain rule, the commutativity of $\Oop$ with $\nabla_\th$, and the symmetry of $\Oop$,

\begin{align*}
\nabla_\th \ip{\psi_\th}{\Oop\psi_\th}
= 2\Re\ip{\nabla_\th\psi_\th}{\Oop\psi_\th}
= 2\|\psi_\th\|^2 \Re \E_{p_\th(\xv)}\!\biggl[\frac{\Oop\psi_\th(\xv)}{\psi_\th(\xv)}\,\nabla_\th\log\psi_\th^*(\xv)\biggr],
\end{align*}

where the second equality multiplies and divides by $|\psi_\th(\xv)|^2/\|\psi_\th\|^2 = p_\th(\xv)$.
Dividing both sides by $\ip{\psi_\th}{\psi_\th} = \|\psi_\th\|^2$ gives the claim.
\end{proof}

\begin{proof}[Proof of Eq.~\eqref{eq:vmc_grad}]
Consider
\begin{align*}
\nabla_\th \Lc(\th)
&= \frac{\nabla_\th \ip{\psi_\th}{\hmltn\psi_\th}}{\ip{\psi_\th}{\psi_\th}}
-\frac{\ip{\psi_\th}{\hmltn\psi_\th} \nabla_\th \ip{\psi_\th}{\psi_\th}}{\ip{\psi_\th}{\psi_\th}^2}\\
&= \frac{\nabla_\th \ip{\psi_\th}{\hmltn\psi_\th}}{\ip{\psi_\th}{\psi_\th}}
-\Lc(\th) \frac{\nabla_\th \ip{\psi_\th}{\psi_\th}}{\ip{\psi_\th}{\psi_\th}}\\
&\stackrel{(a)}{=} 2\Re\biggl(
\E_{p_\th(\xv)}\Bigl[
\frac{\hmltn\psi_\th(\xv)}{\psi_\th(\xv)}
\nabla_\th \log\psi_\th^*(\xv)
\Bigr]
-\Lc(\th)\E_{p_\th(\xv)}\Bigl[
\nabla_\th \log\psi_\th^*(\xv)
\Bigr]
\biggr)\\
&= 2 \Re\Bigl(\E_{p_\th(\xv)}\Bigl[(\Ec_\th(\xv)-\E_{p_\th(\xv)}[\Ec_\th(\xv)]) \nabla_\th \log \psi_\th^*(\xv) \Bigr]\Bigr).
\end{align*}
Here, (a) follows from Lemma~\ref{lem:grad_basic}.
When $\psi_\th(\xv)\in\Real$ for any $\xv$, note that
\[
\nabla_\th\log\psi_\th^*(\xv)
=\nabla_\th(\log|\psi_\th(\xv)|+\log\sgn(\psi_\th(\xv)))
=\nabla_\th \log|\psi_\th(\xv)|+\nabla_\th \log\sgn(\psi_\th(\xv)).
\]
Here, taking $\log$ as the complex logarithm with its branch cut along the negative imaginary axis, $\nabla_\th\log\sgn(\psi_\th(\xv))$ is well defined and equals $0$ for almost every $\xv$, since $\sgn(\psi_\th(\xv))\in\{-1,1\}$ is locally constant in $\th$ wherever $\psi_\th(\xv)\neq0$.
Thus we can conclude $\nabla_\th\log\psi_\th^*(\xv)=\nabla_\th\log|\psi_\th(\xv)|$, and numerically we ignore this sign term and differentiate $\log|\psi_\th|$ alone.
This gives Eq.~\eqref{eq:vmc_grad} with $\nabla_\th\log|\psi_\th|$ in place of $\nabla_\th\log\psi_\th^*$ for a real-valued ansatz.
\end{proof}

\subsubsection{Derivation of the LoRA gradient}
\label{app:sec:unified_gradient_derivation}
\begin{lemma}
Define
\begin{align*}
\Gamma_{\kidx\ell}^{\Oop}
&\defeq \frac{\nabla_\th \ip{\psi_k}{\Oop\psi_\ell}}{\ip{\psi_k}{\psi_k}}.
\end{align*}
Then, we have
\begin{align}
\nabla_\th \Rm_{\kidx\ell}^{\Oop}
&= \Gamma_{\kidx\ell}^{\Oop} - \Rm_{k\ell}^{\Oop} \Gamma_{\kidx\kidx}^{\idop}.
\label{eq:generic_grad}
\end{align}
\end{lemma}
\begin{proof}
The quotient rule on $\Rm_{\kidx\ell}^{\Oop} = \ip{\psi_k}{\Oop\psi_\ell} / \ip{\psi_k}{\psi_k}$ gives $\nabla_\th \Rm_{\kidx\ell}^{\Oop} = \Gamma_{\kidx\ell}^{\Oop} - \Rm_{k\ell}^{\Oop}\,\Gamma_{\kidx\kidx}^{\idop}$.
\end{proof}

\begin{lemma}\label{lem:grad}
Define 
\begin{align*}
\Delta_{\kidx\ell}^{\Oop}
&\defeq \frac{\ip{\nabla_\th \psi_\kidx}{\Oop\psi_\ell}}{\ip{\psi_\kidx}{\psi_\kidx}}
= \E_{p_\kidx(\xv)}\biggl[
\frac{\Oop\psi_\ell(\xv)}{\psi_\kidx(\xv)}
\nabla_\th \log\psi_\kidx^*(\xv)
\biggr].
\end{align*}

Then, for a symmetric operator $\Oop$, we have
\begin{align*}
\Gamma_{\kidx\ell}^{\Oop}
= \Delta_{\kidx\ell}^{\Oop} + 
\frac{\|\psi_\ell\|^2}{\|\psi_\kidx\|^2} (\Delta_{\ell\kidx}^\Oop)^*
\end{align*}
In particular, when $\ell=\kidx$, we have $\Delta_{\kidx\kidx}^{\idop}=\gamma_k$ and
\[
\Gamma_{\kidx\kidx}^{\idop}=\gamma_\kidx+\gamma_\kidx^*,
\]
where we define
\begin{align*}
\gamma_{\kidx}&\defeq 
\E_{p_\kidx}\bigl[ \nabla_\th \log \psi_\kidx^*\bigr].
\end{align*}
\end{lemma}
\begin{proof}
The product rule gives $\nabla_\th\ip{\psi_\kidx}{\Oop\psi_\ell}=\ip{\nabla_\th\psi_\kidx}{\Oop\psi_\ell}+\ip{\psi_\kidx}{\Oop\nabla_\th\psi_\ell}$, and since $\Oop$ is symmetric the second term is $\ip{\nabla_\th\psi_\ell}{\Oop\psi_\kidx}^*$; dividing by $\ip{\psi_\kidx}{\psi_\kidx}$ gives the identity.
Setting $\ell=\kidx$ and $\Oop=\idop$ in the definition of $\Delta_{\kidx\ell}^{\Oop}$ gives $\Delta_{\kidx\kidx}^{\idop}=\gamma_\kidx$.
\end{proof}

The gradient identity used to assemble Eq.~\eqref{eq:nested_gradient} follows from the lemma below.
\begin{lemma}
\label{lem:grad_generic}
Define
\begin{align}
\label{eq:def:grad_cov}
\tilde{\Delta}_{\kidx\ell}^{\Oop}
&\defeq
\Cov_{p_\kidx}\biggl(\frac{\Oop \psi_{\ell}}{\psi_\kidx}, \nabla_\th \log \psi_\kidx^*\biggr)
= \Delta_{\kidx\ell}^{\Oop}
-\Rm_{\kidx\ell}^{\Oop} \gamma_\kidx.
\end{align}
Then,
\begin{align*}
\nabla_\th \Rm_{\kidx\kidx}^{\Top}
=\tilde{\Delta}_{\kidx\kidx}^{\Top}+(\tilde{\Delta}_{\kidx\kidx}^{\Top})^*
= 2\Re(\tilde{\Delta}_{\kidx\kidx}^{\Top}).
\end{align*}
For a symmetric operator $\Oop$, we have
\begin{align*}
\nabla_\th \Rm_{\kidx\ell}^{\Oop} \Rm_{\ell\kidx}^{\idop}
&= 
\Rm_{\kidx\ell}^{\Oop} \tilde{\Delta}_{\ell\kidx}^\idop
+ (\Rm_{\ell\kidx}^{\Oop}\tilde{\Delta}_{\kidx\ell}^\idop)^*
+\Rm_{\ell\kidx}^{\idop} \tilde{\Delta}_{\kidx\ell}^\Oop
+ (\Rm_{\kidx\ell}^{\idop}\tilde{\Delta}_{\ell\kidx}^\Oop)^*\\
&=2\Re \Bigl\{\Rm_{\kidx\ell}^{\Oop} \tilde{\Delta}_{\ell\kidx}^\idop
+\Rm_{\ell\kidx}^{\idop} \tilde{\Delta}_{\kidx\ell}^\Oop\Bigr\}\quad\text{if }\Oop=\idop.
\end{align*}
\end{lemma}

\begin{proof}
Continuing from Eq.~\eqref{eq:generic_grad},
\begin{align*}
\nabla_\th \Rm_{\kidx\ell}^{\Oop}
&= \Gamma_{\kidx\ell}^{\Oop} - \Rm_{k\ell}^{\Oop} \Gamma_{\kidx\kidx}^{\idop}\\
&= \Delta_{\kidx\ell}^{\Oop} 
+ \frac{\|\psi_\ell\|^2}{\|\psi_\kidx\|^2} (\Delta_{\ell\kidx}^\Oop)^*
- \Rm_{k\ell}^{\Oop} (\gamma_\kidx+\gamma_\kidx^*)\\
&= (\Delta_{\kidx\ell}^{\Oop} - \Rm_{k\ell}^{\Oop} \gamma_\kidx)
+ \frac{\|\psi_\ell\|^2}{\|\psi_\kidx\|^2} (
\Delta_{\ell\kidx}^\Oop
- \Rm_{\ell\kidx}^{\Oop} \gamma_\ell
)^*
+ \frac{\|\psi_\ell\|^2}{\|\psi_\kidx\|^2} (\Rm_{\ell\kidx}^{\Oop} \gamma_\ell)^*
-\Rm_{\kidx\ell}^{\Oop} \gamma_\kidx^*
\\
&= \tilde{\Delta}_{\kidx\ell}^\Oop
+ \frac{\|\psi_\ell\|^2}{\|\psi_\kidx\|^2} 
(\tilde{\Delta}_{\ell\kidx}^\Oop)^*
+ \frac{\|\psi_\ell\|^2}{\|\psi_\kidx\|^2} (\Rm_{\ell\kidx}^{\Oop} \gamma_\ell)^*
-\Rm_{\kidx\ell}^{\Oop} \gamma_\kidx^*.
\end{align*}

This leads to:
\begin{align*}
\Rm_{\kidx\ell}^{\Oop} 
\nabla_\th \Rm_{\ell\kidx}^{\idop}
&= 
\Rm_{\kidx\ell}^{\Oop} \tilde{\Delta}_{\ell\kidx}^\idop
+ (\Rm_{\ell\kidx}^{\Oop}\tilde{\Delta}_{\kidx\ell}^\idop)^*
+ (\Rm_{\ell\kidx}^{\Oop}\Rm_{\kidx\ell}^{\idop} \gamma_\kidx)^*
-\Rm_{\kidx\ell}^{\Oop} \Rm_{\ell\kidx}^{\idop} \gamma_\ell^*,\\
\Rm_{\ell\kidx}^{\idop} 
\nabla_\th \Rm_{\kidx\ell}^{\Oop}
&= 
\Rm_{\ell\kidx}^{\idop} \tilde{\Delta}_{\kidx\ell}^\Oop
+ (\Rm_{\kidx\ell}^{\idop}\tilde{\Delta}_{\ell\kidx}^\Oop)^*
+ (\Rm_{\kidx\ell}^{\idop}
\Rm_{\ell\kidx}^{\Oop} \gamma_\ell)^*
-\Rm_{\ell\kidx}^{\idop} \Rm_{\kidx\ell}^{\Oop} \gamma_\kidx^*.
\end{align*}
Summing them up, we get
\begin{align*}
\nabla_\th \Rm_{\kidx\ell}^{\Oop} \Rm_{\ell\kidx}^{\idop}
&= 
\Rm_{\kidx\ell}^{\Oop} \nabla_\th \Rm_{\ell\kidx}^{\idop}
+\Rm_{\ell\kidx}^{\idop} \nabla_\th \Rm_{\kidx\ell}^{\Oop} \\
&= 
\Bigl(
\Rm_{\kidx\ell}^{\Oop} \tilde{\Delta}_{\ell\kidx}^\idop
+ (\Rm_{\ell\kidx}^{\Oop}\tilde{\Delta}_{\kidx\ell}^\idop)^*
+\Rm_{\ell\kidx}^{\idop} \tilde{\Delta}_{\kidx\ell}^\Oop
+ (\Rm_{\kidx\ell}^{\idop}\tilde{\Delta}_{\ell\kidx}^\Oop)^*\Bigr)\\
&\quad+ 
\Bigl(
(\Rm_{\ell\kidx}^{\Oop}\Rm_{\kidx\ell}^{\idop} )^*- \Rm_{\ell\kidx}^{\idop} \Rm_{\kidx\ell}^{\Oop}\Bigr)
(\gamma_\kidx^*+\gamma_\ell^*)
\\
&= 
\Rm_{\kidx\ell}^{\Oop} \tilde{\Delta}_{\ell\kidx}^\idop
+ (\Rm_{\ell\kidx}^{\Oop}\tilde{\Delta}_{\kidx\ell}^\idop)^*
+\Rm_{\ell\kidx}^{\idop} \tilde{\Delta}_{\kidx\ell}^\Oop
+ (\Rm_{\kidx\ell}^{\idop}\tilde{\Delta}_{\ell\kidx}^\Oop)^*.
\end{align*}
This concludes the proof.
\end{proof}

\subsubsection{Final gradient expression for implementation}
\label{app:sec:final_custom_gradient}
Recall that $\tilde{\Delta}^{\Oop}$ is defined in Eq.~\eqref{eq:def:grad_cov}.
For simplicity, we assume real ansatzes without loss of generality.
Let $\th_k$ denote the parameters of network $k$; with $\nabla_{\th_k}$ in place of $\nabla_\th$ in Eq.~\eqref{eq:def:grad_cov}, $\tilde{\Delta}_{ij}^{\Oop}=0$ for $i\neq k$.
\begin{theorem}[LoRA gradient]
\label{thm:lora_grad}
\begin{align*}
\begin{cases}
\displaystyle
\nabla_{\th_k} \Jc_{\lora}(\psi_{1:k},\eigvalparam_{1:k})
&
\displaystyle
= -4\eigvalparam_k\tilde{\Delta}_{kk}^{\Top}
+ 4\sum_{\ell=1}^k
\eigvalparam_k\eigvalparam_{\ell}
\Rm_{\ell k}^{\idop}\tilde{\Delta}_{k\ell}^{\idop},\\
\displaystyle
\partial_{\eigvalparam_k} \Jc_{\lora}(\psi_{1:k},\eigvalparam_{1:k})
&
\displaystyle
= -2\Rm_{\kidx\kidx}^{\Top}
+ 2\sum_{\ell=1}^k
\eigvalparam_\ell
\Rm_{\ell k}^{\idop}\Rm_{k\ell}^{\idop}.
\end{cases}
\end{align*}
\end{theorem}
Sequential nesting applies this gradient for each $k$.

\paragraph{Surrogate objective for autograd.}
To compute the centered gradient of Eq.~\eqref{eq:def_g} with a generic autograd, we differentiate a surrogate objective whose derivative gives this gradient.
The surrogate, whose autograd derivative is twice the centered gradient, is
\begin{align}
\label{eq:unified_surrogate_obj}
\breve{\Lc}_{\lora}(\psi_{1:K},\eigvalparam_{1:K}) \defeq 2 \sum_{k=1}^K \E_{p_k}\!\bigl[\stopgrad(g_k - \E_{p_k}[g_k])\,\log |\psi_k|\bigr],
\end{align}
where $\stopgrad$ is the stop-gradient operator.
The surrogate depends on $\theta$ only through $\log|\psi_k|$, in which it is linear, so a generic autograd differentiates it directly.
In $g_k$, $\Rm_{\ell k}^{\idop}$ is estimated by its batch mean.

\subsubsection{Sequential nesting identity}
\label{app:sec:proof_sequential_nesting}

We state the lemma that gives the overlap term of Eq.~\eqref{eq:nested_gradient} in terms of the covariance
\begin{align*}
\tilde{\Delta}_{k\ell}^{\Oop} \defeq \Cov_{p_k}\!\bigl(\Oop\psi_\ell / \psi_k,\, \nabla_\th \log \psi_k^*\bigr) \qquad \text{for } \Oop \in \{\Top, \idop\}.
\end{align*}

\begin{lemma}[Sequential nesting, LoRA case]
\label{lem:sequential_nesting}
The gradient contribution from the second term of $\Jc_{\lora}(\psi_{1:k},\eigvalparam_{1:k})$, without its weights $\eigvalparam_i\eigvalparam_j$, summed over $k$ is
\begin{align*}
\sum_{k=1}^K \bigl\langle \nabla_\th\psi_k~\big|~\partial_{\psi_k}\textstyle\sum_{i,j=1}^k \Rm_{ij}^{\idop} \Rm_{ji}^{\idop}\bigr\rangle
= 4\sum_{k,\ell=1}^K \matmask_{k\ell}^\seq\,\Rm_{\ell k}^{\idop}\,\tilde{\Delta}_{k\ell}^{\idop}.
\end{align*}
\end{lemma}

\begin{proof}
For $\ell\neq k$, the quotient rule gives

\begin{align*}
\frac{\partial\Rm_{k\ell}^{\Oop}}{\partial\psi_k(\xv)}
=\frac{\psi_k(\xv)}{\|\psi_k\|^2}\biggl(\frac{\Oop\psi_\ell(\xv)}{\psi_k(\xv)}-2\Rm_{k\ell}^\Oop\biggr).
\end{align*}

Therefore, using $\nabla_\th\psi_k/\psi_k=\nabla_\th\log|\psi_k|$ on the support of $p_k$,

\begin{align*}
\ip{\nabla_\th\psi_k}{\partial_{\psi_k}\Rm_{k\ell}^{\Oop}}
&= \int \nabla_\th\psi_k(\xv)\,\frac{\partial\Rm_{k\ell}^{\Oop}}{\partial\psi_k(\xv)} \diff\xv
= \E_{p_k(\xv)}\biggl[\biggl(\frac{\Oop\psi_\ell(\xv)}{\psi_k(\xv)}-2\Rm_{k\ell}^\Oop\biggr)\frac{\nabla_\th\psi_k(\xv)}{\psi_k(\xv)}\biggr] \\
&= \Delta_{k\ell}^{\Oop}-2\gamma_k \Rm_{k\ell}^{\Oop} = \tilde{\Delta}_{k\ell}^{\Oop}-\gamma_k \Rm_{k\ell}^{\Oop}.
\end{align*}

For the second factor $\Rm_{\ell k}^{\idop}$, for $\ell\neq k$, $\ip{\nabla_\th\psi_k}{\partial_{\psi_k}\Rm_{\ell k}^{\idop}} = (\|\psi_k\|^2/\|\psi_\ell\|^2)\,\Delta_{k\ell}^{\idop}$.
Combining both, applying the identities $(\|\psi_k\|^2/\|\psi_\ell\|^2)\Rm_{k\ell}^{\Oop}=\Rm_{\ell k}^{\Oop}$ and $\Rm_{k\ell}^{\Oop}\Rm_{\ell k}^{\idop}=\Rm_{\ell k}^{\Oop}\Rm_{k\ell}^{\idop}$, and using $\Delta - \gamma R = \tilde\Delta$, when $\ell \neq k$:

\begin{align*}
\ip{\nabla_\th\psi_k}{\partial_{\psi_k}\Rm_{k\ell}^{\Oop}\Rm_{\ell k}^{\idop}}
&= \Rm_{\ell k}^{\idop}\tilde{\Delta}_{k\ell}^{\Oop} - \gamma_k \Rm_{k\ell}^{\Oop}\Rm_{\ell k}^{\idop} + \Rm_{k\ell}^{\Oop}\,(\|\psi_k\|^2/\|\psi_\ell\|^2)\,\Delta_{k\ell}^{\idop} \\
&= \Rm_{\ell k}^{\idop}\tilde{\Delta}_{k\ell}^{\Oop} + \Rm_{\ell k}^{\Oop}(\Delta_{k\ell}^{\idop} - \gamma_k \Rm_{k\ell}^{\idop})
= \Rm_{\ell k}^{\Oop} \tilde{\Delta}_{k\ell}^\idop + \Rm_{\ell k}^{\idop} \tilde{\Delta}_{k\ell}^\Oop.
\end{align*}

In $\sum_{i,j=1}^k \Rm_{ij}^{\Oop}\Rm_{ji}^{\idop}$, each $\ell<k$ appears in the two terms $(i,j)=(k,\ell)$ and $(\ell,k)$, which are equal by the second identity.
The term $(k,k)$ is $\Rm_{kk}^{\Oop}$, since $\Rm_{kk}^{\idop}=1$, and the quotient rule gives $\ip{\nabla_\th\psi_k}{\partial_{\psi_k}\Rm_{kk}^{\Oop}}=2\tilde\Delta_{kk}^{\Oop}$, which is the $\ell=k$ summand below since $\tilde\Delta_{kk}^{\idop}=0$.
Hence,

\begin{align*}
\biggl\langle \nabla_\th\psi_k~\bigg|~ \partial_{\psi_k}\sum_{i,j=1}^k \Rm_{ij}^{\Oop} \Rm_{ji}^{\idop}\biggr\rangle
= 2\sum_{\ell=1}^k (\Rm_{\ell k}^{\Oop} \tilde{\Delta}_{k\ell}^\idop + \Rm_{\ell k}^{\idop} \tilde{\Delta}_{k\ell}^\Oop).
\end{align*}
At $\Oop=\idop$ the bracket is $2\Rm_{\ell k}^{\idop}\tilde\Delta_{k\ell}^{\idop}$ and vanishes at $\ell=k$, so summing over $k$ gives the lemma.
\end{proof}

\subsection{Derivation of the analytic eigenvalue update in Eq.~\eqref{eq:analytic_eigvals}}
\label{app:sec:analytic_eigvals}

We derive the closed form of Eq.~\eqref{eq:analytic_eigvals} step by step.
For each $k\in[K]$, the partial derivative of the LoRA objective with respect to $\eigvalparam_k$ at the sequentially nested level is
\begin{align*}
\frac{\partial}{\partial \eigvalparam_k}
\Jc_{\lora}(\psi_{1:k},\eigvalparam_{1:k})
&=
-2\Rm_{kk}^\Top
+ 2\sum_{\ell=1}^k \eigvalparam_\ell \Rm_{k\ell}^{\idop}\Rm_{\ell k}^{\idop}.
\end{align*}
Setting this to zero,
\begin{align*}
\eigvalparam_k^\star
=\Rm_{kk}^\Top - \sum_{\ell=1}^{k-1}\eigvalparam_\ell^\star \Rm_{k\ell}^{\idop}\Rm_{\ell k}^{\idop}.
\end{align*}
Solving sequentially in $k$ yields
\begin{align*}
\eigvalparam_1^\star&=\Rm_{11}^{\Top},\\
\eigvalparam_2^\star&=\Rm_{22}^{\Top}-\eigvalparam_1^\star\Rm_{21}^{\idop}\Rm_{12}^{\idop},\\
&\vdots\\
\eigvalparam_K^\star&=\Rm_{KK}^{\Top}-\sum_{\ell=1}^{K-1}\eigvalparam_\ell^\star \Rm_{K\ell}^{\idop}\Rm_{\ell K}^{\idop}.
\end{align*}
In matrix-vector form,
\begin{align*}
\boldsymbol{\eigvalparam}^\star&=\diag(\Rm^{\Top})-\begin{bmatrix}
0\\
\eigvalparam_1^\star\Rm_{21}^{\idop}\Rm_{12}^{\idop}\\
\vdots\\
\eigvalparam_1^\star\Rm_{K,1}^{\idop}\Rm_{1,K}^{\idop}+\ldots+\eigvalparam_{K-1}^\star\Rm_{K,K-1}^{\idop}\Rm_{K-1,K}^{\idop}
\end{bmatrix}\\
&= \diag(\Rm^{\Top})
-\bigl(\matmask^{\seq}\odot(\Rm^{\idop}\odot(\Rm^{\idop})^\intercal)\bigr)\boldsymbol{\eigvalparam}^\star,
\end{align*}
which rearranges to the closed form quoted in the main text:
\begin{align*}
\boldsymbol{\eigvalparam}^\star
&= \bigl(\Imat + \matmask^{\seq}\odot(\Rm^{\idop}\odot(\Rm^{\idop})^\intercal)\bigr)^{-1}\diag(\Rm^{\Top}).
\end{align*}
In our implementation, $\Rm^{\Top}$ and $\Rm^{\idop}$ in this solve are estimated by exponential moving averages (EMAs) of their per-step batch means, which smooths $\boldsymbol{\eigvalparam}^\star$ across steps.
The EMA is warm-started: at step $1$ it takes the first batch statistic itself rather than blending against an arbitrary prior $\hat x_0$, so the smoothing starts at step $2$.
The gradient coefficient $g_k$ of Eq.~\eqref{eq:def_g} uses the per-step batch means.

\section{Monte Carlo statistics}
\label{app:sec:mc_statistics}

This section gives the estimates of energies and overlaps that Section~\ref{sec:implementation} takes from the walkers.

\paragraph{Averages over steps.}
An exponential moving average (EMA) over training steps $t$ of a batch statistic $x_t$ is $m_t=\alpha_t x_t+(1-\alpha_t)\,m_{t-1}$.
The inputs $\Rm^{\Top}$ and $\Rm^{\idop}$ of the eigenvalue solve (Appendix~\ref{app:sec:analytic_eigvals}) and the logarithm of the norm-ratio estimate (Appendix~\ref{app:sec:overlap_statistics}) are EMAs with $\alpha_t=0.01$, started at $m_1=x_1$.
The energies that rank the networks (Appendix~\ref{app:sec:ewm_ordering}) are EMAs with the schedule $\alpha_t=\max(10^{-3},\,1/(2+t/10))$ of DeepQMC~\citep{deepqmc}, which falls from $1/2$ to $10^{-3}$ within the first ${\sim}10^4$ steps.

\subsection{Energy statistics}
\label{app:sec:energy_statistics}

\paragraph{Clipping.} The local energy in $\tilde{\Delta}_{kk}^{\Top}$, $x = \Ec_k(\xv)$ plus $\hat S^2_{\mathrm{loc}}(\xv)$ when the spin penalty is on (Appendix~\ref{app:sec:spin_penalty_impl}), is clipped to the window $[m - w\,d,\, m + w\,d]$ around its batch median $m$, with $w = 10$ and $d \defeq \E[|x - m|]$.
The clipped values enter the energy term of $g_k$ and the energy input of the eigenvalue solver.

\subsection{Overlap statistics}
\label{app:sec:overlap_statistics}

The ratio $\psi_\ell(\xv)/\psi_k(\xv)$ in $\tilde{\Delta}_{k\ell}^{\idop}$ enters training unclipped, because a clip of this heavy-tailed ratio biases the overlap estimate (Appendices~\ref{app:sec:overlap_heavy_tail} and~\ref{app:sec:overlap_measurements}).
The coefficient $\Rm^{\idop}_{\ell k}$ of $g_k$ in Eq.~\eqref{eq:def_g} is the unclipped batch mean of the reverse ratio $\psi_k/\psi_\ell$ on the walkers of $\psi_\ell$, and the eigenvalue solve of Eq.~\eqref{eq:analytic_eigvals} takes the EMA of the same unclipped batch means as $\Rm^{\idop}$.
Before these means are formed, each ratio is divided by the square root of an estimate of the norm ratio $\ip{\psi_\ell}{\psi_\ell}/\ip{\psi_k}{\psi_k} = \E_{p_k}[(\psi_\ell/\psi_k)^2]$, obtained by exponentiating an EMA (Appendix~\ref{app:sec:mc_statistics}) of the logarithm of its batch mean.
This division makes the batch mean of a ratio an estimate of the normalized overlap $\tilde{\Sm}_{k\ell}$.
The two ratios of a pair are rescaled by estimates from different chains, and the overlap term of $g_k$ multiplies one rescaled ratio by the batch mean of the other, so this term is multiplied by a factor that is one at the exact norm ratios; the estimate could be replaced by the iterative bridge sampling of \citet{Gao--Grutschus--Noe--Guennemann2026}, or the rescaling removed.

\subsubsection{Heavy-tail behavior of the overlap-term ratio at nodal surfaces}
\label{app:sec:overlap_heavy_tail}

The ratio $\psi_\ell(\xv)/\psi_k(\xv)$ at $\xv \sim p_k$ has a heavy tail near the nodes of $\psi_k$, and it is the per-sample integrand of the overlap term of $g_k$ in Eq.~\eqref{eq:def_g}.
At a node of $\psi_k$, $|\psi_k|$ vanishes while $\psi_\ell$ is generically nonzero, so the ratio diverges for a walker close to a node of the denominator, while its mean $\Rm_{k\ell}^{\idop}$ stays finite.

\paragraph{Closed-form example: Hydrogen $2p_z$ vs $2p_x$.}
The simplest non-trivial pair already shows the heavy tail: take $\psi_k = \psi_{2p_z}$ as the denominator and $\psi_\ell = \psi_{2p_x}$ as the numerator on Hydrogen.
Both have the same radial factor, which cancels, so the overlap ratio reduces to the angular ratio
\begin{align*}
R(\xv) \;=\; \frac{\psi_{2p_x}(\xv)}{\psi_{2p_z}(\xv)} \;=\; \tan\theta \cdot \cos\phi,
\end{align*}
sampled at $\xv \sim |\psi_{2p_z}|^2$.
Setting $u = \cos\theta$, the squared density factorizes as $\frac{3}{2} u^2 \,\mathrm{d}u \cdot \frac{1}{2\pi}\,\mathrm{d}\phi$, so the polar and azimuthal factors are independent.
The polar tail is $\Pr(|\tan\theta| > w) = (1+w^2)^{-3/2}$ exactly.
Integrating over $\phi$ and differentiating gives the closed-form density
\begin{align}
f_R(r) \;=\; \frac{2}{\pi}\,\frac{1}{(1 + r^2)^2}, \qquad r \in \mathbb{R},
\label{eq:overlap_density}
\end{align}
which (after the rescaling $x = \sqrt{3}\,r$) is exactly Student's $t$ distribution with $\nu = 3$ degrees of freedom.
The tail decays as $\Pr(|R| > t) \sim 4 / (3\pi t^3)$ as $t \to \infty$, and consequently $\E[|R|^p] < \infty$ if and only if $p < 3$: the per-sample overlap estimator has a finite mean and a finite variance, but its third absolute moment diverges logarithmically.

\paragraph{General nodal sets.}
\begin{proposition}\label{prop:nodal_tail}
Assume, for some $M,\eta>0$, that
\begin{enumerate}[label=(\roman*),leftmargin=*,itemsep=1pt,topsep=2pt]
\item $\psi_k$ is real and continuously differentiable,
\item $|\psi_\ell|\le M$,
\item $\nabla\psi_k\neq 0$ wherever $|\psi_k|<\eta$, and
\item for $u,v\in\mathbb{R}$ the function $\Gamma(u,v)\defeq\int_{\{\psi_k=u\}}\mathbf{1}\{|\psi_\ell|>|v|\}\,|\nabla\psi_k|^{-1}\,\mathrm{d}\sigma$ satisfies $\Gamma(u,v)\to\Gamma(0,v)$ as $u\to 0$ for almost every $v$, and $\Gamma(u,v)\le\bar\Gamma(v)$ for $|u|<\eta$ with $\int v^2\,\bar\Gamma(v)\,\mathrm{d}v<\infty$.
\end{enumerate}
Then $R=\psi_\ell/\psi_k$ at $\xv\sim p_k$ satisfies
\begin{align}
\lim_{t\to\infty} t^3\Pr(|R|>t) \;=\; \frac{2}{3\ip{\psi_k}{\psi_k}}\int_{\{\psi_k=0\}}\frac{|\psi_\ell|^3}{|\nabla\psi_k|}\,\mathrm{d}\sigma .
\label{eq:nodal_tail}
\end{align}
\end{proposition}
\begin{proof}
For $t>M/\eta$ the event $|R|>t$, that is $|\psi_k|<|\psi_\ell|/t$, lies in $\{|\psi_k|<\eta\}$, where the coarea formula gives $\Pr(|R|>t)=\ip{\psi_k}{\psi_k}^{-1}\int_{-\eta}^{\eta}u^2\,\Gamma(u,tu)\,\mathrm{d}u$.
The substitution $u=v/t$ gives $t^3\Pr(|R|>t)=\ip{\psi_k}{\psi_k}^{-1}\int_{-\eta t}^{\eta t}v^2\,\Gamma(v/t,v)\,\mathrm{d}v$, which tends to $\ip{\psi_k}{\psi_k}^{-1}\int v^2\,\Gamma(0,v)\,\mathrm{d}v$ by dominated convergence.
Exchanging the two integrals, each point of $\{\psi_k=0\}$ contributes $\int_{|v|<|\psi_\ell|}v^2\,\mathrm{d}v=\tfrac{2}{3}|\psi_\ell|^3$, which gives Eq.~\eqref{eq:nodal_tail}.
\end{proof}
For the pair of Eq.~\eqref{eq:overlap_density}, with $\psi_k=\psi_{2p_z}=z\,e^{-r/2}$ and $\psi_\ell=\psi_{2p_x}=x\,e^{-r/2}$, the nodal set is the plane $z=0$, the integral in Eq.~\eqref{eq:nodal_tail} is $64$ and $\ip{\psi_k}{\psi_k}=32\pi$, so the limit is the constant $4/(3\pi)$ found above.
The exponent $3$ comes from assumptions (i) and (iii), under which $\psi_k$ changes sign linearly across the hypersurface $\{\psi_k=0\}$.
A complex $\psi_k$, which vanishes on a set of codimension two, gives $t^{-4}$ instead.
A crossing of two nodal surfaces, where $\nabla\psi_k=0$, gives $t^{-3}\log t$.
Assumption (iv) excludes a heavier tail from the region where one electron is far from the nuclei.
No other property of the two functions enters, so the $t^{-3}$ tail applies to molecular wavefunctions that satisfy these conditions.
Under the same conditions, the square $(\psi_\ell/\psi_k)^2$ in the norm-ratio estimate and, where $\nabla_\th\psi_k\neq0$ on the node, the product of the ratio with the score in Eq.~\eqref{eq:def_g} have a finite mean and an infinite variance.

\paragraph{Distributions at convergence.}
The local energy and the ratio differ in how their distributions behave as the networks converge.
At an eigenstate the local energy is constant, $\Ec_k(\xv)=E_k$ for every $\xv$, so its distribution under $p_k$ shrinks to a point, every sample falls inside the clip window, and the energy clip does not act.
At the eigenfunctions the ratio $R=\varphi_\ell/\varphi_k$ keeps the tail $\Pr(|R|>t)\sim t^{-3}$ of Proposition~\ref{prop:nodal_tail} near every node of $\varphi_k$.

\subsubsection{Measurements on trained networks}
\label{app:sec:overlap_measurements}

On trained networks, a clip biases the batch means of the overlap statistics by much more than it reduces their spread (Section~\ref{sec:clipping}).
The networks trained with the unclipped ratio are close to orthogonal (Figure~\ref{fig:overlap_x2}).

\paragraph{Long chains.}
Two long Metropolis chains for each pair of networks $\psi_k$ and $\psi_\ell$, run on saved checkpoints, give the reference value of $\E_{p_k}[\psi_\ell/\psi_k]$ and the normalized overlap of the pair.
The chains sample the density proportional to $\rho=|\psi_k|^2+e^{-a}|\psi_\ell|^2$, so they cross the nodes of $\psi_k$ more often than the chain of $\psi_k$.
With $\hat\E_\rho[f]$ the mean of $f$ over their walkers, we estimate
\begin{align*}
\E_{p_k}\!\left[\frac{\psi_\ell}{\psi_k}\right]=\frac{\ip{\psi_k}{\psi_\ell}}{\|\psi_k\|^2}&\approx\frac{\hat\E_\rho[\psi_k\psi_\ell/\rho]}{\hat\E_\rho[|\psi_k|^2/\rho]},\\
\frac{\ip{\psi_k}{\psi_\ell}}{\|\psi_k\|\,\|\psi_\ell\|}&\approx\frac{\hat\E_\rho[\psi_k\psi_\ell/\rho]}{\sqrt{\hat\E_\rho[|\psi_k|^2/\rho]\,\hat\E_\rho[|\psi_\ell|^2/\rho]}}.
\end{align*}
Since the walkers follow $\rho/Z_\rho$ with $Z_\rho=\int\rho$, $\hat\E_\rho[f/\rho]$ tends to $Z_\rho^{-1}\int f$, so $Z_\rho$ cancels in each ratio and every $a$ gives the same limit.
The constant $a$ sets only the share of walkers on $|\psi_k|^2$, $\|\psi_k\|^2/(\|\psi_k\|^2+e^{-a}\|\psi_\ell\|^2)$.
Optimally, $e^{a}$ is the norm ratio $\|\psi_\ell\|^2/\|\psi_k\|^2$ estimated from shorter chains, which makes this share about $1/2$.

\begin{figure}[t]
  \centering
  \includegraphics[width=\linewidth]{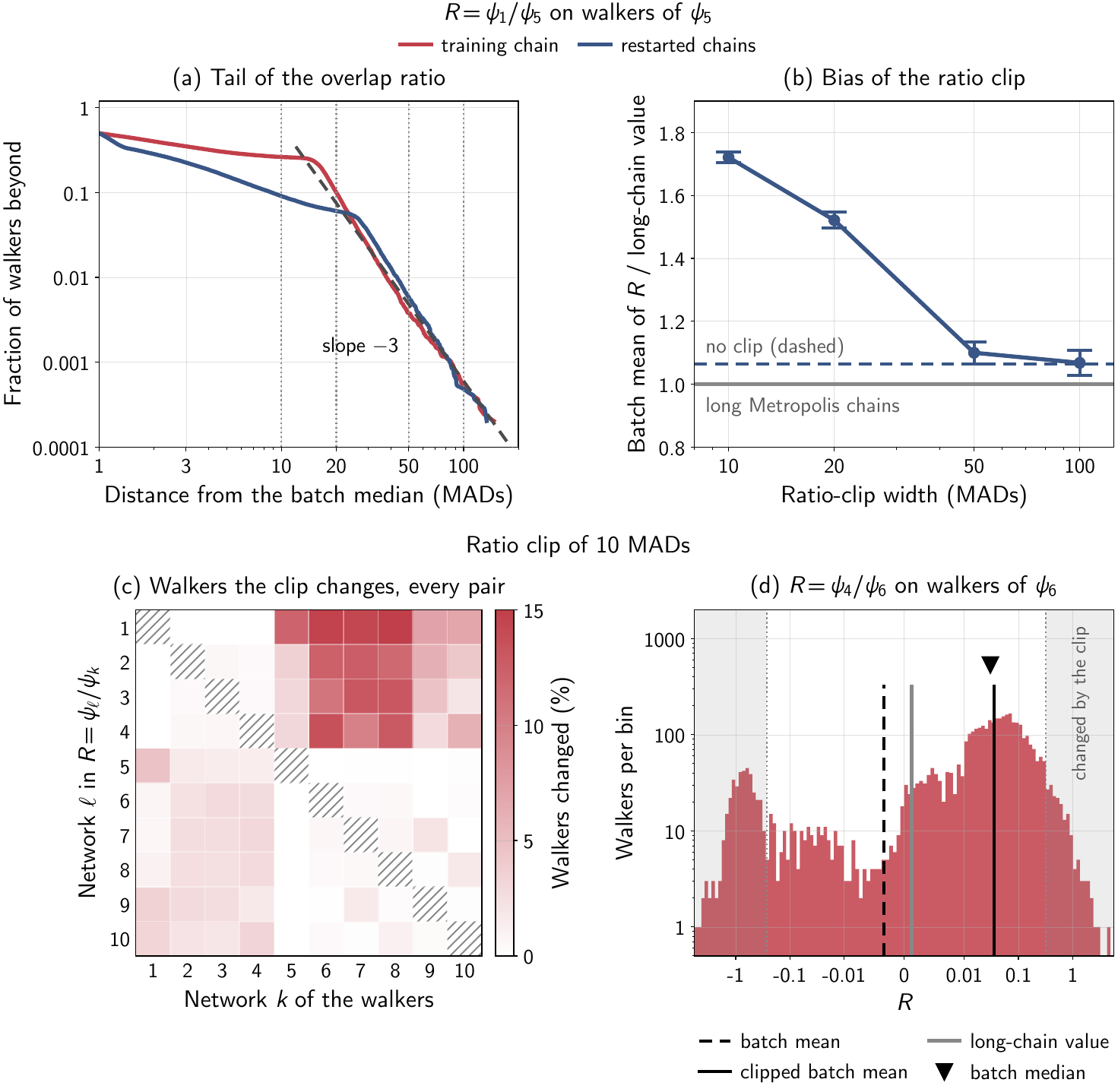}
  \caption{Overlap ratio $R=\psi_\ell/\psi_k$ of networks $k$ and $\ell$ of the Li ansatz, on the walkers of $\psi_k$. Panels (a) and (b) are at training step $1{,}000$ and number the networks by ascending energy at that step; panels (c) and (d) are at step $100{,}000$ and number them as in Table~\ref{tab:per_state_master}. Dotted lines in (a) mark the clip widths of (b); error bars in (b) are standard errors over six batches; the axis of $R$ in (d) is logarithmic beyond $\pm0.01$. Panel (b) uses the restarted chains (Appendix~\ref{app:sec:overlap_measurements}).}
  \label{fig:overlap_clip_bias}
\end{figure}

\paragraph{Bias of the clip (Figure~\ref{fig:overlap_clip_bias}a\&~\ref{fig:overlap_clip_bias}b).}
Clipping $R$ at a fixed number of median absolute deviations (MADs) strongly biases its batch mean; the bias remains at a width of $50$ MADs and is gone only at $100$ MADs (Figure~\ref{fig:overlap_clip_bias}b).
Figure~\ref{fig:overlap_clip_bias}b shows that the bias of the clip is many times the spread across batches, even though the clip lowers this spread by removing the tail of $R$.
The clipped batch mean is thus a precise estimate of a shifted value.
For the reverse ratio $\psi_5/\psi_1$ on the chain of $\psi_1$, the unclipped batch mean lies near the reference.
The ratio clip lowers it to half of it or less.

We provisionally apply the ratio clip of the DeepQMC code~\citep{deepqmc} at $10$ MADs to $R=\psi_1/\psi_5$ at step $1{,}000$ of a Li run.
The fraction of walkers beyond $s$ MADs in Figure~\ref{fig:overlap_clip_bias}a is the fraction a ratio clip of width $s$ changes.
The walkers of $\psi_5$ come from its \emph{training chain} and from \emph{restarted chains}, which cross the nodes of $\psi_5$ more often.
The two long chains of this pair use $a=0$, which gives the same limit, with $4{,}096$ walkers each and $45{,}000$ steps; their walkers, resampled to $p_5$ and then moved by $300$ Metropolis steps on $p_5$, start the restarted chains.
Without a clip, the batch mean agrees with the long-chain reference within its spread across batches on the restarted chains and has the opposite sign on the training chain, so Figure~\ref{fig:overlap_clip_bias}b shows the restarted chains only.

\paragraph{The ratio at the end of training (Figure~\ref{fig:overlap_clip_bias}c\&~\ref{fig:overlap_clip_bias}d).}
The ratio $R=\psi_\ell/\psi_k$ keeps a heavy, power-law tail to the end of training, as the analysis of Appendix~\ref{app:sec:overlap_heavy_tail} supports.
At step $100{,}000$ of the seed-$1$ Li run, we measure the tail of each of the $90$ ratios by the fraction of the $4{,}096$ training walkers of $\psi_k$ at which $R$ lies more than $s$ MADs from its median.
Proposition~\ref{prop:nodal_tail} gives this fraction an asymptotic log-log slope of $-3$ in $s$; the slopes fitted to the $90$ ratios have median $-2.5$ and quartiles $-3.0$ and $-1.9$.

At the end of training, where the overlaps should vanish, the ratio clip at $10$ MADs still changes the walkers of many ratios (Figure~\ref{fig:overlap_clip_bias}c) and biases their batch means.
The clip window is centered on the batch median, so the clip removes more walkers on the heavier side of the tail and moves the batch mean toward the median (Figure~\ref{fig:overlap_clip_bias}d).
For $R=\psi_4/\psi_6$, the ratio whose batch mean the clip moves farthest relative to its long-chain value, the clip moves the batch mean from $-0.0033$ to $+0.0354$, close to the median $+0.0302$, while the long chains give $+0.0012\pm0.0006$.

\paragraph{Overlaps of the trained networks.}
The near-orthogonality of the trained networks in Figure~\ref{fig:overlap_x2} of Section~\ref{sec:exp_overlap} is measured on long chains, with two chains for each pair that repeat the same measurement as a check on each other.
The figure covers the pairs within a spin sector at the final step of the seed-$1$ runs of Li to F, since pairs in different sectors are orthogonal by spin symmetry.
Each chain has $2{,}048$ walkers and runs $5{,}000$ steps of burn-in and up to $10^5$ steps.
The standard error of each estimate is computed by resampling the walkers, each with its whole trajectory, which accounts for the correlation along a stationary chain.

Table~\ref{tab:overlap_x2_classes} classes each pair by whether its two long chains agree and are long enough: \emph{converged} when they agree within three combined standard errors, every integrated autocorrelation time $\tau_{\mathrm{int}}$ is resolved, and each chain is at least $50\,\tau_{\mathrm{int}}$ long; \emph{short} when only the last condition fails; \emph{differ} when the chains differ by three combined standard errors or more; and \emph{unresolved} when a $\tau_{\mathrm{int}}$ window reaches the end of its chain.
A chain shorter than $50\,\tau_{\mathrm{int}}$ was extended once, except on F.

\begin{table}[ht]
\centering
\small
\caption{Pairs of networks of each atom in Figure~\ref{fig:overlap_x2} by the convergence of their two long chains, and the largest $|\tilde{\Sm}_{k\ell}|$ over all pairs of the atom.}
\label{tab:overlap_x2_classes}
\begin{tabular}{lrrrrrc}
\toprule
Atom & Pairs & Converged & Short & Differ & Unresolved & Largest $|\tilde{\Sm}_{k\ell}|$ \\
\midrule
Li & $45$ & $29$ & $14$ & $0$ & $2$ & $0.037$ \\
Be & $21$ & $21$ & $0$ & $0$ & $0$ & $0.037$ \\
B & $24$ & $18$ & $3$ & $2$ & $1$ & $0.032$ \\
C & $18$ & $18$ & $0$ & $0$ & $0$ & $0.026$ \\
N & $29$ & $28$ & $0$ & $1$ & $0$ & $0.021$ \\
O & $18$ & $18$ & $0$ & $0$ & $0$ & $0.017$ \\
F & $21$ & $9$ & $12$ & $0$ & $0$ & $0.021$ \\
\bottomrule
\end{tabular}
\end{table}

\section{Implementation details}
\label{app:sec:implementation_details}

Every choice in this section keeps the eigenfunctions of $\Top=-\hmltn$ and changes how training reaches them.
Appendix~\ref{app:sec:mc_statistics} gives the Monte Carlo estimates of energies and overlaps.

\subsection{The per-network shifted operator}
\label{app:sec:shifted_operator}

\paragraph{The weight from the analytic solve.}
The overlap-term weights $\lambda_\ell$ of $g_k$ under the shifted operator $\Top_k = E_{\mathrm{ref},k}\idop-\hmltn$ follow from the analytic solve of Eq.~\eqref{eq:analytic_eigvals}, and one extra triangular solve gives them for all $K$ networks.
The shift adds $E_{\mathrm{ref},k}$ to each $\Rm^{\Top}_{\ell\ell}$ and leaves $\Rm^{\idop}$ unchanged, and the solve is linear in $\diag(\Rm^{\Top})$, so the new weights, the eigenvalue parameters of $\Top_k$, are
\begin{align}
\label{eq:shifted_weights}
\boldsymbol{\lambda}^{(k)} = \boldsymbol{\lambda}^\star + E_{\mathrm{ref},k}\,\mathbf{u},
\qquad
\mathbf{u} \defeq \bigl(\Imat + \matmask^{\seq}\odot(\Rm^{\idop}\odot(\Rm^{\idop})^\intercal)\bigr)^{-1}\mathbf{1},
\end{align}
with $\mathbf{1}$ the all-ones vector.
Network $k$ uses $\lambda^{(k)}_\ell$ in place of $\lambda_\ell$ in $g_k(\xv)$.
When the overlaps between networks vanish, $\mathbf{u} = \mathbf{1}$.
With the prefactor $\lambda_k$ dropped, the shift adds the constant $-E_{\mathrm{ref},k}$ to the energy term of $g_k(\xv)$, which cancels in the centered gradient, so only the weights of the overlap term change.

\paragraph{Controlled experiment on carbon.}
The shift lowers the curvature of a carbon $^1\!D$ network toward the lower levels from about $724$ times the smaller curvature outside the level to at most $17$ times it, which stabilizes the energy convergence (Section~\ref{sec:exp_shifted_operator}).
For a network of the $^1\!D$ level other than the first in order, the smaller curvature outside the level is the $^1\!D$--$^1\!S$ gap of $52.2$\,mHa~\citep{Sansonetti--Martin2005}.
The first has no lower network in its spin sector, so the shift does not change its curvatures.
With $\Top=-\hmltn$, that is $E_{\mathrm{ref},k}=0$, the curvature toward the lower levels is $|E_k| \approx 37.8$\,Ha, so the ratio $\kappa$ of this curvature to the gap is about $724$; with $E_{\mathrm{ref},k} = -\lambda_k + c\,\sigma_k$ at $c=5$ it is $c\,\sigma_k \lesssim 0.87$\,Ha on the seed-$1$ carbon run of Table~\ref{tab:atoms_mae} after step $15{,}000$, above the gap, so $\kappa \lesssim 17$.

\paragraph{Weight floor.}
At an exact eigenstate, $-\lambda_k=E_k$ and $\sigma_k=0$, so $E_{\mathrm{ref},k}=E_k$ and the lower-state curvature vanishes.
A trained network is an approximate eigenstate, so in practice $\sigma_k$ stays positive.
In theory, if $E_\ell=E_k$ for some $\ell<k$, the weight $\lambda^{(k)}_\ell=E_{\mathrm{ref},k}-E_\ell$ is $c\,\sigma_k=0$, and a rotation of $\psi_k$ toward $\psi_\ell$ within their level keeps $\sigma_k=0$, so the objective leaves their overlap free.
This case arises only at an exact eigenstate, outside the practical regime, and as a guard the implementation replaces every weight $\lambda^{(k)}_\ell$ of Eq.~\eqref{eq:shifted_weights} by $\max(\lambda^{(k)}_\ell,1\,\mathrm{mHa})$.

\subsection{State ordering}
\label{app:sec:ewm_ordering}

Within each spin sector we permute the order indices of the networks at every step, ranking the networks by the EMA (Appendix~\ref{app:sec:mc_statistics}) of each network's batch mean of $\Ec_k + \hat S^2_{\mathrm{loc}}$, taken before clipping (Appendix~\ref{app:sec:spin_penalty_impl}).
The EMAs of $\Rm^{\Top}$, $\Rm^{\idop}$ and the norm ratios are kept per network, and the eigenvalue solve permutes $\Rm^{\Top}$ and $\Rm^{\idop}$ into the step's order.
At the fixed point of Corollary~\ref{cor:nested_fixed_point} the local energies are constant and the networks are already in energy order, so the assignment leaves the fixed point unchanged; within a degenerate level the order is free, as the corollary allows.
After step $10{,}000$, in our runs of Li to F, the assignment changes between networks that end in different levels in three runs: on one Li seed the networks of the $3p$ $^2\!P^\circ$ and $3d$ $^2\!D$ levels swap near step $17{,}000$, and on two B seeds two networks swap within the $3p$ $^2\!P^\circ$ level until one of them descends to the $^2\!D$ level.

\subsection{Spin penalty}
\label{app:sec:spin_penalty_impl}

The spin penalty replaces the operator $-\hmltn$ of the Hilbert--Schmidt objective by $-(\hmltn + \hat S^2)$, which keeps the eigenstates and raises each level of total spin $S$ by $S(S{+}1)$.
The molecular Hamiltonian commutes with $\hat S^2$ for spin-independent (Coulomb) interactions, so $\hmltn$ and $\hmltn + \hat S^2$ share an eigenbasis.

In \algo, $g_k$ in Eq.~\eqref{eq:def_g} replaces $\Top\psi_k/\psi_k$ with the augmented local quantity:
\begin{align*}
\frac{(\hmltn + \hat S^2)\psi_k(\xv)}{\psi_k(\xv)} = \Ec_k(\xv) + \hat S^2_{\mathrm{loc}}(\xv).
\end{align*}

The local spin $\hat S^2_{\mathrm{loc}}$ is evaluated by the L\"owdin swap-ratio formula on the same MCMC sample $\xv$ (Eq.~44 of \citet{Pfau--Axelrod--Sutterud--von-Glehn--Spencer2024}).
For $n_\uparrow \ge n_\downarrow$ up- and down-spin electrons, the L\"owdin formula is:
\begin{align*}
\hat S^2_{\mathrm{loc}}(\xv) = m_S(m_S{+}1) + n_\downarrow - \sum_{i\in\uparrow,\,j\in\downarrow}\frac{\psi_k(\tau_{ij}\xv)}{\psi_k(\xv)},
\end{align*}
where $\tau_{ij}$ exchanges the positions of up-spin electron $i$ and down-spin electron $j$.

Because both local quantities are evaluated on identical walkers, the per-walker sum $\Ec_k + \hat S^2_{\mathrm{loc}}$ is the local energy of $\hmltn + \hat S^2$, and we clip this sum as one quantity with the window of Appendix~\ref{app:sec:energy_statistics} (as the NES-VMC implementation does~\citep{ferminet_github}).
Clipping $\Ec_k$ alone and subtracting a separate batch-mean $\langle\hat S^2\rangle$ would leave the heavy tails of $\hat S^2_{\mathrm{loc}}$ unclipped.

We add $\hat S^2$ to the operator in the run of every atom, in one $m_S$ sector per spin multiplicity (Appendix~\ref{app:sec:atom_setup}), so the penalty enters only the $K$ local quantities $g_k$, one for each network.
\citet{Szabo--Schatzle--Entwistle--Noe2024} instead add the penalty to the loss as a separate term (Appendix~\ref{app:sec:penalty_explicit}), and NES-VMC~\citep{Pfau--Axelrod--Sutterud--von-Glehn--Spencer2024} adds a multiple of $\hat S^2$ to the Hamiltonian for the singlet states of cyclopentadienone and tetrazine, where the penalty enters all $K^2$ entries of its local-energy matrix.

\section{Experimental setup details}
\label{app:sec:exp_details}

This section gives the settings of the runs reported in Section~\ref{sec:exp_setup}.
Each network $\psi_k$ is a Psiformer (Appendix~\ref{app:sec:architectures}), and we discuss its computational stability in float32 (Appendix~\ref{app:sec:comp_stability}).
Each network samples its own walkers by Metropolis steps (Appendix~\ref{app:sec:samplers}) and is optimized by natural gradient descent with KFAC (Appendix~\ref{app:sec:ngd_kfac}).
The per-atom settings give the spin sectors, the terms each sector holds and the CASSCF active space (Appendix~\ref{app:sec:atom_setup}), and pretraining fits each network to one root of the CASSCF (Appendix~\ref{app:sec:pretrain_details}).
With the envelope shift, the local energies, $\log\psi_k$ and updates of the reported runs stayed finite, and non-finite values that could later occur are handled by dropping the affected walkers and undoing the affected steps (Appendix~\ref{app:sec:nan_handling}).

\subsection{Architectures}
\label{app:sec:architectures}

All experiments use Psiformer~\citep{glehn2023a} in the DeepQMC framework~\citep{deepqmc}; each per-state ansatz $\psi_k$ in our \algo method is an independent Psiformer instance.
As in FermiNet~\citep{Pfau--Spencer--Matthews--Foulkes2020}, each network has two stages: a permutation-equivariant backbone computes one feature vector per electron, and orbitals obtained from these features by a linear map enter determinants, which make $\psi_k$ antisymmetric.

\paragraph{Permutation-equivariant backbones.}
Each layer of the Psiformer or FermiNet backbone updates the feature of an electron from its own feature and from the unordered set of the features of all electrons, so permuting the electrons permutes the features in the same way.
In the Psiformer each layer is a self-attention over all electrons followed by an MLP~\citep{glehn2023a}, and in FermiNet each layer appends to the feature of an electron the means of the one- and two-electron features over the electrons of each spin~\citep{Pfau--Spencer--Matthews--Foulkes2020}.

Our Psiformer instance has four self-attention layers, four heads and MLP hidden dimension $256$ (Table~\ref{tab:psiformer_settings}).
Relative to the reference implementation~\citep{ferminet_github}, the DeepQMC Psiformer omits the layer normalization after each residual sum.

\begin{figure}[t]
  \centering
  \includegraphics[width=\linewidth]{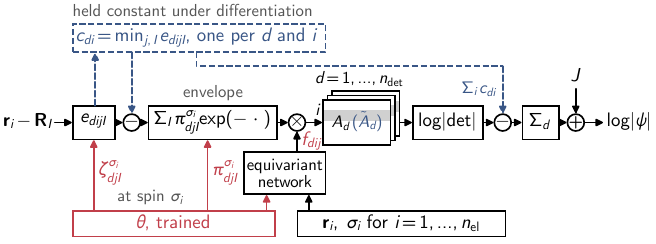}
  \caption{Computation of $\log|\psi|$ with the envelope shift, from the electron positions and spins and the trained parameters $\theta$. The constant $c_{di}$, held fixed under differentiation, is subtracted from the exponents in row $i$ of $A_d$ and its sum over $i$ from $\log|\det\tilde A_d|$.}
  \label{fig:envelope_flow}
\end{figure}

\paragraph{Orbitals and determinants.}
The second stage completes the antisymmetry: orbitals computed by a linear map from the feature of electron $i$ fill row $i$ of each determinant $\det A_d$, so the exchange of two electrons of the same spin swaps two rows of every determinant and changes the sign of $\psi_k$, and every network therefore lies in $\Hc$.
Figure~\ref{fig:envelope_flow} shows the detailed structure.

Each network is $\psi=e^{J}\sum_{d=1}^{n_{\det}}\det A_d$ with a Jastrow factor $e^{J}$ symmetric in the electrons and Slater matrices
\begin{equation*}
A_d=\begin{pmatrix}
\phi_{d1}(\rv_1)&\cdots&\phi_{d\numel}(\rv_1)\\
\vdots&&\vdots\\
\phi_{d1}(\rv_{\numel})&\cdots&\phi_{d\numel}(\rv_{\numel})
\end{pmatrix},\qquad
\phi_{dj}(\rv_i)=f_{dij}\sum_{I=1}^{\numnu}\pi^{\sigma_i}_{djI}\exp(-e_{dijI}),
\end{equation*}
with exponents $e_{dijI}=|\zeta^{\sigma_i}_{djI}|\,\|\rv_i-\Rv_I\|$.
Row $i\in[\numel]$ holds the orbitals $j\in[\numel]$ at electron $i$.
The index $d\in[n_{\det}]$ runs over the determinants and $I\in[\numnu]$ over the nuclei.
Each orbital is the network output $f_{dij}$ times the envelope $\sum_I\pi^{\sigma_i}_{djI}\exp(-e_{dijI})$.
The output $f_{dij}$ depends on all electrons.
The envelope parameters $\pi^{\sigma}_{djI}$ and $\zeta^{\sigma}_{djI}$ are trained for each spin $\sigma$ and taken at the spin $\sigma_i$ of electron $i$.

\begin{table}[t]
  \centering
  \small
  \caption{Settings of each network $\psi_k$.}
  \label{tab:psiformer_settings}
  \begin{tabular}{@{}lp{0.7\linewidth}@{}}
    \toprule
    Input of electron $i$ & $\rv_i-\Rv_I$ and $\|\rv_i-\Rv_I\|$ for every nucleus $I$, each scaled by $\log(1+\|\rv_i-\Rv_I\|)/\|\rv_i-\Rv_I\|$, and the spin $\sigma_i=\pm1$ \\
    Embedding & linear map without bias to dimension $256$ \\
    Layers & $4$, each a self-attention over all electrons with $4$ heads followed by a two-layer $\tanh$ MLP of width $256$, both with residual sums \\
    \midrule
    Orbitals & linear map without bias from the embedding of electron $i$ to $f_{dij}$ \\
    Determinants & $n_{\det}=16$, each over all $\numel$ electrons, summed with unit weights \\
    Envelopes & $\pi^{\sigma}_{djI}$ and $\zeta^{\sigma}_{djI}$ initialized to $1$ \\
    Jastrow factor & $J=-\sum_{i<j}c_{ij}\,\alpha^2/(\alpha+\|\rv_i-\rv_j\|)$ with $c_{ij}=1/4$ for equal and $1/2$ for opposite spins, and one trained $\alpha$ for each case, initialized to $1$ \\
    \bottomrule
  \end{tabular}
\end{table}

\subsubsection{Computational stability}
\label{app:sec:comp_stability}

Below we discuss the issues that can arise in float32 from the envelopes and from the derivative of the electron--nucleus distance, together with their solutions.

\paragraph{Envelope underflow.}
We keep $\log|\psi|$ and its derivatives finite in float32 by shifting the exponents in each row of the Slater matrix $A_d$ above, a shift that leaves them exact.
If a row of the shifted matrix $\tilde A_d$ still has largest absolute entry below the smallest normal float32 value $1.2\times10^{-38}$, its precision is lost and we set $\det A_d$ and its derivative to zero.
Each entry of $A_d$ is the network output $f_{dij}$ times an envelope that decays exponentially in the distance of electron $i$ from the nuclei, so without the shift the entries of an electron far from every nucleus are stored as zero and the derivative of $\log|\psi|$ becomes non-finite.

Consider an electron $i$ far from every nucleus, so that every exponent $e_{dijI}$ is large.
Each term $\exp(-e_{dijI})$ falls below $1.2\times10^{-38}$ once $e_{dijI}\gtrsim 87$ and is stored as zero once $e_{dijI}\gtrsim 103$, so row $i$ is zero or subnormal in some or all $A_d$.
The derivative of $\log|\det A_d|$ is the transposed inverse $(A_d^{-1})^{\top}$, and the inverse of a matrix with a zero or subnormal row is non-finite.
A determinant that contributes nothing to $\psi$ still passes this value on, because its zero weight multiplies a non-finite derivative.

We subtract the smallest exponent in row $i$ of $A_d$, $c_{di}=\min_{j,I}e_{dijI}$, from every exponent $e_{dijI}$ of that row (Figure~\ref{fig:envelope_flow}).
This multiplies row $i$ by $e^{c_{di}}$ and gives the shifted matrix $\tilde A_d$, whose largest exponential in row $i$ is $\exp(0)=1$.
The shift is the subtract-max trick of the softmax applied to the logits $-e_{dijI}$, whose largest value is $-c_{di}$, and it leaves $\log|\psi|$ and its derivatives exact.
In the softmax the trick prevents overflow of large logits; here every logit $-e_{dijI}$ is at most zero, and for a distant electron the largest one, $-c_{di}$, is large and negative, so the subtraction raises every entry of row $i$ and prevents underflow.

A determinant is linear in each row, so $\log|\det A_d|=\log|\det\tilde A_d|-\sum_i c_{di}$ for every value of $c_{di}$.
We compute $\log|\det\tilde A_d|$ from rows stored at full precision and subtract $\sum_i c_{di}$ before the determinants are summed.
Since the identity holds for every $c_{di}$, we hold $c_{di}$ constant under differentiation.
A second subtract-max, over the determinants instead of the entries of a row, sums the determinants:
\begin{equation*}
x_d=\log|\det\tilde A_d|-\sum_{i=1}^{\numel}c_{di},\quad
M=\max_d x_d,\quad
\log|\psi|=J+M+\log\Bigl|\sum_{d=1}^{n_{\det}}\mathrm{sgn}(\det\tilde A_d)\,e^{x_d-M}\Bigr|.
\end{equation*}
The first shift keeps the entries of $\tilde A_d$ from underflowing, and the second keeps the terms $e^{x_d}$ of the sum, lowered by $\sum_i c_{di}$, from underflowing.

\paragraph{Safe norm of the electron--nucleus distance.}
The reported runs were trained with the safe norm $\rho=\sqrt{r^2+a^2}$ of the DeepQMC code~\citep{deepqmc}, where $a^2$ is the float32 machine epsilon ($a=3.45\times10^{-4}$\,bohr), in place of each electron--nucleus distance $r=\|\rv_i-\Rv_I\|$ in the exponents of the envelopes and in the input features.
This helps with the undefined derivative $(\rv_i-\Rv_I)/r$ of $r$ where $r$ vanishes, since that of $\rho$ is finite everywhere.

While this choice may be numerically stable, it also induces rare jumps of the local energy, and the reported energies evaluate the network and the local energy with exact interparticle distances (Appendix~\ref{app:sec:per_state_mae}), which removes these jumps.
The potential $-Z/r$ takes the exact distance $r$, so within about $a$ of a nucleus the safe norm breaks the cancellation between the kinetic energy and the nuclear Coulomb term.
For an envelope $e^{-\zeta\rho}$, the term $-\partial_r\psi/(r\psi)$ of the kinetic energy equals $\zeta/\rho$ and is at most $\zeta/a$, whereas for $e^{-\zeta r}$ it equals $\zeta/r$ and cancels $-Z/r$ at the nuclear cusp when $\zeta=Z$.
A walker with an electron within about $a$ of a nucleus thus has a local energy far below its value with exact distances, which makes the unclipped batch mean of the local energy jump at the step that samples it.

We conjecture that training with the exact distance $r$ removes these jumps and gives more accurate energies, since the envelopes then keep this cancellation at every distance from a nucleus.
We also suspect that the safe norm is a broader precaution than the computation needs, because the FermiNet code~\citep{ferminet_github} keeps the exact electron--nucleus distance and masks only the difference of each electron with itself, and the same mask appears applicable here.

\subsection{Samplers}
\label{app:sec:samplers}

The training chains follow the Metropolis sampler of DeepQMC~v1.2.0~\citep{deepqmc,Schatzle--Szabo--Mezera--Hermann--Noe2023}: each network $\psi_k$ has its own chain of $4{,}096$ walkers, which samples $p_k$ (Section~\ref{sec:exp_setup}).
A proposal moves all electrons of a walker by a Gaussian step of width $\tau$ and is accepted with probability $\min\{1,|\psi_k(\xv')|^2/|\psi_k(\xv)|^2\}$.
Each training step takes $30$ Metropolis steps, and after each Metropolis step $\tau$ is multiplied by $\max(a_{\mathrm{acc}},0.05)/0.57$, where $a_{\mathrm{acc}}$ is the acceptance rate over the walkers of the chain; $\tau$ starts at $1$ bohr.
The chains start from the walkers at the end of pretraining and are equilibrated under the pretrained networks before training.

\subsection{Natural-gradient descent with KFAC}
\label{app:sec:ngd_kfac}

Optimization uses per-state natural gradient descent: each per-state gradient is left-multiplied by the inverse Fisher information $F_k^{-1}$, with
\begin{align*}
F_k \defeq \E_{p_k}[\nabla_\theta \log p_k(\xv)\,\nabla_\theta \log p_k(\xv)^\intercal].
\end{align*}
The parameter update direction is
\begin{align}
\label{eq:ngd_update}
\Delta\theta = 2\sum_{k=1}^K F_k^{-1}\,\E_{p_k(\xv)}\!\bigl[(g_{k}(\xv) - \E_{p_k}[g_{k}])\,\nabla_\theta \log |\psi_k|\bigr].
\end{align}
In practice each $F_k$ is approximated by KFAC~\citep{pmlr-v37-martens15}, applied per state.
The factorization by state holds because each network samples its own Born density (Section~\ref{sec:scaling}): the walkers $\xv \sim p_k$ enter only $F_k$, so the KFAC blocks decouple across $k$.

\paragraph{Hyperparameters.}
The optimizer uses an inverse-time learning-rate schedule $\eta_n = 0.05 / (1 + n / 5\times 10^4)$.
KFAC-specific settings: norm constraint $10^{-3}$, constant damping $10^{-3}$, Fisher factors computed exactly from the registered output $\log|\psi_k|$ rather than from sampled targets, inverse update period $1$, no burn-in.
The norm constraint is applied to each spin sector separately: the update of the networks of a sector is multiplied by $\min\{1, (10^{-3}/(\eta_n^2 G))^{1/2}\}$, where $G$ is the sum over the networks of that sector of the inner product of each network's preconditioned gradient with its gradient.

\subsection{Per-atom setup}
\label{app:sec:atom_setup}

\begin{table}[h]
\centering
\footnotesize
\caption{Per-atom physical configuration. Each row lists the terms one spin sector $m_S=S$ (Section~\ref{sec:exp_setup}) holds, in NIST term-centroid order, with the number of networks assigned to each term in parentheses. A fully covered term gets its $2L{+}1$ orbital sub-states; a term the $K{=}10$ window cuts keeps fewer (e.g.~Li's ${}^2\!D(2)$ keeps $2$ of its $5$).
The CAS $(N_e, N_o)$ is chosen per atom by counting the orbitals of the $K{=}10$ configurations (Li, C, Ne) or by enlarging the valence shell by $\{3s,3p\}$ (Be, B, N, O, F).}
\label{tab:atom_setup}
\begin{tabular}{lclc}
\toprule
atom & $m_S$ & $K{=}10$ block & CAS $(N_e, N_o)$ \\
\midrule
Li & $1/2$ & ${}^2\!S(1),\,{}^2\!P^\circ(3),\,{}^2\!S(1),\,{}^2\!P^\circ(3),\,{}^2\!D(2)$ & $(1, 13)$ \\
\midrule
Be & $0$ & ${}^1\!S(1),\,{}^1\!P^\circ(3),\,{}^1\!S(1),\,{}^1\!D(1)$ & $(2, 8)$ \\
   & $1$ & ${}^3\!P^\circ(3),\,{}^3\!S(1)$ & \\
\midrule
B  & $1/2$ & ${}^2\!P^\circ(3),\,{}^2\!S(1),\,{}^2\!D(3)$ & $(3, 8)$ \\
   & $3/2$ & ${}^4\!P(3)$ & \\
\midrule
C  & $0$ & ${}^1\!D(5),\,{}^1\!S(1)$ & $(4, 4)$ \\
   & $1$ & ${}^3\!P(3)$ & \\
   & $2$ & ${}^5\!S^\circ(1)$ & \\
\midrule
N  & $1/2$ & ${}^2\!D^\circ(5),\,{}^2\!P^\circ(3)$ & $(5, 8)$ \\
   & $3/2$ & ${}^4\!S^\circ(1),\,{}^4\!P(1)$ & \\
\midrule
O  & $0$ & ${}^1\!D(5),\,{}^1\!S(1)$ & $(6, 8)$ \\
   & $1$ & ${}^3\!P(3)$ & \\
   & $2$ & ${}^5\!S^\circ(1)$ & \\
\midrule
F  & $1/2$ & ${}^2\!P^\circ(3),\,{}^2\!P(3)$ & $(7, 8)$ \\
   & $3/2$ & ${}^4\!P(3),\,{}^4\!P^\circ(1)$ & \\
\midrule
Ne\textsuperscript{$\dagger$} & $0$ & ${}^1\!S(1),\,{}^2[1/2]^{\!\circ}(3)$ & $(8, 8)$ \\
   & $1$ & ${}^2[3/2]^{\!\circ}(3),\,{}^2[1/2](1),\,{}^2[5/2](1),\,{}^2[3/2](1)$ & \\
\bottomrule
\end{tabular}
\par\smallskip
\parbox{\linewidth}{\textsuperscript{$\dagger$}NIST names each level by its total angular momentum $J$, which spin--orbit coupling conserves, and by the basis state with the largest weight in that level~\citep{Sansonetti--Martin2005}.
For Li--F this basis state is an LS term, and for the excited levels of Ne it is a $J_1l$-coupled state ${}^{2S_2+1}[K]^\pi_J$: the core's angular momentum $J_1$ and the outer electron's orbital angular momentum $l$ give $K$, and the leading $2$ is the multiplicity of the outer electron's spin.}
\end{table}

\paragraph{Sector decision.} Every atom runs \algo on $\Top=-(\hmltn+\hat S^2)$ in one $m_S$ sector per spin multiplicity, and the states of all sectors are merged by energy (Section~\ref{sec:exp_setup}).
The augmented Hamiltonian shifts each state by $S(S{+}1)$\,Ha.
This shift differs by at least $2$\,Ha between two values of $S$ in one sector, far above the excitation energies in the $K{=}10$ window (Table~\ref{tab:per_state_master}), so a single-sector run with the spin penalty would lift every state with $S>|m_S|$ out of the $K{=}10$ window.
One $m_S$ sector per spin multiplicity is also the recipe of \citet{Szabo--Schatzle--Entwistle--Noe2024} (their Section~2.1.3), who fix $m_S$ to reach each higher spin multiplicity and keep the spin penalty on.

\subsection{Pretraining}
\label{app:sec:pretrain_details}

\paragraph{Per-state CASSCF targets.}
Network $k$ of a sector is pretrained to the leading determinants of the $k$-th root of the sector's state-averaged CASSCF.

\paragraph{Pretraining sampler.} As in FermiNet's excited-state pretraining~\citep{ferminet_github}, each state's pretraining walkers come from a chain on the density of its CASSCF reference that continues from one iteration to the next, whereas the penalty baseline~\citep{Szabo--Schatzle--Entwistle--Noe2024} restarts its chains at every iteration.

\paragraph{Optimization.}
The pretrain loss is the mean-squared error between the network orbital matrices and the per-state CASSCF target.
We optimize with Lamb at learning rate $3\times 10^{-4}$ ($\beta_1{=}0.9$, $\beta_2{=}0.999$) for $10{,}000$ steps in all reported runs.

\subsection{NaN handling in training}
\label{app:sec:nan_handling}

A non-finite $\log\psi_k$, local energy or cross-state ratio at a single walker makes every batch mean it enters non-finite.
A non-finite update then leaves the parameters non-finite at every later step.
Although the reported runs kept $\log\psi_k$, the local energy and the parameters finite with the envelope shift of Appendix~\ref{app:sec:comp_stability}, we adopt two safeguards from the NES-VMC stability handler of \citet[Section~S8]{Pfau--Axelrod--Sutterud--von-Glehn--Spencer2024}.
(i)~Per-walker exclusion: any walker with non-finite $\log\psi_k$, local energy or cross-state ratio is dropped from every batch reduction in $g_k$, including the analytic-eigenvalue solver inputs and the EMAs.
(ii)~Step undo with a failure counter: a step that leaves the parameters, the optimizer state or the statistics fed back to the next step non-finite is undone, as FermiNet does~\citep{ferminet_github}.
A step with a non-finite walker or an undone update increments a counter of consecutive failures, and at $100$ the run restarts from its last checkpoint.
The other two safeguards of that handler address the joint matrix of NES-VMC and larger systems.
Extra MCMC on zero entries $\psi_i(\xv^j)$ treats a sparse joint matrix with nonzero $\det\Psi$.
In \algo each network $k$ samples its walkers from $|\psi_k|^2$, so $\psi_k$ is nonzero at every one of them.
The median-of-minibatches energy estimator was introduced for systems with $>25$ electrons, beyond the first-row atoms studied here.

\section{Per-step cost analysis}
\label{app:sec:cost_analysis}

\algo evaluates $K$ Laplacians per training step and NES-VMC evaluates $K^2$.
This section counts the cost of one step for both methods and measures the wall time of \algo on Ne as $K$ grows.

One forward evaluation of a network costs $\Cc_{\mathrm{eval}}=\Cc_{\mathrm{bb}}+\Cc_{\mathrm{post}}$, the sum over the two stages of Appendix~\ref{app:sec:architectures}, with $\Cc_{\mathrm{bb}}\sim\numel^2$ and $\Cc_{\mathrm{post}}\sim\numel^3$ at fixed depth and width.
The backbone of a network is its permutation-equivariant layers, of order $\numel^2$ from the self-attention of Psiformer or the symmetric features of FermiNet.
The orbitals, envelopes and determinants of one state have the order $\numel^3$ from the determinants.
A Laplacian costs $\Cc_{\mathrm{lap}}\sim3\numel\,\Cc_{\mathrm{eval}}\sim\numel^4$ (Appendix~\ref{app:sec:folx}).
A training step takes $n_{\mathrm{MH}}=30$ Metropolis steps (Appendix~\ref{app:sec:samplers}).

\subsection{Forward-Laplacian primitive (folx)}
\label{app:sec:folx}

We compute every Laplacian $\Delta f=\sum_{k=1}^{3\numel}\partial^2 f/\partial x_k^2$ with the forward Laplacian of \texttt{folx}~\citep{gao2023folx}, at about half the cost of the standard Hessian-diagonal loop.
DeepQMC~\citep{deepqmc} and FermiNet~\citep{ferminet_github} by default run $3\numel$ forward-mode Jacobian--vector products through the gradient $\nabla_\xv f$, each about one forward and one backward pass, $\sim2\cdot3\numel\,\Cc_{\mathrm{eval}}$ in total.
The forward Laplacian instead propagates $(h,\nabla h,\Delta h)$ with $\nabla h\in\Real^{3\numel\times d_l}$ through the network in one pass, and the Jacobian products $(3\numel\times d_l)\cdot(d_l\times d_{l+1})$ of its layers sum to $\Cc_{\mathrm{lap}}\sim3\numel\,\Cc_{\mathrm{eval}}$.

The sparse mode of \texttt{folx} shrinks the Jacobian to $(s\times d_l)$ in layers whose outputs depend on at most $s$ input coordinates, which holds with $s=3$ only in the leading per-electron embedding and MLP.
From the first self-attention of Psiformer~\citep{glehn2023a} or the first mean over electrons of FermiNet~\citep{Pfau--Spencer--Matthews--Foulkes2020} on, the features of an electron depend on all $3\numel$ coordinates, so the Laplacian cost stays at $\sim3\numel\,\Cc_{\mathrm{eval}}$.

\subsection{Cost decomposition: Hamiltonian coupling and ansatz parameterization}
\label{app:sec:nes_costs}

Table~\ref{tab:cost} counts every component of a step under the two ansatz parameterizations, a joint backbone with $K$ heads and $K$ disjoint networks.

\begin{table}[t]
\centering
\small
\setlength{\tabcolsep}{4pt}
\caption{Cost of one training step by component. $\Cc_{\mathrm{bb}}$ is one forward evaluation of the backbone, $\Cc_{\mathrm{post}}$ that of one state's orbitals, envelopes and determinants, and $\Cc_{\mathrm{eval}}=\Cc_{\mathrm{bb}}+\Cc_{\mathrm{post}}$; a training step takes $n_{\mathrm{MH}}$ Metropolis steps. The last column gives the leading order in $K$ and $\numel$. Bold: the Laplacian, the largest single evaluation.}
\label{tab:cost}
\begin{tabular}{@{}llccc@{}}
\toprule
 & Component & Joint backbone, $K$ heads & $K$ disjoint networks & Order \\
\midrule
NES-VMC & Laplacian & $\boldsymbol{3\numel(K\Cc_{\mathrm{bb}}+K^2\Cc_{\mathrm{post}})}$ & $\boldsymbol{3\numel K^2\Cc_{\mathrm{eval}}}$ & $K^2\numel^4$ \\
 & Sampling & $n_{\mathrm{MH}}(K\Cc_{\mathrm{bb}}+K^2\Cc_{\mathrm{post}})$ & $n_{\mathrm{MH}}K^2\Cc_{\mathrm{eval}}$ & $n_{\mathrm{MH}}K^2\numel^3$ \\
 & Inverse of $\Psi$ & $K^3$ & $K^3$ & $K^3$ \\
\midrule
\algo, penalty & Laplacian & $\boldsymbol{3\numel K\Cc_{\mathrm{eval}}}$ & $\boldsymbol{3\numel K\Cc_{\mathrm{eval}}}$ & $K\numel^4$ \\
 & Sampling & $n_{\mathrm{MH}}K\Cc_{\mathrm{eval}}$ & $n_{\mathrm{MH}}K\Cc_{\mathrm{eval}}$ & $n_{\mathrm{MH}}K\numel^3$ \\
 & Overlaps & $K^2\Cc_{\mathrm{post}}$ & $K^2\Cc_{\mathrm{eval}}$ & $K^2\numel^3$ \\
\bottomrule
\end{tabular}
\end{table}

\paragraph{Hamiltonian coupling: off-diagonal Laplacian count.}
The $K^2$ Laplacians of NES-VMC come from its local-energy matrix $\locmat(\Xv) = \Psi(\Xv)^{-1}(\hmltn\Psi)(\Xv)$, which contains $(\hmltn\psi_k)(\xv^i)$ at every state-walker pair $(i,k)$.
In \algo and the penalty methods the cross term is the overlap $\ip{\psi_k}{\psi_\ell}$, a forward-only evaluation, so only the $K$ diagonal local energies $\hmltn\psi_k/\psi_k$ need a Laplacian, a factor $K$ fewer than NES-VMC.
The inverse $\Psi^{-1}$ and the product $\Psi^{-1}(\hmltn\Psi)$ add $O(K^3)$ per step to NES-VMC, below its Laplacian cost $K^2\numel^4$ for $\numel\gtrsim K$.

\paragraph{Ansatz parameterization and walker-target factorization.}
The parameterization changes the overlap term of \algo, and the Laplacian and sampling terms of NES-VMC.
The walker target $|\psi_1|^2 \cdots |\psi_K|^2$ of \algo and the penalty methods factorizes into $K$ Markov chains, so sampling and Laplacians evaluate each network at its own walkers under either parameterization.
The overlaps evaluate every network at every walker set, at $K^2\,\Cc_{\mathrm{post}}$ with a joint backbone, whose $K$ heads share the backbone features of a walker set, and at $K^2\,\Cc_{\mathrm{eval}}$ with the $K$ disjoint networks that \algo uses, following \citet{Szabo--Schatzle--Entwistle--Noe2024}.
The target $|\det\Psi|^2$ of NES-VMC couples the $K$ walker sets, so its sampling and Laplacians evaluate the full matrix $\Psi(\Xv)$, at $K\,\Cc_{\mathrm{bb}} + K^2\,\Cc_{\mathrm{post}}$ with a joint backbone, as in the NES-VMC reference implementation~\citep{ferminet_github}, and at $K^2\,\Cc_{\mathrm{eval}}$ with disjoint networks.

\paragraph{Cost of the spin penalty.}
For the first-row atoms, with $\numel\le10$, the L\"owdin sum of Appendix~\ref{app:sec:spin_penalty_impl} costs less than the $K$ Laplacians of a step and much less than the $K(K-1)$ Laplacians that \algo saves against NES-VMC.
Each of its $n_\uparrow n_\downarrow$ terms is one forward evaluation of $\psi_k$, because every orbital of the Psiformer depends on all electron positions.
Evaluated at every walker, it adds $K n_\uparrow n_\downarrow \le K\numel^2/4$ forward evaluations per walker and step to the forward counts of Table~\ref{tab:cost}, of order $K\numel^5$ against the order $K\numel^4$ of the Laplacians.

In larger systems the spin term can therefore cost more than the Laplacians, but a small random subsample of the walkers suffices to estimate it, because the eigenvalues $S(S+1)$ of neighboring spin sectors differ by at least $2$, far above the excitation energies.
Evaluating the spin term on a random eighth of the walkers keeps its cost below that of the Laplacians up to about a hundred electrons, $\numel<96$.
On Ne, training with the spin term on a random eighth of the walkers reached the same accuracy as training with it on every walker.

\subsection{Empirical wall-time scaling on Ne}
\label{app:sec:scaling}

We confirm the scaling in $K$ claimed in Section~\ref{sec:scaling} by sweeping $K$ on Ne.
Measurements run on a single host with $8{\times}$ RTX~5090 (FP32, JAX 0.9.2).
For each value of $K$ we run 100 main-loop warm-up steps followed by 100 measurement steps.
The sweep places all $K$ networks in one spin sector and runs the training step with the spin penalty off.
With two or more sectors, overlaps are evaluated only between networks of the same sector, and the spin penalty adds the cost given in Appendix~\ref{app:sec:nes_costs}.
Figure~\ref{fig:scaling} reports two batch sizes.
At the training batch of $4{,}096$ the walkers no longer fit in memory beyond $K{=}24$.
The exponent fitted at this batch in Figure~\ref{fig:scaling} has standard error $0.02$ ($95\%$ interval $\pm0.04$) and describes the measured range $K{=}6$ to $24$ on this host.

\begin{figure}[t]
  \centering
  \includegraphics[width=0.7\linewidth]{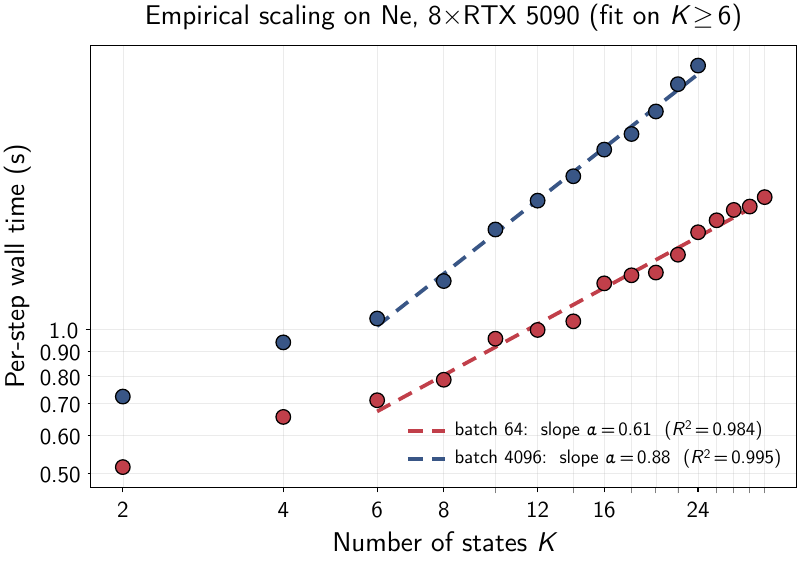}
  \caption{Empirical per-step wall time of \algo's main optimization loop on Ne, $8{\times}$RTX~5090, FP32, at batches of $4{,}096$ and $64$. Dashed lines are least-squares power-law fits. The spin penalty is off.}
  \label{fig:scaling}
\end{figure}

\section{Per-state errors across first-row atoms}
\label{app:sec:per_state_mae}

This section reports the per-state errors behind the per-atom mean absolute error (MAE, in mHa) of Table~\ref{tab:atoms_mae}, for \algo, NES-VMC, and \citet{Szabo--Schatzle--Entwistle--Noe2024} against $(2J{+}1)$-weighted NIST term centroids on each of the eight first-row atoms at $K{=}10$.
Table~\ref{tab:per_state_master} gives the per-state breakdown of one run per atom, whose MAE is in Table~\ref{tab:mae_intervals}.

\paragraph{Per-state stderr and MAE interval.}
Each per-state energy $E_k$ is the unweighted mean of the per-state batch mean of the local energy over training steps $97{,}000$ to $99{,}999$, and its stderr is the standard deviation of the means of $10$ contiguous blocks of $300$ steps, divided by $\sqrt{10}$.
For Ne seed~$1$, read at step $99{,}000$, the mean runs over steps $97{,}000$ to $98{,}999$ and the stderr uses its last $6$ blocks of $300$ steps.
The local energy is unclipped and is evaluated on each step's walkers with exact interparticle distances (Appendix~\ref{app:sec:architectures}).
Blocks of $100$ steps give the same stderr: over the states of the runs in Table~\ref{tab:atoms_mae}, the ratio of the $10$-block stderr to the $30$-block stderr has median $1.00$, within the precision of a standard deviation estimated from $10$ blocks.

The interval of the MAE comes from a block bootstrap, because the sampling error of the anchor $E_1$ enters every per-state error $\varepsilon_k = (E_k - E_1) - t_k$, with $t_k$ the target excitation energy.
Each of $4{,}000$ resamples draws the blocks with replacement, the same blocks for every state, and recomputes the MAE from the resampled per-state means: it sorts them, subtracts the lowest and scores the excitations against the targets $t_k$.
The $95\%$ interval runs from the $2.5$th to the $97.5$th percentile of the resampled MAE.
Table~\ref{tab:mae_intervals} gives the interval of every run in Table~\ref{tab:atoms_mae}; blocks of $100$ and of $300$ steps give nearly the same interval.
The interval measures the sampling noise of one run, and on B it is narrower than the spread across the three seeds.

\begin{table}[ht]
\centering
\small
\caption{MAE (mHa) of each run entering Table~\ref{tab:atoms_mae}, with its $95\%$ block-bootstrap interval for blocks of $100$ and of $300$ training steps over the last $3{,}000$ steps ($2{,}000$ for Ne seed~$1$). Steps is the number of training steps the run had completed when it was read.}
\label{tab:mae_intervals}
\begin{tabular}{lrrccc}
\toprule
Atom & Seed & Steps & MAE & Blocks of $100$ & Blocks of $300$ \\
\midrule
Li & 1 & 100{,}000 & 0.046 & [0.034, 0.073] & [0.036, 0.070] \\
Li & 2 & 100{,}000 & 0.066 & [0.049, 0.093] & [0.051, 0.086] \\
Li & 3 & 100{,}000 & 0.059 & [0.046, 0.083] & [0.046, 0.086] \\
Be & 1 & 100{,}000 & 0.019 & [0.015, 0.042] & [0.015, 0.042] \\
Be & 2 & 100{,}000 & 0.035 & [0.028, 0.054] & [0.025, 0.056] \\
Be & 3 & 100{,}000 & 0.036 & [0.022, 0.064] & [0.026, 0.059] \\
B & 1 & 100{,}000 & 0.498 & [0.489, 0.533] & [0.487, 0.531] \\
B & 2 & 100{,}000 & 0.521 & [0.507, 0.558] & [0.505, 0.561] \\
B & 3 & 100{,}000 & 0.915 & [0.901, 0.954] & [0.897, 0.962] \\
C & 1 & 100{,}000 & 0.177 & [0.147, 0.207] & [0.150, 0.201] \\
C & 2 & 100{,}000 & 0.193 & [0.148, 0.243] & [0.139, 0.260] \\
N & 1 & 100{,}000 & 0.276 & [0.213, 0.338] & [0.214, 0.337] \\
O & 1 & 100{,}000 & 0.265 & [0.234, 0.368] & [0.243, 0.353] \\
F & 1 & 100{,}000 & 0.647 & [0.555, 0.781] & [0.561, 0.780] \\
Ne & 1 & 99{,}000 & 1.944 & [1.746, 2.139] & [1.665, 2.118] \\
Ne & 2 & 100{,}000 & 1.153 & [1.051, 1.280] & [1.068, 1.253] \\
\bottomrule
\end{tabular}
\end{table}

\paragraph{B: tenth state.}
On B, the tenth states of \algo and of NES-VMC lie near the $2s^2 3p$ $^2\!P^\circ$ level ($\Delta E_{\text{NIST}} = 0.22145$\,Ha), not at a third $^2\!D$ substate.
The $k{=}10$ row of Table~\ref{tab:per_state_master} scores both against the $^2\!D$ centroid.

{\footnotesize
\makeatletter
\def\LT@makecaption#1#2#3{%
  \LT@mcol\LT@cols{@{}l}{\hbox to\z@{\parbox[t]\textwidth{%
    \normalsize\reset@font #1{#2: }#3\endgraf\vskip\baselineskip}\hss}}}
\makeatother
\setlength{\tabcolsep}{2.5pt}
\begin{longtable}{@{}llllrrrrrrr@{}}
\caption{Per-state $K{=}10$ results across first-row atoms (Li--Ne), one run per atom: seed~$1$ (Ne: seed~$2$), read at the step given in Table~\ref{tab:mae_intervals}. Within an atom, the rows list the target terms by ascending $\Delta E_{\text{NIST}}$, each repeated by its number of networks in Table~\ref{tab:atom_setup}, against the run's ten lowest states in ascending energy. $E$ is the per-state mean defined above, with its stderr in parentheses. Errors in mHa against the NIST term centroids of Section~\ref{sec:exp_setup}; baseline errors from~\citet[Table~S2]{Pfau--Axelrod--Sutterud--von-Glehn--Spencer2024} and~\citet[SI Table~S1]{Szabo--Schatzle--Entwistle--Noe2024}. Config.\ omits the closed 1s$^{2}$ shell (Ne: 1s$^{2}$2s$^{2}$); each column is rounded from unrounded values.}
\label{tab:per_state_master} \\
\toprule
& & & & \multicolumn{3}{c}{This work} & NIST & & \multicolumn{2}{c}{Baselines (mHa)} \\
\cmidrule(lr){5-7} \cmidrule(lr){10-11}
System & $k$ & Config. & Term & $E$ (Ha) & $\langle\hat S^2\rangle$ & $\Delta E$ (Ha) & $\Delta E_{\text{NIST}}$ (Ha) & Err (mHa) & \shortstack{NES-VMC\\\citep{Pfau--Axelrod--Sutterud--von-Glehn--Spencer2024}} & \shortstack{Szab\'o\\\citep{Szabo--Schatzle--Entwistle--Noe2024}} \\
\midrule
\endfirsthead
\multicolumn{11}{c}{\tablename\ \thetable{} -- continued from previous page} \\
\toprule
& & & & \multicolumn{3}{c}{This work} & NIST & & \multicolumn{2}{c}{Baselines (mHa)} \\
\cmidrule(lr){5-7} \cmidrule(lr){10-11}
System & $k$ & Config. & Term & $E$ (Ha) & $\langle\hat S^2\rangle$ & $\Delta E$ (Ha) & $\Delta E_{\text{NIST}}$ (Ha) & Err (mHa) & \shortstack{NES-VMC\\\citep{Pfau--Axelrod--Sutterud--von-Glehn--Spencer2024}} & \shortstack{Szab\'o\\\citep{Szabo--Schatzle--Entwistle--Noe2024}} \\
\midrule
\endhead
\midrule \multicolumn{11}{r}{\textit{continued on next page}} \\
\endfoot
\bottomrule
\endlastfoot

  Li & 1 & 2s & ${}^{2}S$ & -7.47805(1) & 0.750 & 0.00000 & 0.00000 & +0.00 & --- & --- \\
  Li & 2 & 2p & ${}^{2}P^{\circ}$ & -7.41015(2) & 0.750 & 0.06790(2) & 0.06791 & +0.00 & +0.00 & +0.05 \\
  Li & 3 & 2p & ${}^{2}P^{\circ}$ & -7.41014(2) & 0.750 & 0.06791(2) & 0.06791 & +0.00 & +0.01 & +0.07 \\
  Li & 4 & 2p & ${}^{2}P^{\circ}$ & -7.41010(1) & 0.750 & 0.06795(2) & 0.06791 & +0.04 & +0.01 & +0.14 \\
  Li & 5 & 3s & ${}^{2}S$ & -7.35408(3) & 0.750 & 0.12397(3) & 0.12396 & +0.01 & +0.01 & +0.22 \\
  Li & 6 & 3p & ${}^{2}P^{\circ}$ & -7.33709(1) & 0.750 & 0.14096(2) & 0.14091 & +0.05 & +0.01 & +0.57 \\
  Li & 7 & 3p & ${}^{2}P^{\circ}$ & -7.33708(1) & 0.750 & 0.14097(2) & 0.14091 & +0.06 & +0.02 & +2.12 \\
  Li & 8 & 3p & ${}^{2}P^{\circ}$ & -7.33704(2) & 0.750 & 0.14101(2) & 0.14091 & +0.10 & +0.03 & +2.25 \\
  Li & 9 & 3d & ${}^{2}D$ & -7.33552(2) & 0.750 & 0.14253(2) & 0.14254 & -0.01 & +0.02 & +1.45 \\
  Li & 10 & 3d & ${}^{2}D$ & -7.33538(2) & 0.750 & 0.14267(2) & 0.14254 & +0.14 & +0.03 & +2.61 \\
\cmidrule(lr){2-11}
  Be & 1 & 2s$^{2}$ & ${}^{1}S$ & -14.66737(2) & 0.000 & 0.00000 & 0.00000 & +0.00 & --- & --- \\
  Be & 2 & 2s2p & ${}^{3}P^{\circ}$ & -14.56723(2) & 2.000 & 0.10013(2) & 0.10015 & -0.01 & -0.05 & -0.04 \\
  Be & 3 & 2s2p & ${}^{3}P^{\circ}$ & -14.56722(1) & 2.000 & 0.10015(2) & 0.10015 & +0.00 & -0.04 & +0.01 \\
  Be & 4 & 2s2p & ${}^{3}P^{\circ}$ & -14.56721(1) & 2.000 & 0.10016(2) & 0.10015 & +0.01 & -0.04 & +0.02 \\
  Be & 5 & 2s2p & ${}^{1}P^{\circ}$ & -14.47344(1) & 0.000 & 0.19393(2) & 0.19394 & -0.01 & -0.05 & -0.09 \\
  Be & 6 & 2s2p & ${}^{1}P^{\circ}$ & -14.47343(2) & 0.000 & 0.19393(2) & 0.19394 & -0.01 & -0.03 & +0.07 \\
  Be & 7 & 2s2p & ${}^{1}P^{\circ}$ & -14.47342(2) & 0.000 & 0.19395(2) & 0.19394 & +0.01 & -0.03 & +0.07 \\
  Be & 8 & 2s3s & ${}^{3}S$ & -14.43002(2) & 2.000 & 0.23735(3) & 0.23730 & +0.05 & +0.03 & +0.07 \\
  Be & 9 & 2s3s & ${}^{1}S$ & -14.41819(3) & 0.000 & 0.24918(3) & 0.24913 & +0.05 & +0.10 & +1.27 \\
  Be & 10 & 2p$^{2}$ & ${}^{1}D$ & -14.40821(2) & 0.000 & 0.25916(3) & 0.25917 & -0.01 & -0.00 & +0.14 \\
\cmidrule(lr){2-11}
  B & 1 & 2s$^{2}$2p & ${}^{2}P^{\circ}$ & -24.65381(2) & 0.750 & 0.00000 & 0.00000 & +0.00 & --- & --- \\
  B & 2 & 2s$^{2}$2p & ${}^{2}P^{\circ}$ & -24.65381(2) & 0.750 & 0.00000(3) & 0.00000 & +0.00 & +0.02 & +0.20 \\
  B & 3 & 2s$^{2}$2p & ${}^{2}P^{\circ}$ & -24.65379(2) & 0.750 & 0.00001(2) & 0.00000 & +0.01 & +0.02 & +0.20 \\
  B & 4 & 2s2p$^{2}$ & ${}^{4}P$ & -24.52201(2) & 3.750 & 0.13179(3) & 0.13153 & +0.27 & +0.33 & -0.47 \\
  B & 5 & 2s2p$^{2}$ & ${}^{4}P$ & -24.52201(1) & 3.750 & 0.13180(2) & 0.13153 & +0.27 & +0.33 & -0.42 \\
  B & 6 & 2s2p$^{2}$ & ${}^{4}P$ & -24.52200(2) & 3.750 & 0.13180(2) & 0.13153 & +0.27 & +0.34 & -0.33 \\
  B & 7 & 2s$^{2}$3s & ${}^{2}S$ & -24.47139(3) & 0.750 & 0.18242(4) & 0.18239 & +0.03 & +0.10 & -0.09 \\
  B & 8 & 2s2p$^{2}$ & ${}^{2}D$ & -24.43588(3) & 0.750 & 0.21793(4) & 0.21801 & -0.08 & +0.07 & +0.89 \\
  B & 9 & 2s2p$^{2}$ & ${}^{2}D$ & -24.43584(4) & 0.750 & 0.21797(4) & 0.21801 & -0.03 & +0.07 & +1.39 \\
  B & 10 & 2s2p$^{2}$ & ${}^{2}D$ & -24.43229(3) & 0.750 & 0.22151(3) & 0.21801 & +3.51 & +3.62 & +1.79 \\
\cmidrule(lr){2-11}
  C & 1 & 2s$^{2}$2p$^{2}$ & ${}^{3}P$ & -37.84489(2) & 2.000 & 0.00000 & 0.00000 & +0.00 & --- & --- \\
  C & 2 & 2s$^{2}$2p$^{2}$ & ${}^{3}P$ & -37.84482(3) & 2.000 & 0.00007(2) & 0.00000 & +0.07 & +0.04 & +0.00 \\
  C & 3 & 2s$^{2}$2p$^{2}$ & ${}^{3}P$ & -37.84472(4) & 2.000 & 0.00016(4) & 0.00000 & +0.16 & +0.06 & +0.34 \\
  C & 4 & 2s$^{2}$2p$^{2}$ & ${}^{1}D$ & -37.79844(4) & 0.000 & 0.04645(5) & 0.04631 & +0.14 & +0.03 & +0.19 \\
  C & 5 & 2s$^{2}$2p$^{2}$ & ${}^{1}D$ & -37.79843(2) & 0.000 & 0.04645(2) & 0.04631 & +0.15 & +0.03 & +0.54 \\
  C & 6 & 2s$^{2}$2p$^{2}$ & ${}^{1}D$ & -37.79841(3) & 0.000 & 0.04648(3) & 0.04631 & +0.17 & +0.04 & +0.79 \\
  C & 7 & 2s$^{2}$2p$^{2}$ & ${}^{1}D$ & -37.79839(2) & 0.000 & 0.04650(3) & 0.04631 & +0.19 & +0.04 & +0.89 \\
  C & 8 & 2s$^{2}$2p$^{2}$ & ${}^{1}D$ & -37.79837(4) & 0.000 & 0.04652(5) & 0.04631 & +0.21 & +0.11 & +1.05 \\
  C & 9 & 2s$^{2}$2p$^{2}$ & ${}^{1}S$ & -37.74631(3) & 0.000 & 0.09857(3) & 0.09850 & +0.07 & +0.04 & +1.00 \\
  C & 10 & 2s2p$^{3}$ & ${}^{5}S^{\circ}$ & -37.69174(1) & 6.000 & 0.15315(2) & 0.15357 & -0.42 & -0.40 & -0.52 \\
\cmidrule(lr){2-11}
  N & 1 & 2s$^{2}$2p$^{3}$ & ${}^{4}S^{\circ}$ & -54.58905(3) & 3.750 & 0.00000 & 0.00000 & +0.00 & --- & --- \\
  N & 2 & 2s$^{2}$2p$^{3}$ & ${}^{2}D^{\circ}$ & -54.50120(4) & 0.750 & 0.08784(5) & 0.08761 & +0.23 & +0.00 & +1.89 \\
  N & 3 & 2s$^{2}$2p$^{3}$ & ${}^{2}D^{\circ}$ & -54.50120(5) & 0.750 & 0.08785(6) & 0.08761 & +0.24 & +0.03 & +2.19 \\
  N & 4 & 2s$^{2}$2p$^{3}$ & ${}^{2}D^{\circ}$ & -54.50117(4) & 0.750 & 0.08787(4) & 0.08761 & +0.27 & +0.05 & +2.79 \\
  N & 5 & 2s$^{2}$2p$^{3}$ & ${}^{2}D^{\circ}$ & -54.50113(5) & 0.750 & 0.08792(6) & 0.08761 & +0.31 & +0.05 & +3.09 \\
  N & 6 & 2s$^{2}$2p$^{3}$ & ${}^{2}D^{\circ}$ & -54.50103(6) & 0.750 & 0.08802(8) & 0.08761 & +0.41 & +0.08 & +3.19 \\
  N & 7 & 2s$^{2}$2p$^{3}$ & ${}^{2}P^{\circ}$ & -54.45746(6) & 0.750 & 0.13159(7) & 0.13140 & +0.19 & -0.00 & +3.40 \\
  N & 8 & 2s$^{2}$2p$^{3}$ & ${}^{2}P^{\circ}$ & -54.45737(3) & 0.750 & 0.13168(5) & 0.13140 & +0.28 & +0.02 & +4.30 \\
  N & 9 & 2s$^{2}$2p$^{3}$ & ${}^{2}P^{\circ}$ & -54.45726(6) & 0.750 & 0.13178(6) & 0.13140 & +0.38 & +0.02 & +5.10 \\
  N & 10 & 2s$^{2}$2p$^{2}$(${}^{3}P$)3s & ${}^{4}P$ & -54.20916(4) & 3.750 & 0.37988(5) & 0.37971 & +0.18 & +1.16 & +0.59 \\
\cmidrule(lr){2-11}
  O & 1 & 2s$^{2}$2p$^{4}$ & ${}^{3}P$ & -75.06706(4) & 2.000 & 0.00000 & 0.00000 & +0.00 & --- & --- \\
  O & 2 & 2s$^{2}$2p$^{4}$ & ${}^{3}P$ & -75.06705(3) & 2.000 & 0.00001(5) & 0.00000 & +0.01 & +0.00 & +0.59 \\
  O & 3 & 2s$^{2}$2p$^{4}$ & ${}^{3}P$ & -75.06703(3) & 2.000 & 0.00003(7) & 0.00000 & +0.03 & +0.06 & +0.79 \\
  O & 4 & 2s$^{2}$2p$^{4}$ & ${}^{1}D$ & -74.99483(6) & 0.000 & 0.07223(7) & 0.07194 & +0.29 & +0.02 & -2.85 \\
  O & 5 & 2s$^{2}$2p$^{4}$ & ${}^{1}D$ & -74.99479(13) & 0.000 & 0.07227(15) & 0.07194 & +0.32 & +0.14 & +1.35 \\
  O & 6 & 2s$^{2}$2p$^{4}$ & ${}^{1}D$ & -74.99477(8) & 0.000 & 0.07229(9) & 0.07194 & +0.35 & +0.14 & +1.65 \\
  O & 7 & 2s$^{2}$2p$^{4}$ & ${}^{1}D$ & -74.99476(5) & 0.000 & 0.07230(5) & 0.07194 & +0.36 & +0.14 & +1.65 \\
  O & 8 & 2s$^{2}$2p$^{4}$ & ${}^{1}D$ & -74.99470(5) & 0.000 & 0.07236(6) & 0.07194 & +0.42 & +0.20 & +2.25 \\
  O & 9 & 2s$^{2}$2p$^{4}$ & ${}^{1}S$ & -74.91316(5) & 0.000 & 0.15390(6) & 0.15362 & +0.28 & +0.07 & +1.68 \\
  O & 10 & 2s$^{2}$2p$^{3}$(${}^{4}S^{\circ}$)3s & ${}^{5}S^{\circ}$ & -74.73097(2) & 6.000 & 0.33609(3) & 0.33576 & +0.33 & +1.61 & +0.06 \\
\cmidrule(lr){2-11}
  F & 1 & 2s$^{2}$2p$^{5}$ & ${}^{2}P^{\circ}$ & -99.73354(8) & 0.750 & 0.00000 & 0.00000 & +0.00 & --- & --- \\
  F & 2 & 2s$^{2}$2p$^{5}$ & ${}^{2}P^{\circ}$ & -99.73343(8) & 0.750 & 0.00012(15) & 0.00000 & +0.12 & +0.16 & +1.30 \\
  F & 3 & 2s$^{2}$2p$^{5}$ & ${}^{2}P^{\circ}$ & -99.73342(7) & 0.750 & 0.00013(9) & 0.00000 & +0.13 & +0.21 & +1.70 \\
  F & 4 & 2s$^{2}$2p$^{4}$(${}^{3}P$)3s & ${}^{4}P$ & -99.26623(9) & 3.750 & 0.46731(13) & 0.46673 & +0.59 & +1.20 & +0.17 \\
  F & 5 & 2s$^{2}$2p$^{4}$(${}^{3}P$)3s & ${}^{4}P$ & -99.26618(5) & 3.750 & 0.46736(9) & 0.46673 & +0.63 & +1.30 & +1.17 \\
  F & 6 & 2s$^{2}$2p$^{4}$(${}^{3}P$)3s & ${}^{4}P$ & -99.26602(7) & 3.750 & 0.46752(11) & 0.46673 & +0.79 & +1.42 & +1.27 \\
  F & 7 & 2s$^{2}$2p$^{4}$(${}^{3}P$)3s & ${}^{2}P$ & -99.25567(11) & 0.750 & 0.47788(14) & 0.47707 & +0.81 & +1.04 & +1.93 \\
  F & 8 & 2s$^{2}$2p$^{4}$(${}^{3}P$)3s & ${}^{2}P$ & -99.25564(7) & 0.750 & 0.47790(10) & 0.47707 & +0.83 & +1.04 & +1.93 \\
  F & 9 & 2s$^{2}$2p$^{4}$(${}^{3}P$)3s & ${}^{2}P$ & -99.25554(5) & 0.750 & 0.47800(9) & 0.47707 & +0.93 & +1.06 & +2.23 \\
  F & 10 & 2s$^{2}$2p$^{4}$(${}^{3}P$)3p & ${}^{4}P^{\circ}$ & -99.20465(7) & 3.750 & 0.52890(5) & 0.52790 & +0.99 & +9.68 & -0.90 \\
\cmidrule(lr){2-11}
  Ne & 1 & 2p$^{6}$ & ${}^{1}S$ & -128.93691(5) & 0.000 & 0.00000 & 0.00000 & +0.00 & --- & --- \\
  Ne & 2 & 2p$^{5}$(${}^{2}P^{\circ}_{3/2}$)3s & ${}^{2}[3/2]^{\circ}$ & -128.32451(11) & 2.000 & 0.61240(15) & 0.61145 & +0.95 & +1.40 & +2.95 \\
  Ne & 3 & 2p$^{5}$(${}^{2}P^{\circ}_{3/2}$)3s & ${}^{2}[3/2]^{\circ}$ & -128.32438(8) & 2.000 & 0.61253(11) & 0.61145 & +1.08 & +1.40 & +4.35 \\
  Ne & 4 & 2p$^{5}$(${}^{2}P^{\circ}_{3/2}$)3s & ${}^{2}[3/2]^{\circ}$ & -128.32409(7) & 2.000 & 0.61283(7) & 0.61145 & +1.37 & +1.40 & +5.05 \\
  Ne & 5 & 2p$^{5}$(${}^{2}P^{\circ}_{1/2}$)3s & ${}^{2}[1/2]^{\circ}$ & -128.31793(8) & 0.000 & 0.61898(11) & 0.61794 & +1.05 & +1.56 & -0.44 \\
  Ne & 6 & 2p$^{5}$(${}^{2}P^{\circ}_{1/2}$)3s & ${}^{2}[1/2]^{\circ}$ & -128.31782(8) & 0.000 & 0.61909(11) & 0.61794 & +1.16 & +1.56 & -0.14 \\
  Ne & 7 & 2p$^{5}$(${}^{2}P^{\circ}_{1/2}$)3s & ${}^{2}[1/2]^{\circ}$ & -128.31771(11) & 0.000 & 0.61920(12) & 0.61794 & +1.27 & +1.71 & +0.76 \\
  Ne & 8 & 2p$^{5}$(${}^{2}P^{\circ}_{3/2}$)3p & ${}^{2}[1/2]$ & -128.26008(10) & 2.000 & 0.67684(12) & 0.67854 & -1.71 & +2.14 & +5.86 \\
  Ne & 9 & 2p$^{5}$(${}^{2}P^{\circ}_{3/2}$)3p & ${}^{2}[5/2]$ & -128.25296(12) & 2.000 & 0.68396(12) & 0.68220 & +1.75 & +3.95 & +4.20 \\
  Ne & 10 & 2p$^{5}$(${}^{2}P^{\circ}_{1/2}$)3p & ${}^{2}[3/2]$ & -128.25240(12) & 2.000 & 0.68451(14) & 0.68456 & -0.05 & +2.42 & +2.04 \\

\end{longtable}
}  %

\section{Compute resources and asset licenses}
\label{app:sec:compute_and_licenses}

\paragraph{Compute resources.}
Each run of Table~\ref{tab:atoms_mae} trains for $100{,}000$ KFAC steps at an electron batch of $4{,}096$ on one to four NVIDIA V100 (SXM2, $32$\,GB), two NVIDIA A40, or four NVIDIA H100~NVL.
A training step took $2.8$ to $9.3$\,s, depending on the atom and the cards, and these runs took over $3{,}600$ GPU-hours, estimated as the per-step time times the number of steps and cards.
CASSCF pretraining of Ne took $47$ minutes on four H100~NVL and $140$ minutes on two A40.
Preliminary work used about $2{,}880$ A6000-hours and $2{,}880$ A100-hours.
The wall-time sweep of Appendix~\ref{app:sec:scaling} ran on one host with $8{\times}$ RTX~5090.

\paragraph{Asset licenses.}
This work uses the following open-source software, used under their respective licenses:
\begin{itemize}[leftmargin=*,itemsep=2pt,topsep=2pt]
  \item \textbf{DeepQMC}~v1.2.0~\citep{deepqmc,Schatzle--Szabo--Mezera--Hermann--Noe2023} (MIT, \url{https://github.com/deepqmc/deepqmc}): base code, forked and modified for this work.
The modifications independent of the LoRA objective are the pretraining sampler (Appendix~\ref{app:sec:pretrain_details}), the exact distances in the reported energies (Appendix~\ref{app:sec:architectures}), and the handling of envelope underflow and non-finite steps (Appendix~\ref{app:sec:nan_handling}).
  \item \textbf{JAX}~v0.9.2 (Apache~2.0).
  \item \textbf{KFAC-JAX}~v0.0.8 (Apache~2.0): second-order optimizer.
  \item \textbf{folx}~v0.2.19 (MIT): forward-Laplacian primitive.
\end{itemize}
No new datasets are introduced.
Source code for \algo is available at \url{https://github.com/MinchanJeong/NestedLoRA-VMC} under the MIT license.

\end{document}